\documentclass[review]{elsarticle}
\graphicspath{ {./figures/} }
\usepackage{hyperref}
\usepackage{float}
\usepackage{verbatim} 
\usepackage{apalike}
\restylefloat{figure}
\restylefloat{table}
\usepackage{lineno,hyperref}
\usepackage{amsmath,amssymb,amsfonts}
\usepackage{algorithmic}
\usepackage{graphicx}
\usepackage{textcomp}
\usepackage{xcolor}
\usepackage{subcaption}
\usepackage{comment}
\usepackage{epsfig}
\usepackage{float}
\usepackage{ulem}
\usepackage{longtable}

\journal{Expert Systems with Applications}

\biboptions{authoryear}

\begin{document}
\begin{frontmatter}

\begin{titlepage}
\begin{center}
\vspace*{1cm}

\textbf{ \large Kernel function impact on convolutional neural networks}

\vspace{1.5cm}

M.Amine Mahmoudi$^{a}$ (mohamed.mahmoudi@univ-mascara.dz), Aladine Chetouani$^b$, Fatma Boufera$^a$, Hedi Tabia$^c$ \\

\hspace{10pt}

\begin{flushleft}
\small  
$^a$ Mustapha Stambouli University of Mascara, Algeria \\
$^b$ PRISME laboratory, University of Orleans, Orleans, France \\
$^c$ Université Paris Saclay, IBISC, Univ Evry, Evry, France

\vspace{1cm}
\textbf{Corresponding Author:} \\
M.Amine Mahmoudi \\
Email: mohamed.mahmoudi@univ-mascara.dz

\end{flushleft}        
\end{center}
\end{titlepage}

\title{Kernel function impact on convolutional neural networks}

\author[label1]{M.Amine Mahmoudi  \corref{cor1}}
\ead{mohamed.mahmoudi@univ-mascara.dz}

\author[label2]{Aladine Chetouani}

\author[label1]{Fatma Boufera}

\author[label3]{Hedi Tabia}

\cortext[cor1]{Corresponding author.}
\address[label1]{Mustapha Stambouli University of Mascara, Algeria}
\address[label2]{PRISME laboratory, University of Orleans, Orleans, France}
\address[label3]{Université Paris Saclay, IBISC, Univ Evry, Evry, France}

\begin{abstract}
This paper investigates the usage of kernel functions at the different layers in a convolutional neural network. We carry out extensive studies of their impact on convolutional, pooling and fully-connected layers. We notice that the linear kernel may not be sufficiently effective to fit the input data distributions, whereas high order kernels prone to over-fitting. This leads to conclude that a trade-off between complexity and performance should be reached. We show how one can effectively leverage kernel functions, by introducing a more distortion aware pooling layers which reduces over-fitting while keeping track of the majority of the information fed into subsequent layers. We further propose Kernelized Dense Layers (KDL), which replace fully-connected layers, and capture higher order feature interactions. The experiments on conventional classification datasets i.e. MNIST, FASHION-MNIST and CIFAR-10, show that the proposed techniques improve the performance of the network compared to classical convolution, pooling and fully connected layers. Moreover, experiments on fine-grained classification i.e. facial expression databases, namely RAF-DB, FER2013 and ExpW demonstrate that the discriminative power of the network is boosted, since the proposed techniques improve the awareness to slight visual details and allows the network reaching state-of-the-art results.
\end{abstract}


\end{frontmatter}
\section{Introduction}
\label{sec:Introduction}

Image classification has always been the core operation of computer vision. Several methods have been proposed in the literature addressing this problem, which consist in efficiently assigning the correct label to an image. Recently, with the emergence of Convolutional Neural Network (CNN), the computer vision community has witnessed an era of blossoming result thanks to the use of very large training databases. These databases contain a very large number of different images (i.e. objects, animals...etc). This advance encouraged the computer vision community to go beyond classical image classification that recognizes basic-level categories. The new challenge consists of discriminating categories that were considered previously as a single category and have only small subtle visual differences. This new sub-topic of image classification, called fine-grained image classification, is receiving a special attention from the computer vision community~\citep{liu20203d,tang2020revisiting,chen2020fine,ji2020attention,wang2020weakly,huang2020interpretable,gao2020channel,zhuang2020learning}. Such methods aims at discriminating between classes in a sub-category of objects like birds species~\citep{WelinderEtal2010}, models of cars~\citep{KrauseStarkDengFei_Fei_3DRR2013}, and facial expressions, which makes the classification more difficult due to the high intra-class and low inter-class variations. State-of-the-art approaches typically rely on convolutional neural network as classification backbone and propose a method to improve its awareness to subtle visual details.

A CNN is mainly a stack of three different types of layers: convolution layers, pooling layers and fully-connected layers. Each of these types of layers perform specific task. Convolution layers are the core building block of a CNN by leveraging the fact that an input image is composed of small details, or features, and create a mechanism for analyzing each feature in isolation, which makes a decision about the image as a whole. Pooling layers, on the other hand, are used for the gradual spatial down-sampling of the feature map by reducing the number of parameters and thus decreases both the consumption of the memory and the complexity of computing . In addition, pooling layers widen the receptive field size of the intermediate neurons which allow the latter to receive data from a larger area of the image. These two layers are usually used in alternation until getting the most size-effective representative feature which is finally fed into a fully connected neural network in order to take a final classification decision.

CNNs have been used for a multitude of visual tasks. They showed to perform very competitive results while linear operations are used at different layers of the network. Linear functions are efficient, particularly, when the original data is linearly separable, which should have, in general, a high dimensional representation. In such a case, the decision boundary can be representable as a linear combination of the original features. It is worth noting that not every high dimensional problems are linearly separable~\citep{robert2014machine}. For instance, images may have a high dimensional representation, but individual pixels are not very informative. Moreover, taking in consideration only small regions of the image, dramatically reduces their dimension, which makes linear functions less sensitive to subtle changes in input data. The ability of detecting such differences is crucial essentially for  fine-grained recognition.

To overcome these limitations, some researches investigated different ways to include non-linear functions in CNNs. Starting from non linear activation functions like ReLU, eLU~\citep{clevert2015fast}, SeLU~\citep{klambauer2017self} and more recently~\citep{sitzmann2020implicit}. Moreover, some recent work intended to replace the underlying linear function of a CNN by non linear kernel function without resorting to activation functions. For instance, some of them replaced convolution layers (\citep{zoumpourlis2017non}, \citep{wang2019kervolutional}), while others replaced the pooling layers  (\citep{lin2015bilinear}, \citep{tenenbaum2000separating}, \citep{lin2017improved}, \citep{gao2016compact}, \citep{cui2017kernel}, \citep{gao2019global}, ~\citep{hyun2019universal}). The higher order kernel function are, the more susceptible they are to fit slight changes in data. We leverage this kernel function property to find the best use of these functions in different level of the CNN.

In this paper, we investigate the usage of kernel functions at the different layers in a CNN. We carry out extensive studies of their impact on convolutional, pooling and fully-connected layers. For this purpose, we first replace the convolution operation in CNNs by a non-linear kernel function similarly to Kervolution~\citep{wang2019kervolutional}. We further used a novel pooling layer, based on kernel functions, that keeps the down-scaling aspect of the standard pooling function and brings various new features. This pooling layer relay on learnable weights that generalize ordinary pooling operations (i.e. average pooling and max pooling). Furthermore, it encodes patch-wise non-linearity. In this manner, the discrimination power of the full network is enhanced. The novel pooling, called learnable weights pooling, can be used at any level of the network and is fully differentiable, which allows the network to be trained in an end-to-end fashion. We also use a novel fully-connected layer in which we use kernel functions to create a neuron unit that uses a higher degree kernel function on its inputs rather than computing the weighted sum. The used layer, called Kernelized Dense Layers (KDL)~\citep{9190694,mahmoudi2021taylor,mahmoudi2022kernel}, is also differentiable and demonstrates its usefulness in the improvement of the discrimination power of the full network. We explore different dispositions and configurations of these layers to find which configuration gives the best accuracy without over-fitting

The remainder of this paper is organized as follow: Section~\ref{sec:Related_work} reviews similar work that tried to incorporate kernel functions in CNN models. Section~\ref{sec:Approach} gives more details about the methods (i.e. kervolution, learnable weights pooling and KDL) we used to incorporate kernel functions in a CNN model. Section~\ref{sec:Experiments} presents our experiments, datasets and results; and Section~\ref{sec:Discussion} discusses the obtained results and concludes the paper.

\section{Related work}
\label{sec:Related_work}

Kernel-based learning machines, like kernel Support Vector Machines (SVMs)~\citep{burges1999advances}, kernel Fisher Discriminant (KFD)~\citep{mika1999fisher}, and kernel Principal Component Analysis (KPCA)~\citep{scholkopf1998nonlinear}, have been widely used in literature for various tasks. Their ability to operate in a high-dimensional feature space, allow them to be more accurate than linear models. These approaches have shown practical relevance for some specific fields like computer vision. Even-though CNNs have boosted dramatically the pattern recognition field, some recent works intended to take advantage of kernel-based learning machines to further boost CNN performance.

Recently, Zoumpourlis et al.~\citep{zoumpourlis2017non} developed a second-order convolution method by exploring quadratic forms through the Volterra kernels. Constituting a richer function space, this method is used as approximation of the response profile of visual cells. However, in this method the number of parameters increases the training complexity exponentially. Wang et al.~\citep{wang2019kervolutional} proposed kervolution layers to replace the convolution layers in a CNN. This proposed layer uses non-linear kernel function instead of linear convolution function as core operation. These layers allow the model to capture higher order features at the convolutional level.

In~\citep{lin2015bilinear} a bilinear pooling method for fine-grained recognition has been proposed. This method is based on the second order pooling model proposed  by~\citep{tenenbaum2000separating}. This method applies, first, two CNNs to extract features from the input data. After that, it multiplies using the outer product each location of the outcoming feature maps. At the end, an image descriptor is obtained by sum-pooling the previous product of the outcoming feature maps. The author have further enhanced this model later~\citep{lin2017improved} by applying matrix function normalization. For instance, the matrix square-root, and the matrix logarithm. These methods have also been used for facial expression recognition~\citep{moi} and enhanced the overall accuracy to nearly 3\% over an ordinary CNN. However, these methods have high dimensions and they might be unsuitable for several of image analysis applications, where reduced complexity is needed. To alleviate this issue, Gao et al.~\citep{gao2016compact} proposed a compact representation of these methods. This new method has only a few thousand dimensions, yet it has a similar discriminative power as the full bilinear model. Cui et al~\citep{cui2017kernel} have further generalized~\citep{gao2016compact} method in the form of Taylor series kernel. This method captures non-linear and higher order features. This method is fully differentiable, and can be trained with a CNN in an end-to-end fashion. The methods cited above are used between the convolution layers and the fully connected layers. They expand the basis of the data, which enhances the discrimination power of the fully connected layers. This enhancement in discrimination is then back-propagated to the convolution layers. Even-thought the methods cited above enhances the CNN capacity, they do not carry learning in themselves and rely completely on the underlying CNN. Finally, we have also conducted some works~\cite{mahmoudi2021deep,mahmoudi2022deep} that studies the differences between our methods and the ordinary kernel methods proposed before (like Kernel SVM and Kervolution).

To the best of our knowledge, kernel functions have never been used in fully-connected layers. These layers are always used in their basic linear form. Although non-linearity can be reached by stacking multiple dense layers, it makes the model prone to overfit. Thus many recent architectures omit entirely the fully-connected layers.
In this paper, we investigate the usage of kernel functions at the different layers of a convolutional neural network, in particular, convolution, pooling and fully-connected layers. We build upon the previously cited works and propose novel non-linear functions specific for each type of layer. These functions replace the underlying linear functions of the specific layer, perform the same tasks and bring new features. By doing so, these new layers will be able to extract linear and non-linear relations between features.

\section{Study design}
\label{sec:Approach}

In this section, we present the different proposed kernel based techniques which can be plugged into a CNN. These techniques consist in new feature extraction, pooling and classification layers that can be used in the same manner as the usual CNN layers. Furthermore, these novel layers can be used solely or jointly with the usual CNN layers. This flexibility makes them usable in any architecture or even plugged at any level of a pre-trained CNN model. The novelty in this work is the use of higher order kernel function to replace the underlying function of each layer. These kernel functions allow them to perform the same task and brings additional features. 

\begin{figure}[h]
\begin{center}
\includegraphics[width=0.7\linewidth]{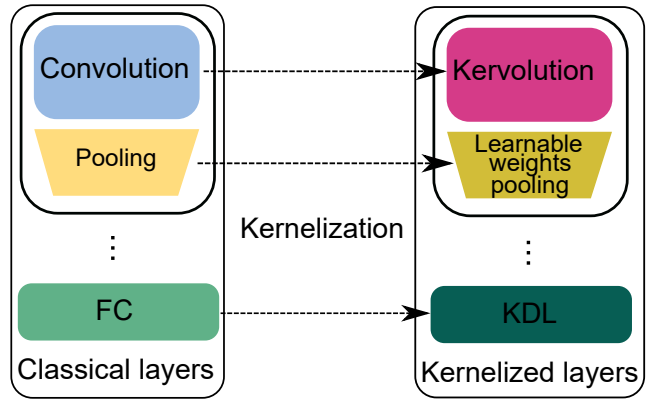}
\end{center}
\caption{The different configurations to replace each layer type of an ordinary CNN with a higher order kernel layer, notably Kervolution, Learnable weights pooling and kernelized dense layers KDL.}
\label{fig:configurations}
\end{figure}

In the following, we give details on how kernel functions can be employed at each level of a CNN. As shown in Figure~\ref{fig:configurations}, we replace each layer type of an ordinary CNN (Fig~\ref{fig:configurations}-(a)) with a higher order kernel layer (Fig~\ref{fig:configurations}-(b)). More precisely, we replace convolution layers with Kervolution layers, Max/AVG pooling layers with Learnable weights pooling layers, and fully connected layers with Kernelized dense layers KDL.

\subsection{Kervolution}

Kervolution has been proposed by Wang et al~\citep{wang2019kervolutional}. It extends the convolution operation which computes the dot product between an input vector $X$ and a weight vector $W$, and adds eventually a bias term, according to Equation \ref{eq:Convolution}:
\begin{equation}
 C_{o,i,j} = (X \times W)_{i,j} = \sum_{g} \sum_{h} X_{g+i,h+j} W_{o,g,h} + B_o,
\label{eq:Convolution}
\end{equation}
Where $o$ corresponds to the output size, $i$ and $j$ are specific locations in the input, $g$ and $h$ are respectively the width and height of the convolution filter and $B$ is a bias term.

Convolution is a linear operation that usually requires adding an activation function to introduce non-linearity. Without these activation functions the CNN performance drops dramatically. Kervolution leverages this fact and proposes to replace the convolution operation in CNNs by a non-linear function that performs the same task as convolution without resorting to activation functions. In this work, we use Kervolution with three kernel functions:

\begin{enumerate}
    \item \textbf{Linear kervolution} which corresponds to the convolution function (Equation~\ref{eq:Convolution});
    \item \textbf{Polynomial kervolution:}
    \begin{equation}
    K_{o,i,j} = \langle X,W\rangle_{i,j} = \sum_{g} \sum_{h} (X_{g+i,h+j} W_{o,g,h} + C)^n + B_o
    \label{eq:Poly_Kervolution}
    \end{equation}
    where $C$ ($C \in \mathbb {R^+}$) is a learnable constant and $n$ ($n \in \mathbb {Z^+}$) is the polynomial order, it extends the feature space to $n$ dimensions;
    
    \item \textbf{Gaussian RBF kervolution:}
    \begin{equation}
     K_{o,i,j} = e^{-{\frac {\|X_{g+i,h+j}-W_{o,g,h}\|^{2}}{2\sigma ^{2}}}}
     \label{eq:RBF_Kervolution}
     \end{equation}
     where $\sigma$ ($\sigma \in \mathbb {R^+} $) is a hyperparameter to control the smoothness of decision boundary. It extends kervolution to infinite dimensions.
\end{enumerate}

The linear kernel measures the similarity between the filter weight vector $W$ and the feature map vector $X$. However, when $n>1$ in Equation~\ref{eq:Poly_Kervolution}, the polynomial kernel encodes the non-linear relations between both $X$ and $W$ vectors, in addition to the linear relation between them. In the case of the Gaussian kernel (Equation~\ref{eq:RBF_Kervolution}), the non-linearity is expanded to the infinity. For this purpose, we can also use other kernels, such as the the Abel kernel defined as $ \mathcal{K}(X,W)=e^{-\alpha |X-W|}, $ or the Laplacian kernel defined by $ \mathcal{K}(X,W)=e^{-\alpha \|X-W\|}$,  $\text{ where, }\alpha >0$ in both kernels. 

\begin{figure}
\begin{center}
\includegraphics[width=0.7\linewidth]{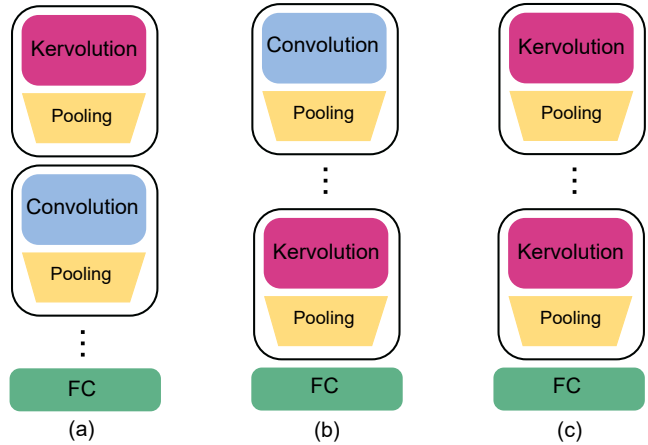}
\end{center}
\caption{The three study configurations of kervolution layers, namely: (a) one kervolution layer at the beginning of the network, (b) one kervolution layer at the end of the network and (c) an end-to-end kervolution network.}
\label{fig:Kerv-configurations}
\end{figure}

Kervolution layers are as flexible as convolution layers and can be plugged at any level of a CNN. In this paper, we use three configurations to study the kervolution layer impact on CNNs. As shown in Figure~\ref{fig:Kerv-configurations}, we use one kervolution layer at the beginning of the network (Fig~\ref{fig:Kerv-configurations}-(a)), one kervolution layer at the end of the network (Fig~\ref{fig:Kerv-configurations}-(b)), and an end-to-end kervolution network (Fig~\ref{fig:Kerv-configurations}-(c)).

\subsection{Learnable Weights Pooling}

Learnable weights pooling has been proposed in~\citep{mahmoudi2020learnable}. It extends the three conventional pooling methods that have usually been employed with CNNs, namely:

\begin{enumerate}
    \item \textbf{Max pooling} $P_{i,j}=Max(C_{i+g,j+h})$ where $C$ is the convolution layer output, $g,h$ is the pooling size and $i,j$ is a specific location. 
    \item \textbf{Average pooling} $P_{i,j}=\frac{\sum_{g}\sum_{h}(C_{i+g,j+h})} {gh}$
    \item \textbf{Strided convolution} (Equation \ref{eq:Convolution}). 
\end{enumerate}

These pooling methods exclude key details that are crucial for fine-grained classification. Particularly for FER, in which, the detection of specific distortions of facial regions is more important than simply detecting it in a given location (which is the case in max-pooling~\citep{Acharya_2018_CVPR_Workshops}).

Learnable weights pooling~\citep{mahmoudi2020learnable} rely on a kernel function that down-scale the input while keeping track of the most important information instead of deleting part of it. This gives pooling layers the ability to detect more sensitive and subtle details in data than the usual pooling methods. This is done by incorporating learnable weights to the pooling layers. In this manner, the usual pooling methods can be considered as a spaecial case of learnable weights pooling that have fixed (non-learnable) weights.

\begin{figure}
\begin{center}
\includegraphics[width=0.7\linewidth]{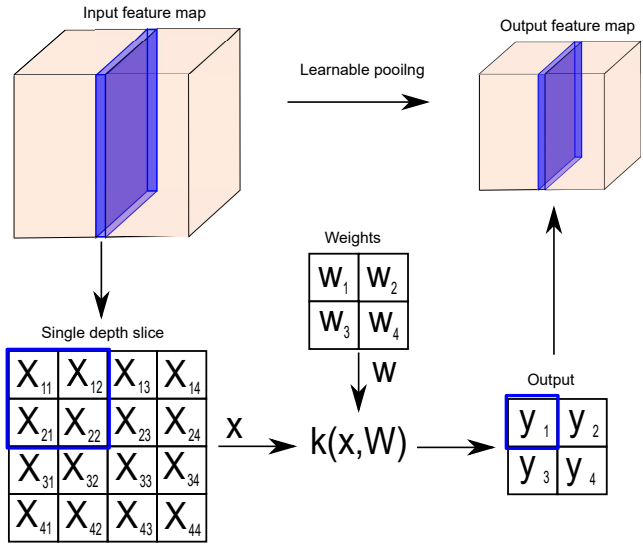}
\end{center}
\caption{The processing of the learnable weights pooling layer is similar to usual pooling layer in the manner that it down-scales the spatial dimensions of the input. Learnable weights pooling rely on learnable weights to encode important relations between features through kernel function.}
\label{fig:NonLinearPooling}
\end{figure}

Learnable weights pooling layer is the same as a usual pooling layer in the manner that it down-scales the spatial dimensions of feature map, on a specific location and a specific stride. It differs from other pooling strategies in the way that it has the capability to extract more important features from the input map. This is specially achieved by learning specific pooling weights at each location of the feature map. These weights are learned similarly to convolutional weights yet with a single depth output (see Figure \ref{fig:NonLinearPooling}). Finally, a combination between the input and the out-coming weights is calculated through a function chosen precisely  to capture non linear  and linear correlations between the weights and the input.

Here, we use three kernel functions as follow:

\begin{enumerate}
    \item Linear pooling is able to learn a suitable pooling from a continuum of methods that ranges from average to max (or extremum) pooling;
    \begin{equation}
        P_{i,j} = \sum_{g} \sum_{h} (C_{g+i,h+j} W_{g,h} + C)
    \end{equation} 
    
    \item Similarly to polynomial kervolution, the polynomial pooling does not only encode linear relation between inputs and weights, but linear relations too. It extends the feature space to $n$ dimension, yet reduces the size;
    \begin{equation}
        P_{i,j} = \sum_{g} \sum_{h} (C_{g+i,h+j} W_{g,h} + C)^n
    \end{equation}
    
    \item Gaussian RBF pooling expands the pooling non-linearity to the infinity;
    \begin{equation}
     P_{i,j} = e^{-{\frac {\|C_{g+i,h+j}-W_{g,h}\|^{2}}{2\sigma ^{2}}}}
     \end{equation}
\end{enumerate}

\begin{figure}[H]
\begin{center}
\includegraphics[width=0.7\linewidth]{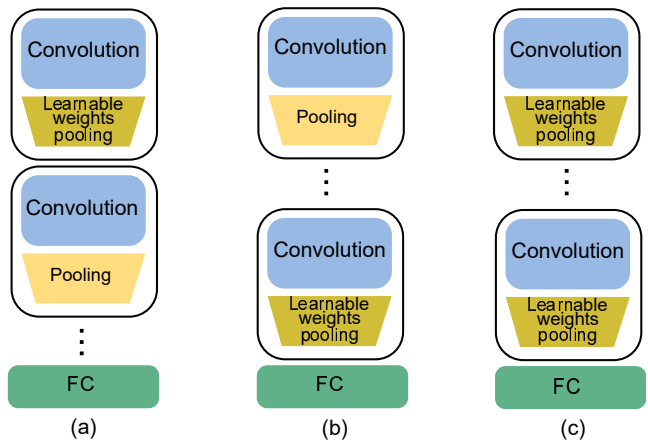}
\end{center}
\caption{The three study configurations of Learnable weights pooling layers, namely:
(a) one Learnable weights pooling layer after the first convolution layer, (b) one Learnable weights pooling layer after the last convolution layer, and (c) an end-to-end Learnable weights pooling network.}
\label{fig:pooling-configurations}
\end{figure}

This method is able to detect the suitable pooling strategy at a patch-wise level. In other words, in the same pooling layer, our method can apply a complex mixture of pooling method ranging from average pooling to max pooling (with a linear kernel) and beyond (with higher order kernel). Yet remaining as simple as an ordinary pooling layer. In order to efficiently study the impact of our proposed pooling layer on a CNN, we test it following the three configurations. As shown in Figure~\ref{fig:pooling-configurations}, we study the impact of one Learnable weights pooling layer after the first convolution layer (Fig~\ref{fig:pooling-configurations}-(a)), one Learnable weights pooling layer after the last convolution layer (Fig~\ref{fig:pooling-configurations}-(b)), and an end-to-end Learnable weights pooling network (Fig~\ref{fig:pooling-configurations}-(c)).

\subsection{Kernelized dense layer}

The inputs of the fully-connected layers consist of the result of the subsequent alternation of convolution and pooling layers. These input values goes through the first fully-connected layer where we multiply them by weights. After that, we apply an activation function (i.e ReLU). Finally, they goes forward through the output layer, in which each neuron represents a classification category. The fully-connected layers back-propagates the most accurate weights where every neuron gets weights that prioritize the most relevant category.

In this paper, we use a novel dense layer composed of neuron that uses a kernel function instead of the usual dot product. Kernelized Dense Layer (KDL), proposed in~\citep{9190694}. In contrary to a classical neuron layer where it computes a dot product between an input vector and a vector of weights, add a bias vector ($b\geq 0$), KDL  applies higher degree kernel function, which permits to the latter to map the input data to a higher space and therefore be more discriminative than a usual linear layer.

\begin{figure}[H]
\begin{center}
\includegraphics[width=0.7\linewidth]{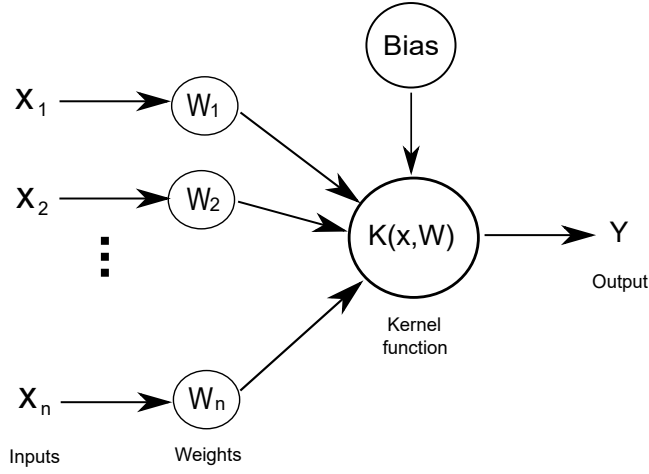}
\end{center}
\caption{The basic unit of KDL is a kernel based neuron. It applies a kernel function on a vector of weights $w = \{w_1,w_2,\dots,w_n\}$ and an input vector $x =\{x_1,x_2,\dots,x_n\}$ and adds a bias term.}
\label{fig:Kernel_Neuron}
\end{figure}

Figure \ref{fig:Kernel_Neuron} shows the processing of a core unit (kernel based neuron) of KDL. Formally, the output $Y$ is calculated using a kernel function $K$ on an input vector $x =\{x_1,x_2,\dots,x_n\}$ and the corresponding vector of weights $W = \{w_1,w_2,\dots,w_n\}$ and, adding the bias vector ($b\geq 0$).

\begin{enumerate}
    \item Linear KDL which correspond to the ordinary dense layer :
    \begin{equation}
    Y_{i} = \sum_{j} X_{j} W_{i,j} + B_i
    \label{eq:FC}
    \end{equation}
    
    \item Polynomial KDL :
    \begin{equation}
    Y_{i} = \sum_{j} (X_{j} W_{i,j}+C)^n + B_i
    \label{eq:Poly_KDL}
    \end{equation}
    where $C$ ($C \in \mathbb {R^+}$) is a learnable constant and $n$ ($n \in \mathbb {Z^+}$) is the polynomial order, it extends the KDL to $n$ dimensions;
    
    \item Gaussian RBF KDL:
    \begin{equation}
     Y_{i} = e^{-{\frac {\|X_{j}-W_{i,j}\|^{2}}{2\sigma ^{2}}}}
     \label{eq:RBF_KDL}
     \end{equation}
     where $\sigma$ ($\sigma \in \mathbb {R^+} $) is a hyperparameter to control the smoothness of decision boundary. It extends KDL to infinite dimensions.
\end{enumerate}

\section{Experiments}
\label{sec:Experiments}

In this section, we evaluate the performance of the kernel-based layers described above, in terms of accuracy rate and convergence speed. Several experiments have been conducted on six well-known datasets, following the configuration shown in Figure~\ref{fig:Kerv-configurations} and \ref{fig:pooling-configurations}. Note that for KDL, only one configuration can be tested, since they can be plugged only at the end of the network. In the following, we detail our experiments process. First we describe the datasets used to evaluate our approach (sec.\ref{sec:datasets}). After that we define the training process of our networks  (sec.\ref{sec:Model_Training}). Then we discuss the obtained results (sec.\ref{sec:AblationStudy}). Finally, we compare our results to state-of-the-art results (sec. \ref{comp-sota}).

\subsection{Datasets}
\label{sec:datasets}

Our experiments have been conducted on six well-known datasets. In order to study different characteristics of kernel functions in CNN, these datasets have been grouped in two categories. The first category is used to study the accuracy enhancement and convergence speed brought by the kernel function. It is composed of the following three datasets: \textbf{MNIST}, \textbf{Fashion-MNIST}~\citep{xiao2017fashion} and \textbf{CIFAR-10}~\citep{krizhevsky2009learning}. The second category is specially used to study the awareness of kernel function to subtle visual details. It is composed of three well-known fine-grained facial expression datasets: RAF-DB~\citep{li2017reliable}, ExpW~\citep{zhang2018facial} and FER2013~\citep{goodfellow2013challenges}. Facial expression datasets contain few classes that are nearly identical, which makes the recognition process more challenging.

\begin{itemize}
\item The \textbf{RAF-DB}~\citep{li2017reliable} or Real-world Affective Face DataBase is composed of  29,672 facial in the wild images. This images are categorized in either seven basic classes or eleven compound classes.

\item The \textbf{ExpW}~\citep{zhang2018facial} or  EXPression  in-the-Wild  dataset is composed of 91,793 facial in the wild images. The annotation was done manually on individual images.

\item The \textbf{FER2013} database was used for the ICML 2013 Challenges in Representation Learning \citep{goodfellow2013challenges}. It is composed of 28,709 training images and 3,589 images for both validation and test.
\end{itemize}

\subsection{ Training process}
\label{sec:Model_Training}

\begin{figure}[ht]
\begin{center}
\includegraphics[width=0.5\linewidth]{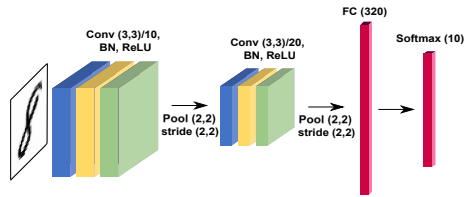}
\end{center}
\caption{Base model-1 architecture: it is composed of two blocks. Each one of these blocks is composed of a convolution layer, a batch normalization layer, a dropout layer and ReLU activation. At the end, two fully-connected layers are added with respectively 320 units and ReLU activation and 10 units with softmax activation.}
\label{fig:Base_Model-1}
\end{figure}

\begin{figure*}
\begin{center}
\includegraphics[width=\linewidth]{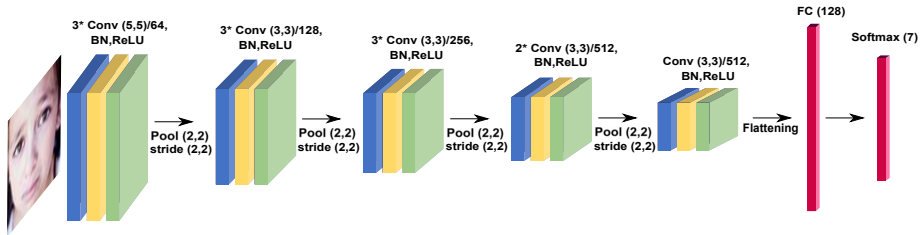}
\end{center}
\caption{Base model-2 architecture: it is composed of five blocks. Each one of these block is composed of a convolution layer, a batch normalization layer, a dropout layer and ReLU activation. At the end, two fully-connected layers are added with respectively 128 units and ReLU activation and 10 or 7 units with softmax activation.}
\label{fig:Base_Model-2}
\end{figure*}

The datasets we have used for our experiments are not similar in terms image complexity. MNIST and Fashion-MNIST contain small and very simple images, while Cifar-10 and FER datasets contain larger complex images. We have thus decided to use two model architectures, as shown in Figures~\ref{fig:Base_Model-1} and~\ref{fig:Base_Model-2}. These architectures are quite simple and can effectively run on cost-effective GPUs. Model-1 (Figure~\ref{fig:Base_Model-1}) is used for MNIST and Fashion-MNIST. It is composed of two blocks. Each one of these blocks is composed of a convolution layer, a batch normalization layer, a dropout layer and ReLU activation. At the end, two fully-connected layers are added with respectively 320 units and ReLU activation and 10 units with softmax activation. On the other hand, model-2 (Figure~\ref{fig:Base_Model-2}) is used for CIFAR-10 and FER datasets. It is composed of five blocks. Each one of these block is composed of a convolution layer, a batch normalization layer, a dropout layer and ReLU activation. At the end, two fully-connected layers are added with respectively 128 units and ReLU activation and 10 units, for CIFAR-10, or 7 units for FER datasets  with softmax activation.

We have used Adam optimiser with a learning rate starting from 0.001 decreased by a factor of 0.5 if the validation accuracy does not increase over two epochs for Model-1 and five epochs for Model-2. We trained Model-1 for 50 epochs and Model-2 for 100 epochs. To avoid over-fitting, for FER datasets, we used data augmentation
with a shear intensity of 0.2, a range degree for random rotations of 20, randomly flip inputs horizontally and a range for random zoom of 0.2. We have also cropped the face region on FER datasets and resize the resulting images to $100 \times 100$ pixels.

\subsection{Ablation Study}
\label{sec:AblationStudy}

In this section, we explore the impact of the three kernel-based layers in terms of accuracy rate and convergence speed. We study the impact of each layer solely, in a full non-linear configuration network. After that, we test them jointly with the usual CNN layer, plugging them either at the beginning or at the end of the network. The kernel functions we have used are: (1) the linear kernel functions, (2) polynomial kernel functions from second degree to degree five and (3) the Gaussian RBF kernel with $\sigma=0.9$. The accuracy rate results are illustrated in tables~\ref{tab:full-kerv}-~\ref{tab:KDL} and convergence speed results are illustrated in figures~\ref{fig:full-kerv}-~\ref{fig:KDL}. In order to distinguish clearly the difference in convergence among kernel, we chose to plot only the first 20 epochs for MNIST, Fashion-MNIST and Cifar-10 datasets.

\subsubsection{Kervolution}

\begin{figure*}[h]
        \begin{subfigure}[b]{0.33\textwidth}
                \centering
                \includegraphics[width=.95\linewidth]{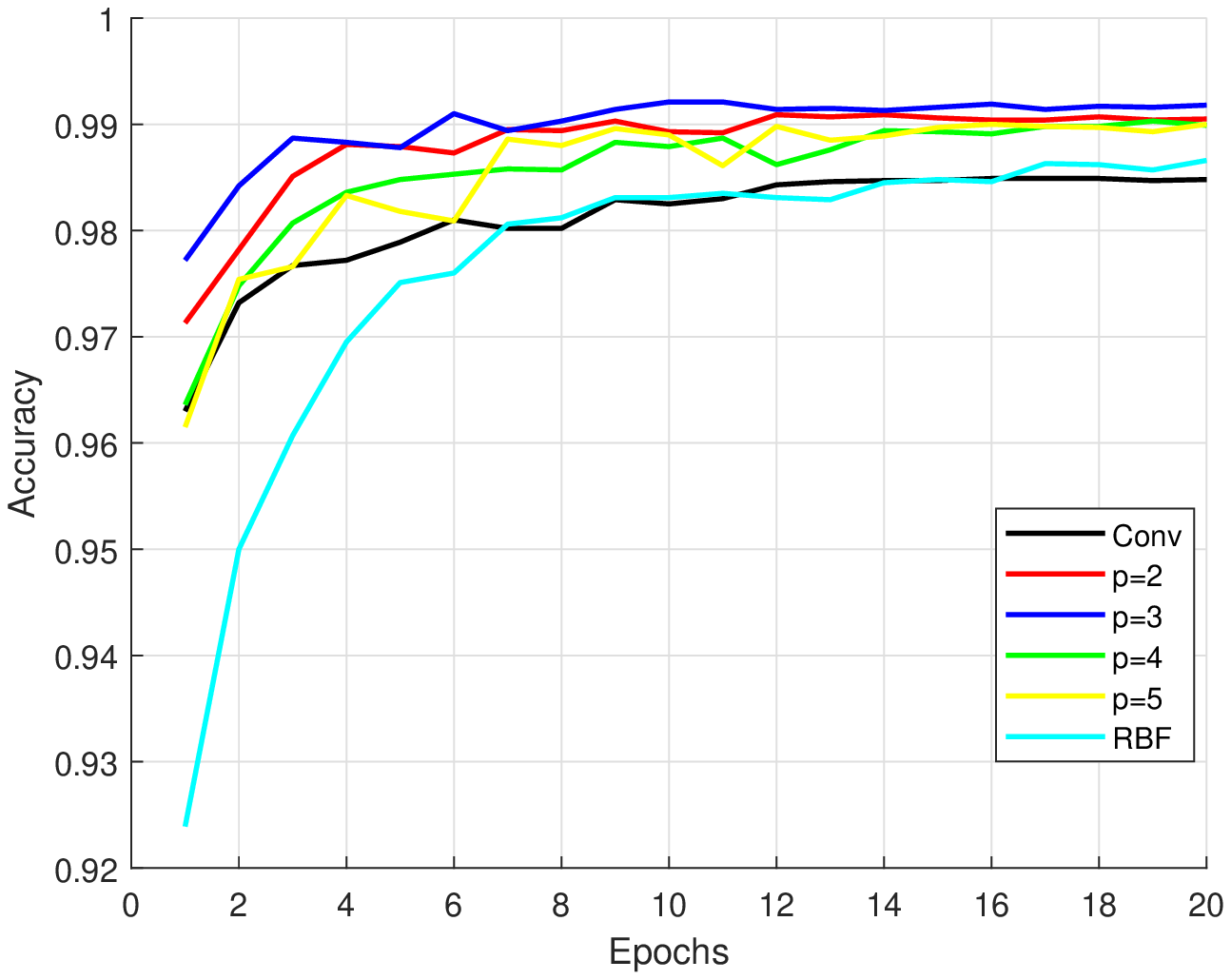}
                \caption{MNIST}
        \end{subfigure}%
        \begin{subfigure}[b]{0.33\textwidth}
                \centering
                \includegraphics[width=.95\linewidth]{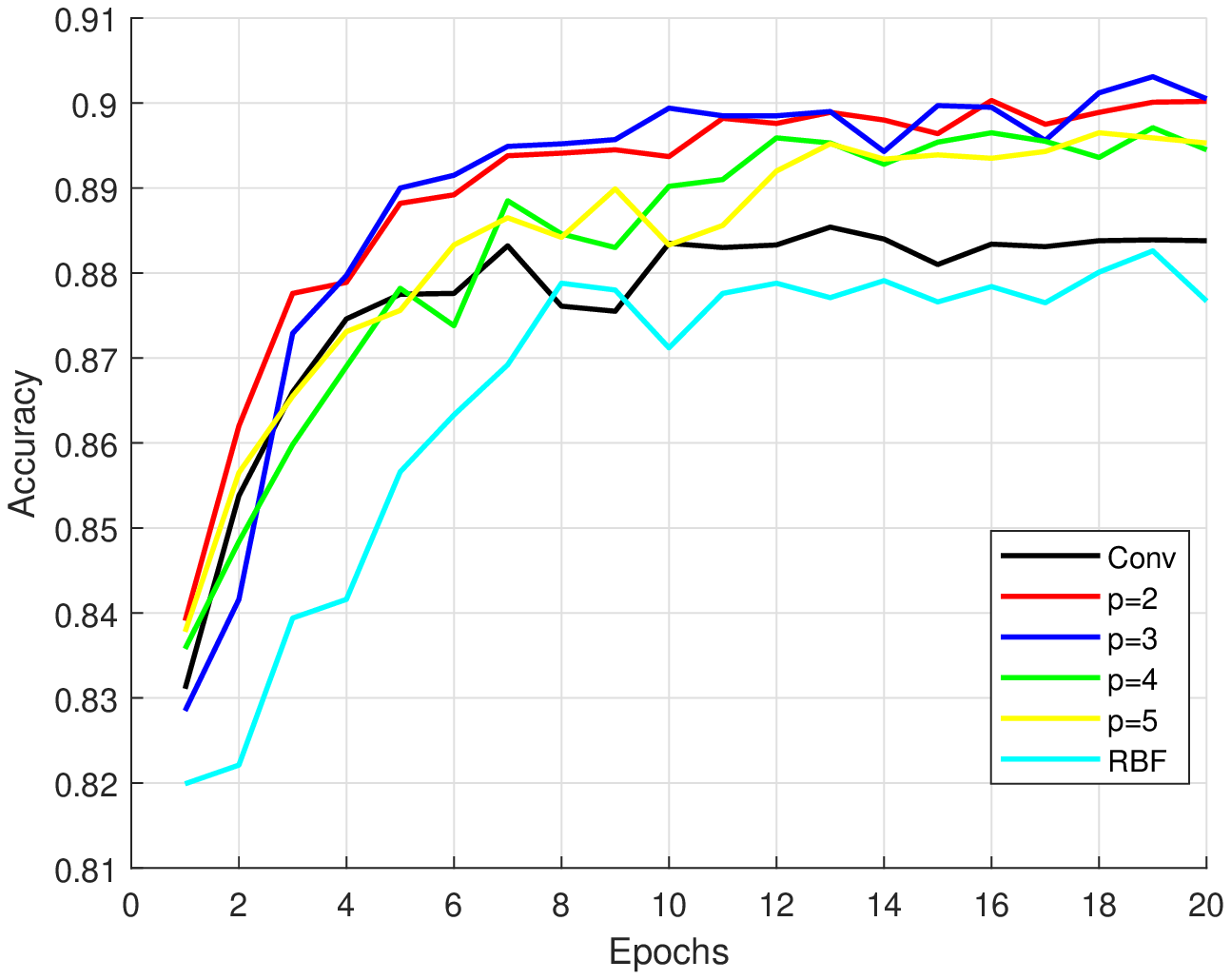}
                \caption{Fashion-MNIST}
        \end{subfigure}%
        \begin{subfigure}[b]{0.33\textwidth}
                \centering
                \includegraphics[width=.95\linewidth]{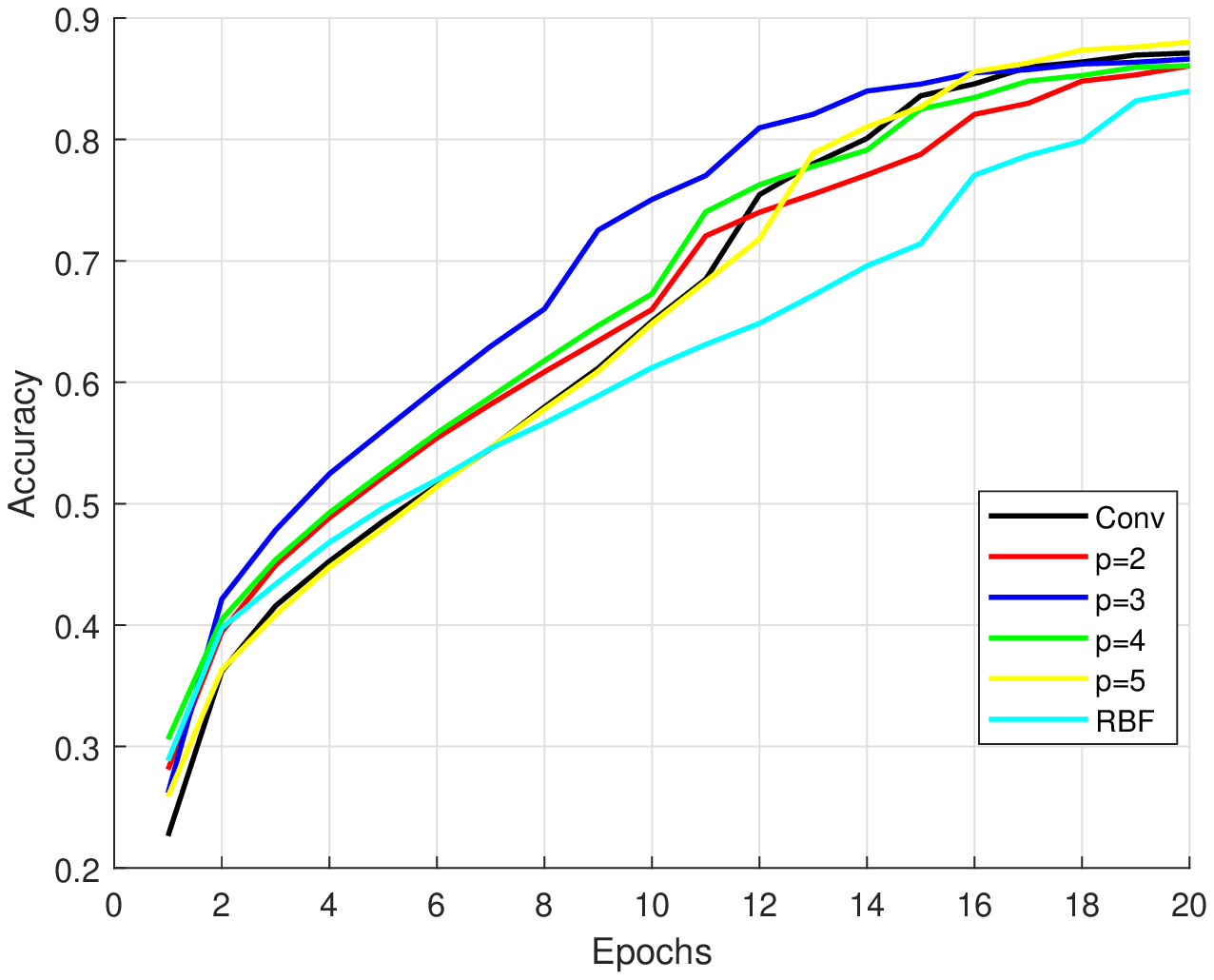}
                \caption{Cifar-10}
        \end{subfigure}
        \begin{subfigure}[b]{0.33\textwidth}
                \centering
                \includegraphics[width=.95\linewidth]{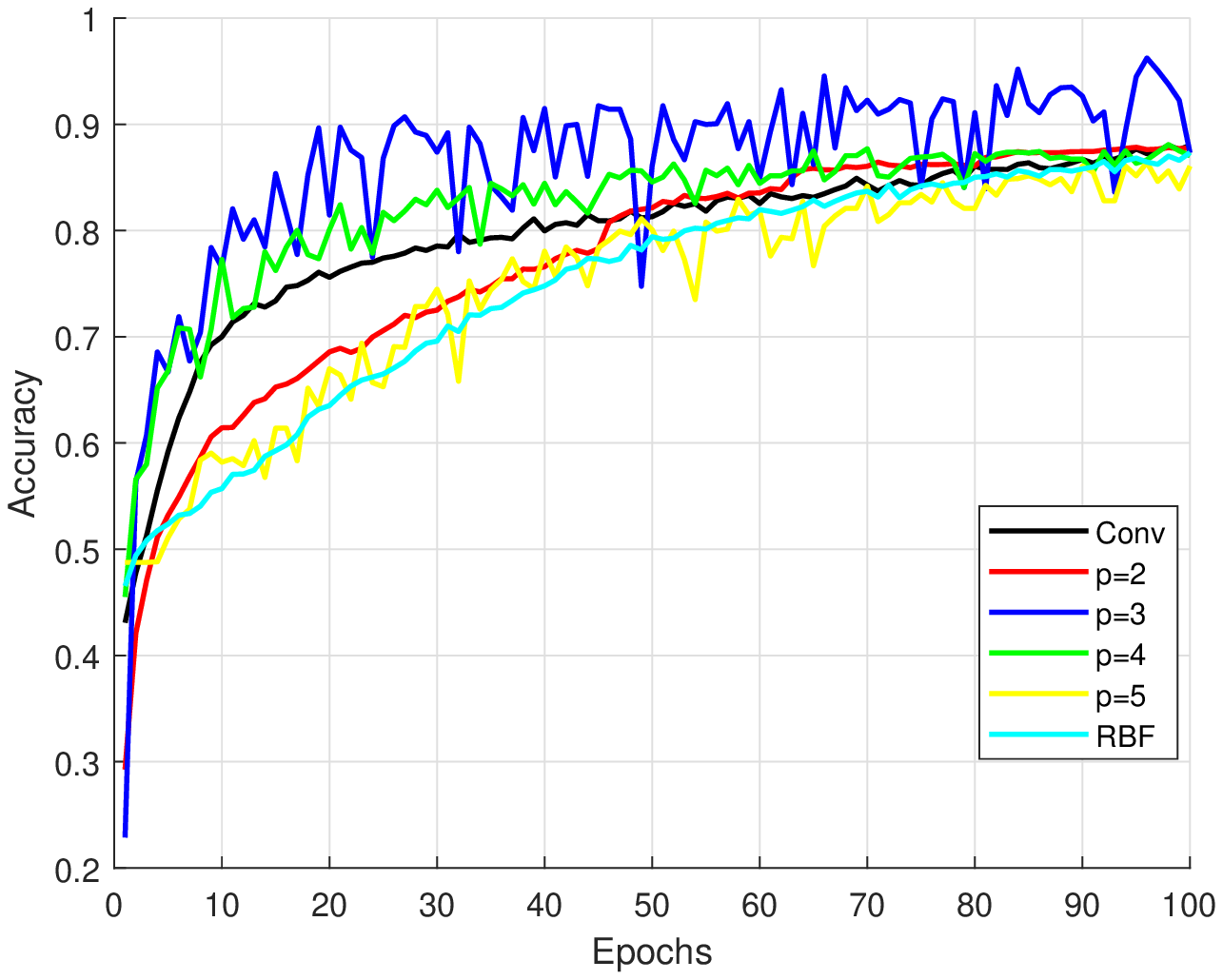}
                \caption{RAF-DB}
        \end{subfigure}%
        \begin{subfigure}[b]{0.33\textwidth}
                \centering
                \includegraphics[width=.95\linewidth]{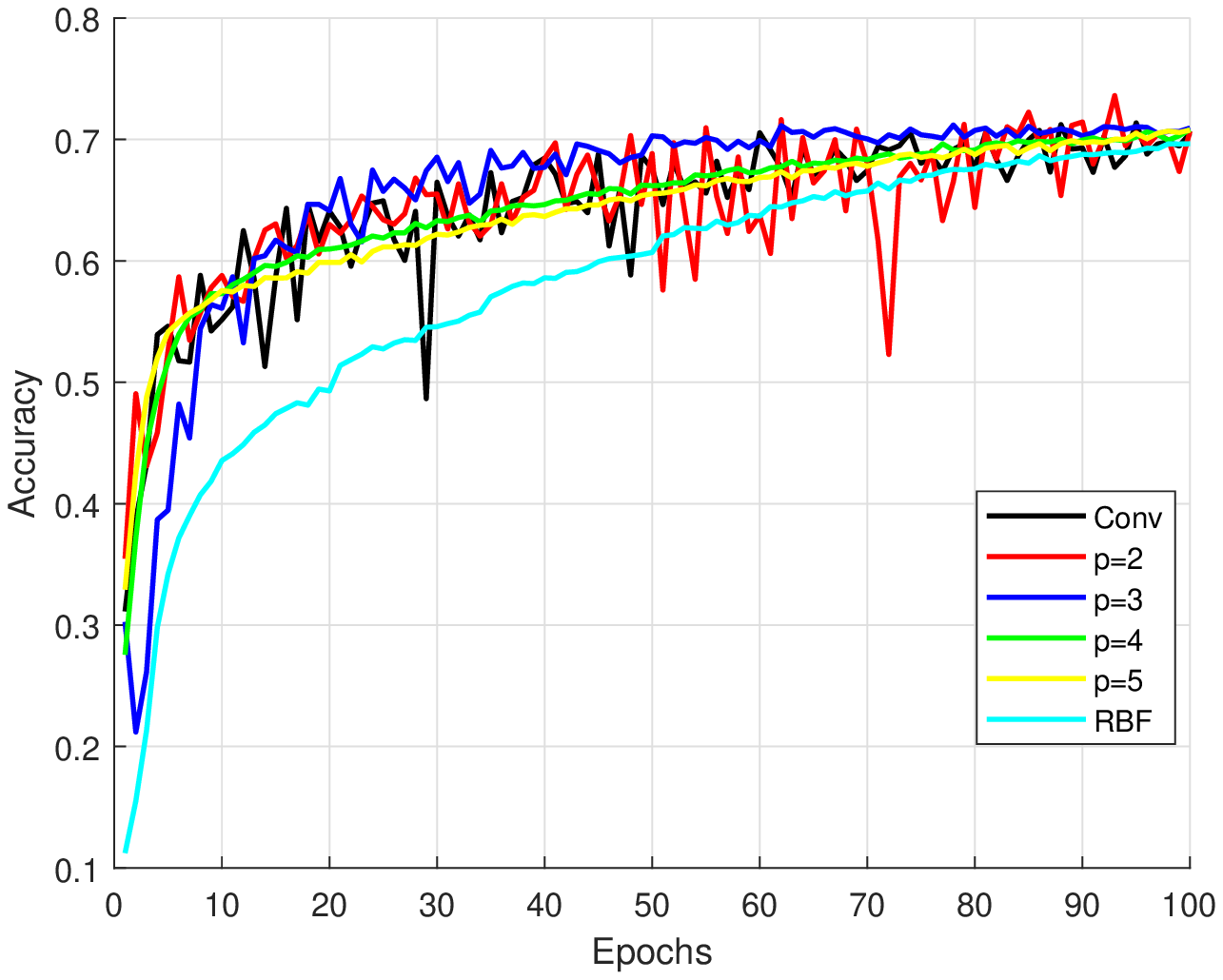}
                \caption{Fer2013}
        \end{subfigure}%
        \begin{subfigure}[b]{0.33\textwidth}
                \centering
                \includegraphics[width=.95\linewidth]{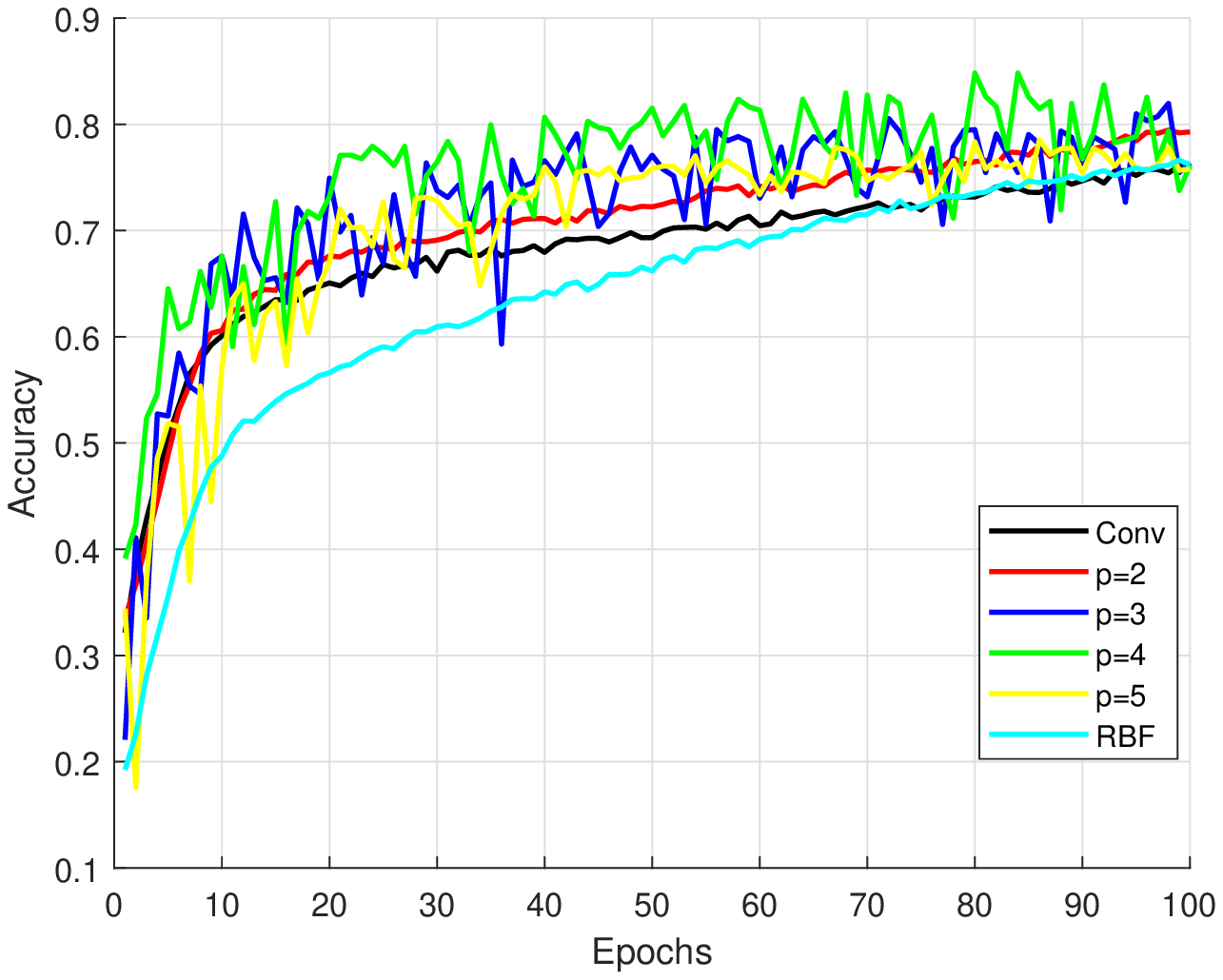}
                \caption{ExpW}
        \end{subfigure}
        \caption{Convergence of full kervolution networks}
        \label{fig:full-kerv}
\end{figure*}

\begin{figure*}
        \begin{subfigure}[b]{0.33\textwidth}
                \centering
                \includegraphics[width=.95\linewidth]{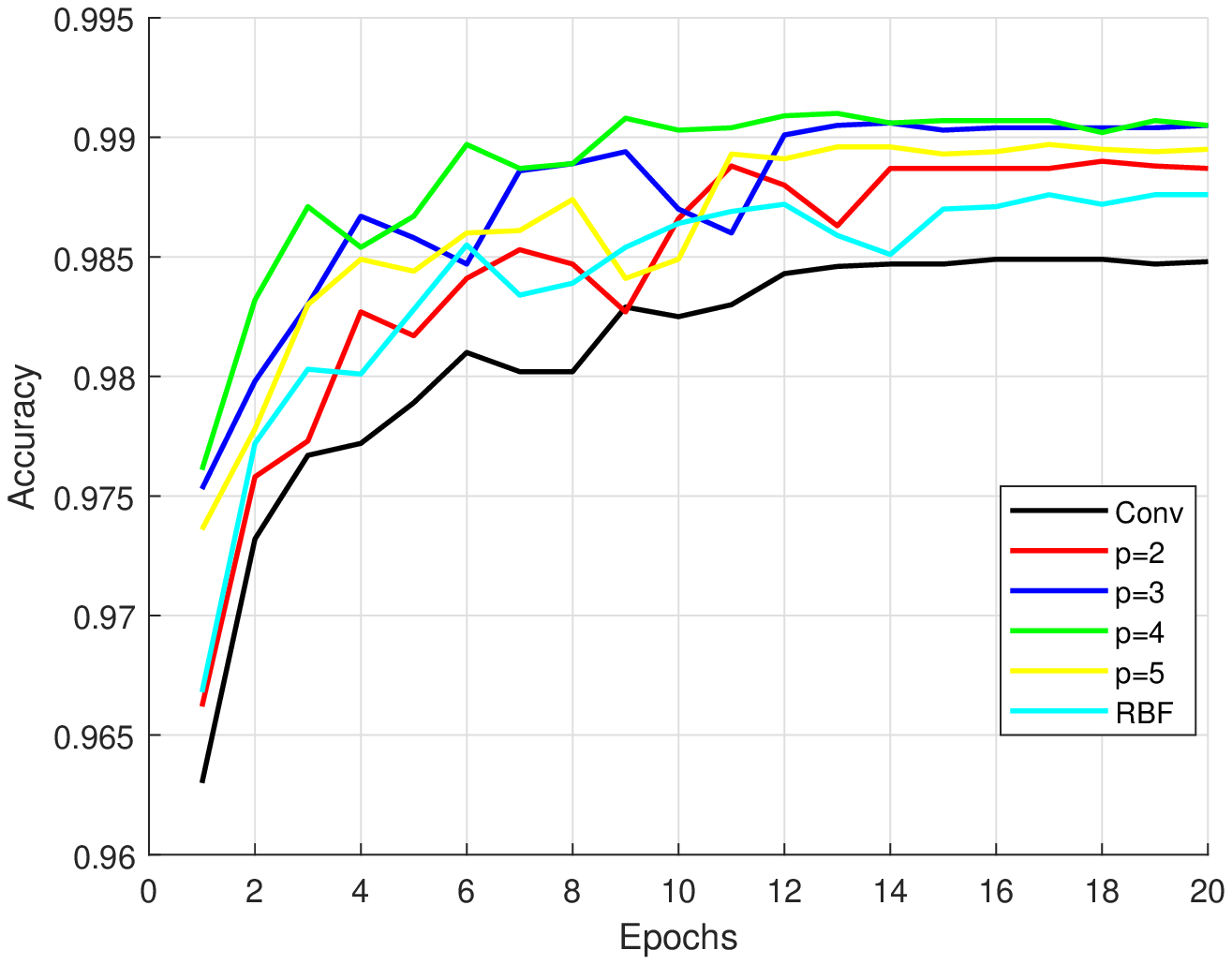}
                \caption{MNIST}
        \end{subfigure}%
        \begin{subfigure}[b]{0.33\textwidth}
                \centering
                \includegraphics[width=.95\linewidth]{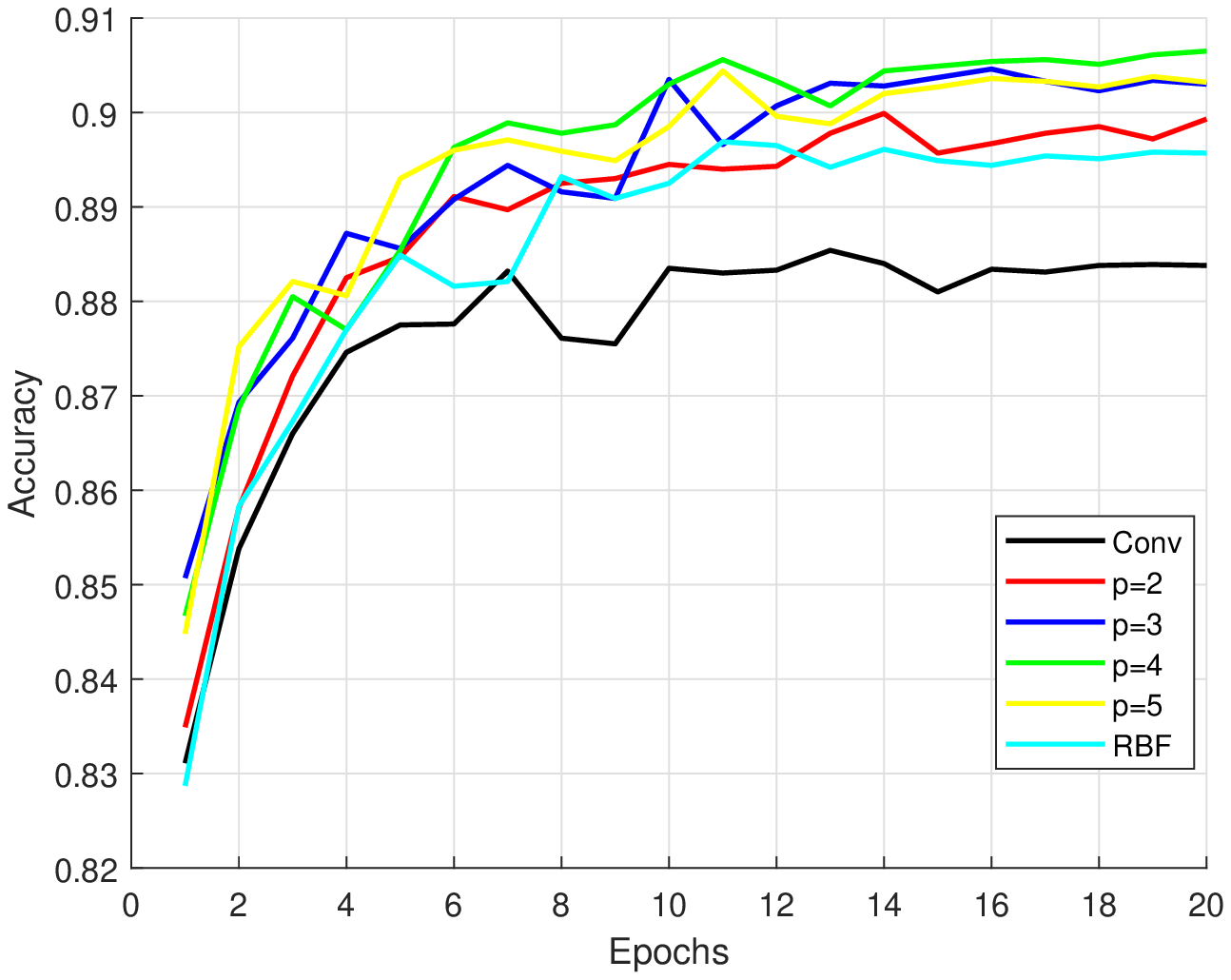}
                \caption{Fashion MNIST}
        \end{subfigure}%
        \begin{subfigure}[b]{0.33\textwidth}
                \centering
                \includegraphics[width=.95\linewidth]{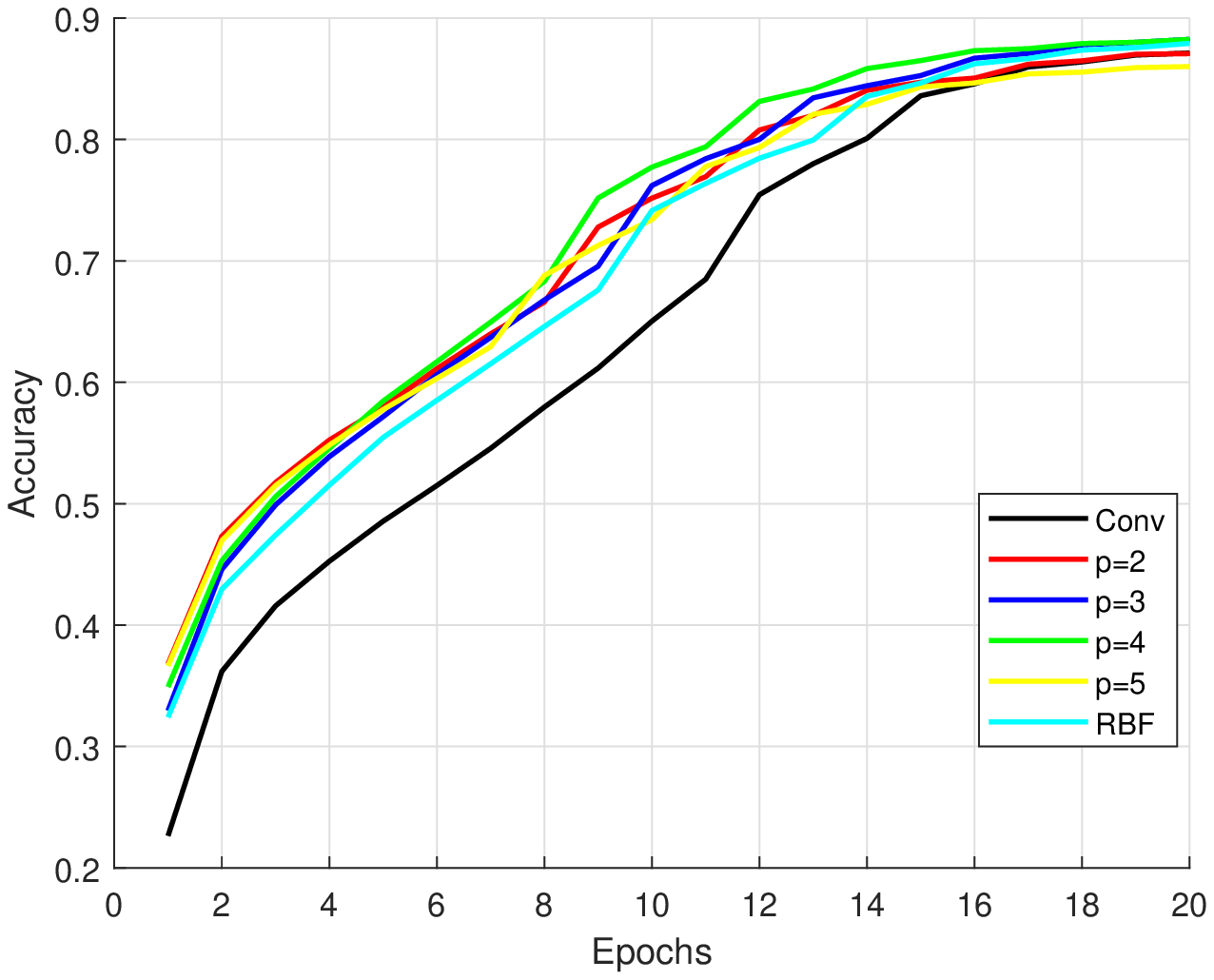}
                \caption{Cifar-10}
        \end{subfigure}
        \begin{subfigure}[b]{0.33\textwidth}
                \centering
                \includegraphics[width=.95\linewidth]{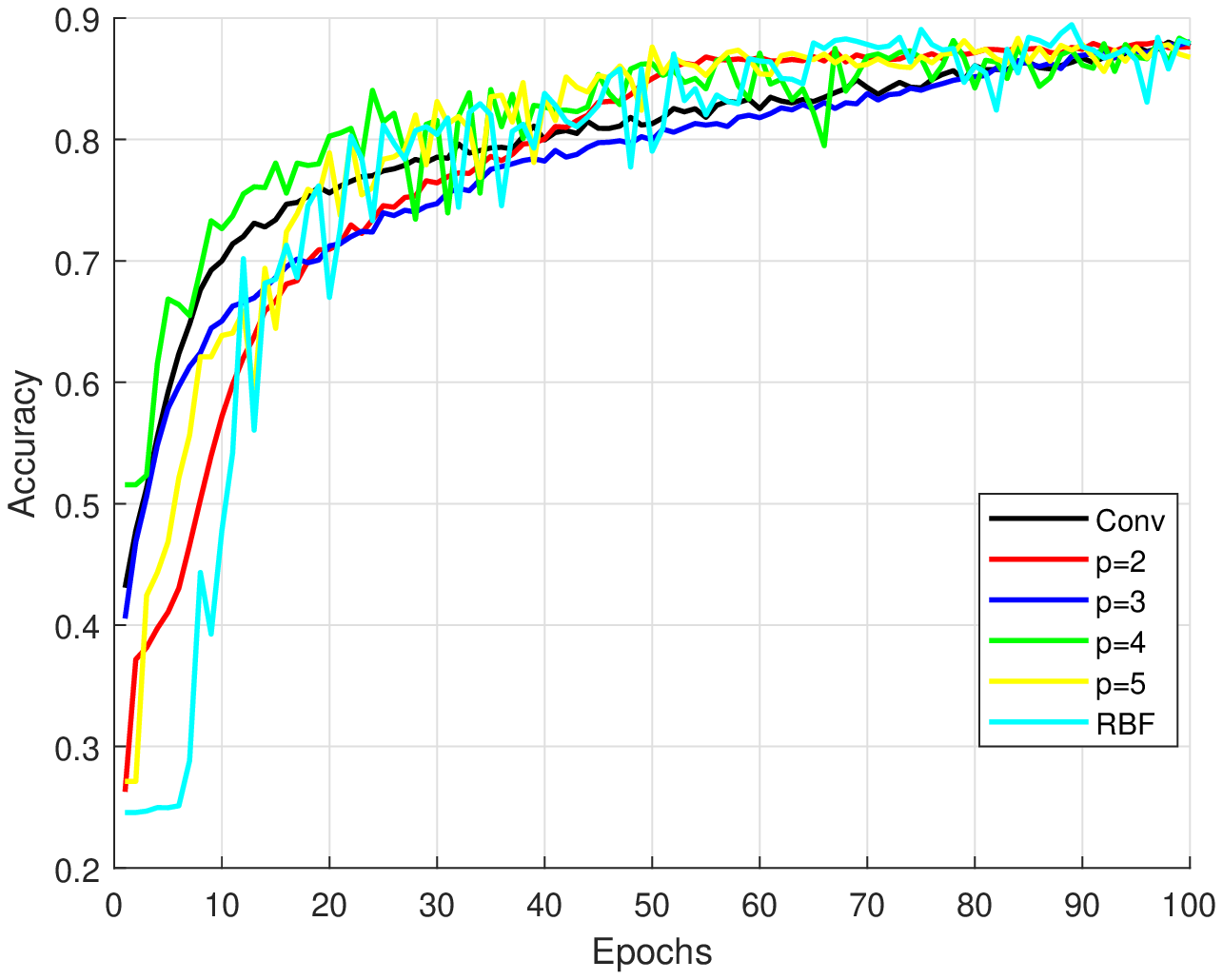}
                \caption{RAF-DB}
        \end{subfigure}%
        \begin{subfigure}[b]{0.33\textwidth}
                \centering
                \includegraphics[width=.95\linewidth]{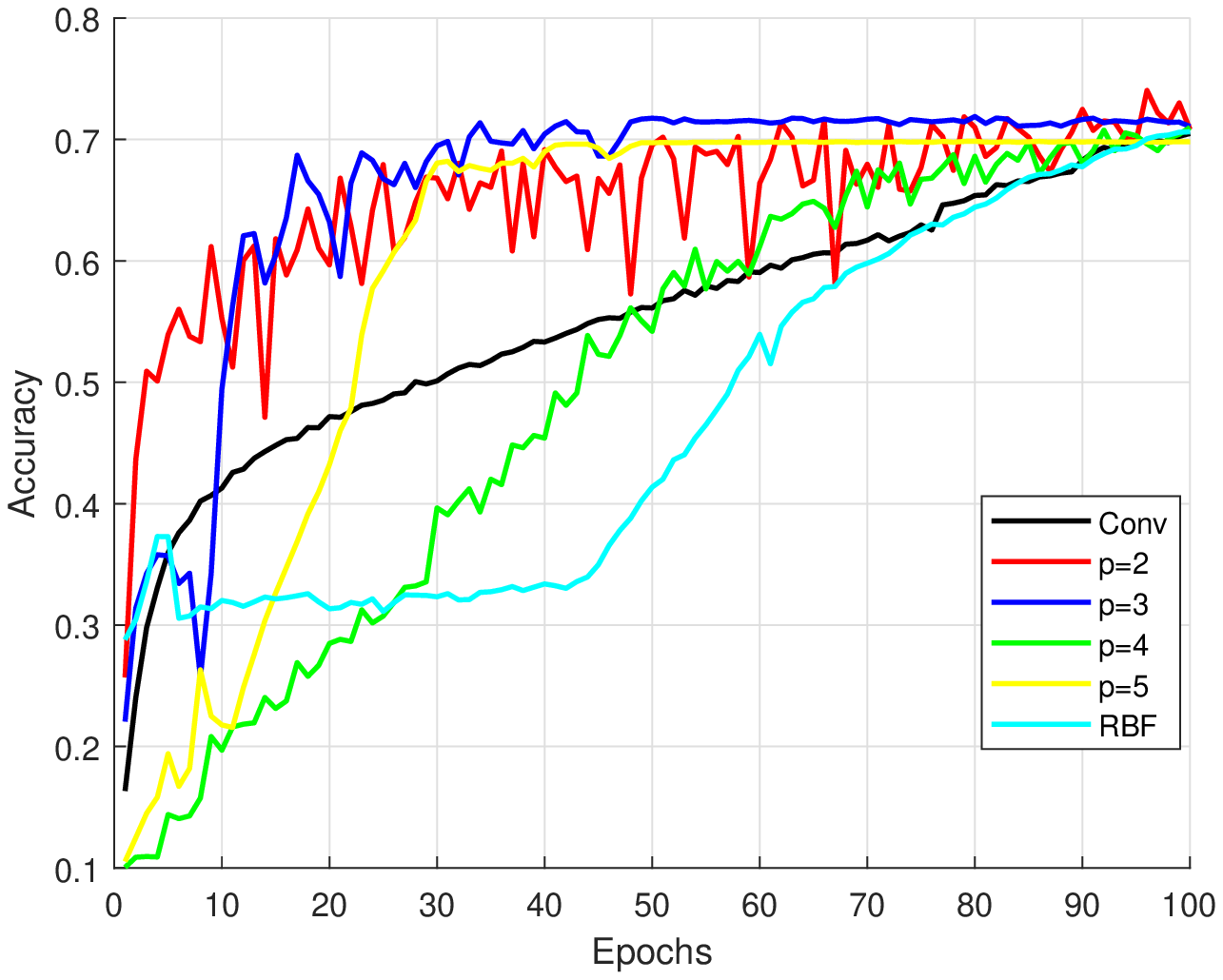}
                \caption{FER2013}
        \end{subfigure}%
        \begin{subfigure}[b]{0.33\textwidth}
                \centering
                \includegraphics[width=.95\linewidth]{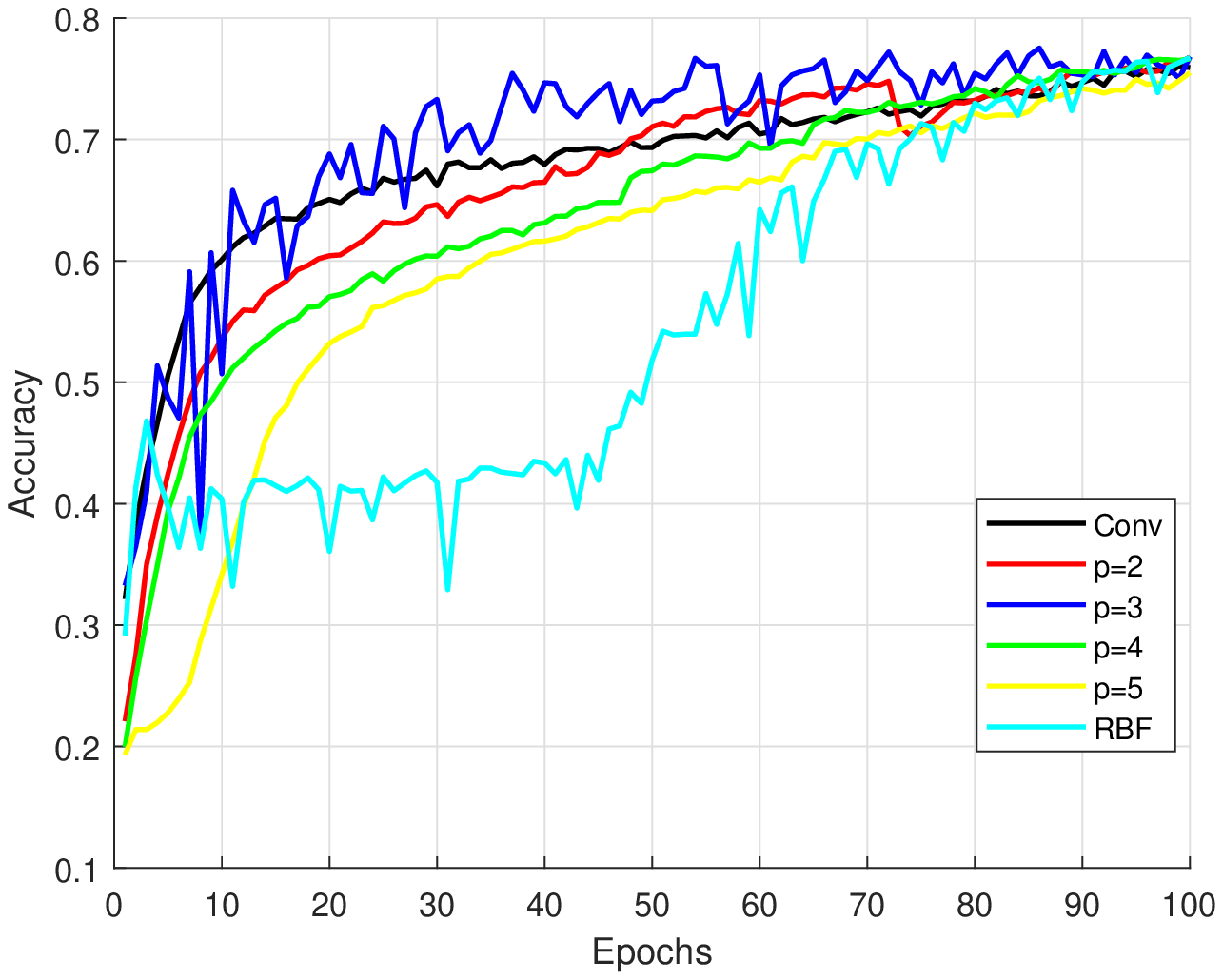}
                \caption{A ExpW}
        \end{subfigure}
        \caption{Convergence of networks with one kervolution layer at the begining}
        \label{fig:kerv-first}
\end{figure*}

\begin{figure*}[h]
        \begin{subfigure}[b]{0.33\textwidth}
                \centering
                \includegraphics[width=.95\linewidth]{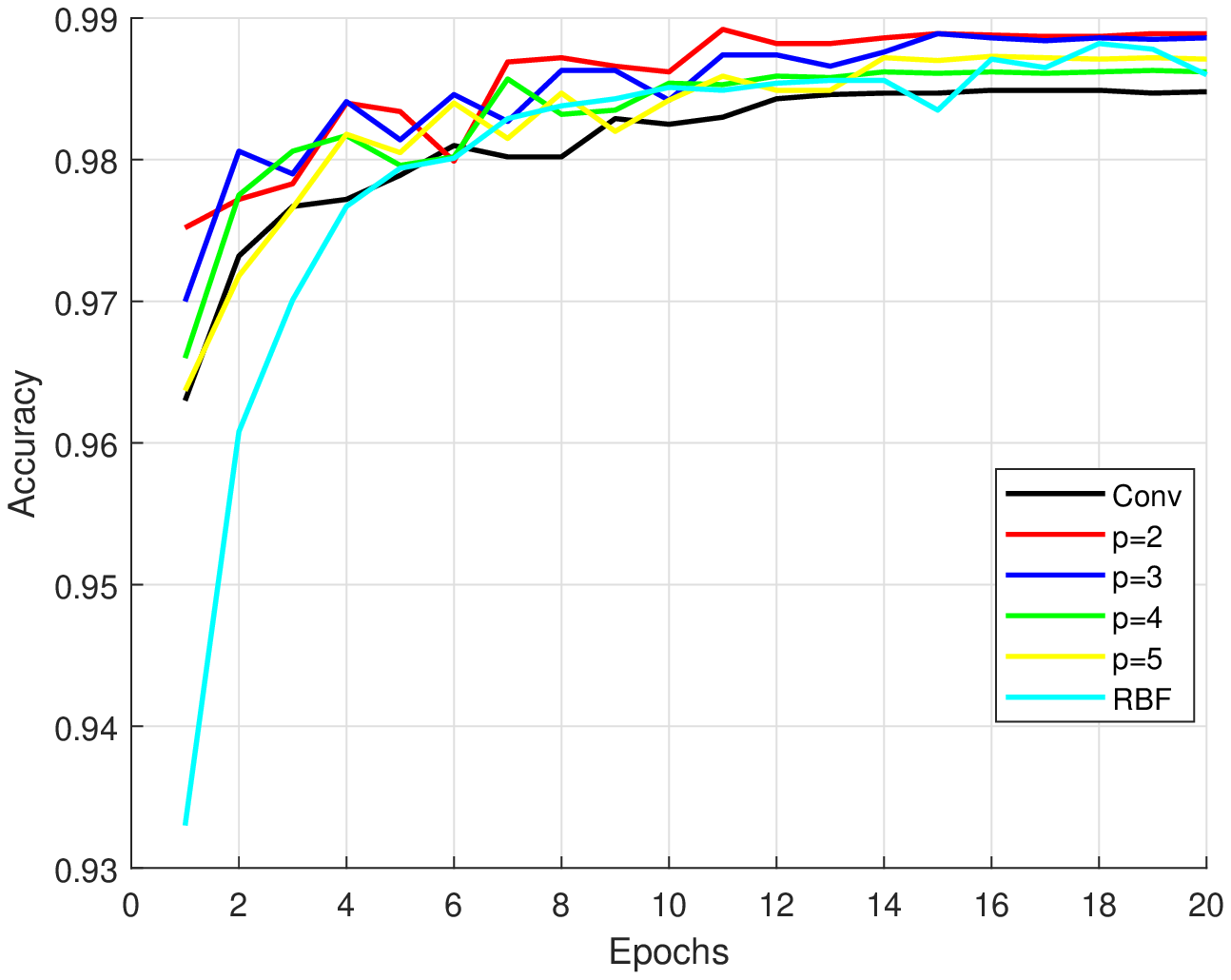}
                \caption{MNIST}
        \end{subfigure}%
        \begin{subfigure}[b]{0.33\textwidth}
                \centering
                \includegraphics[width=.95\linewidth]{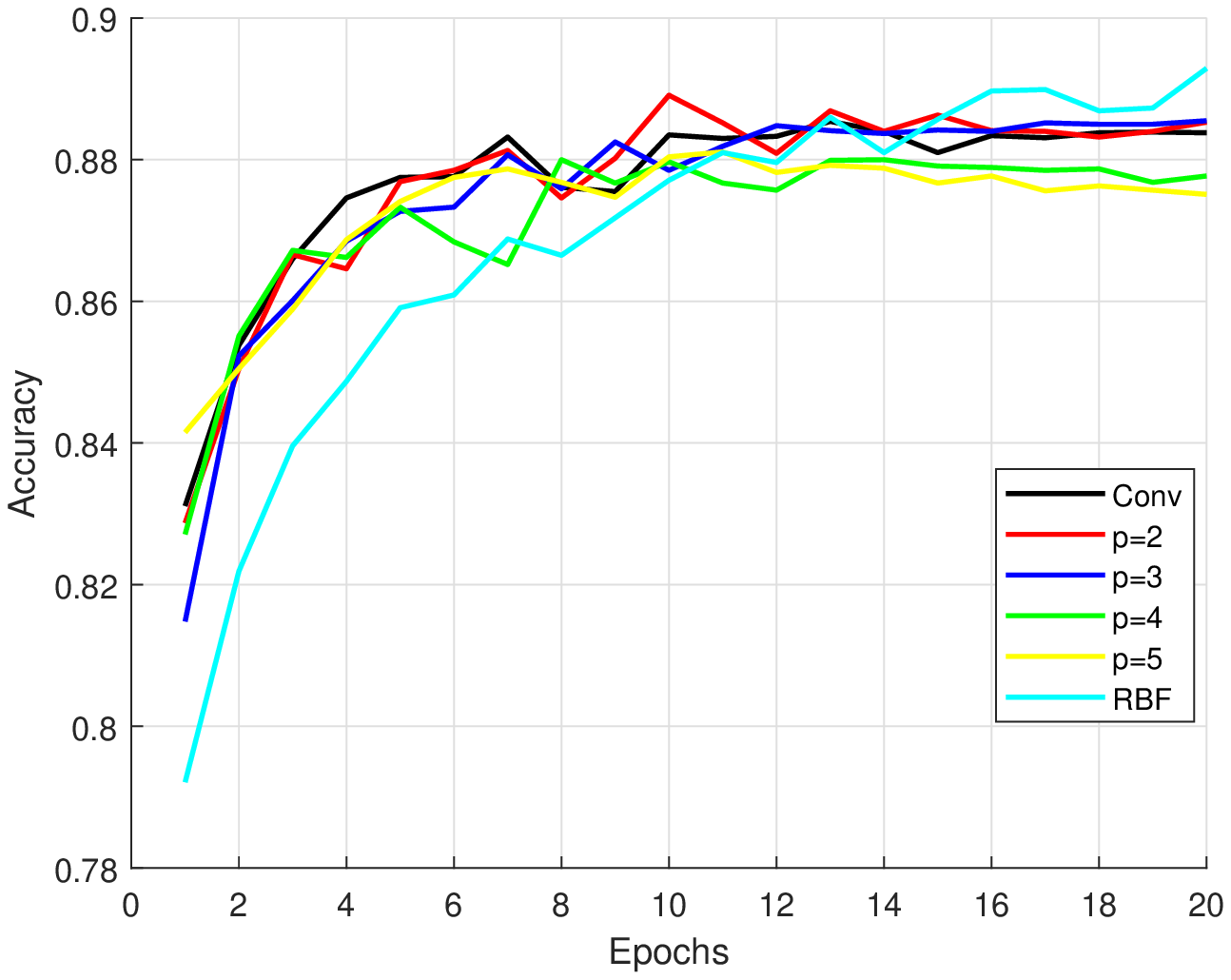}
                \caption{Fashion MNIST}
        \end{subfigure}%
        \begin{subfigure}[b]{0.33\textwidth}
                \centering
                \includegraphics[width=.95\linewidth]{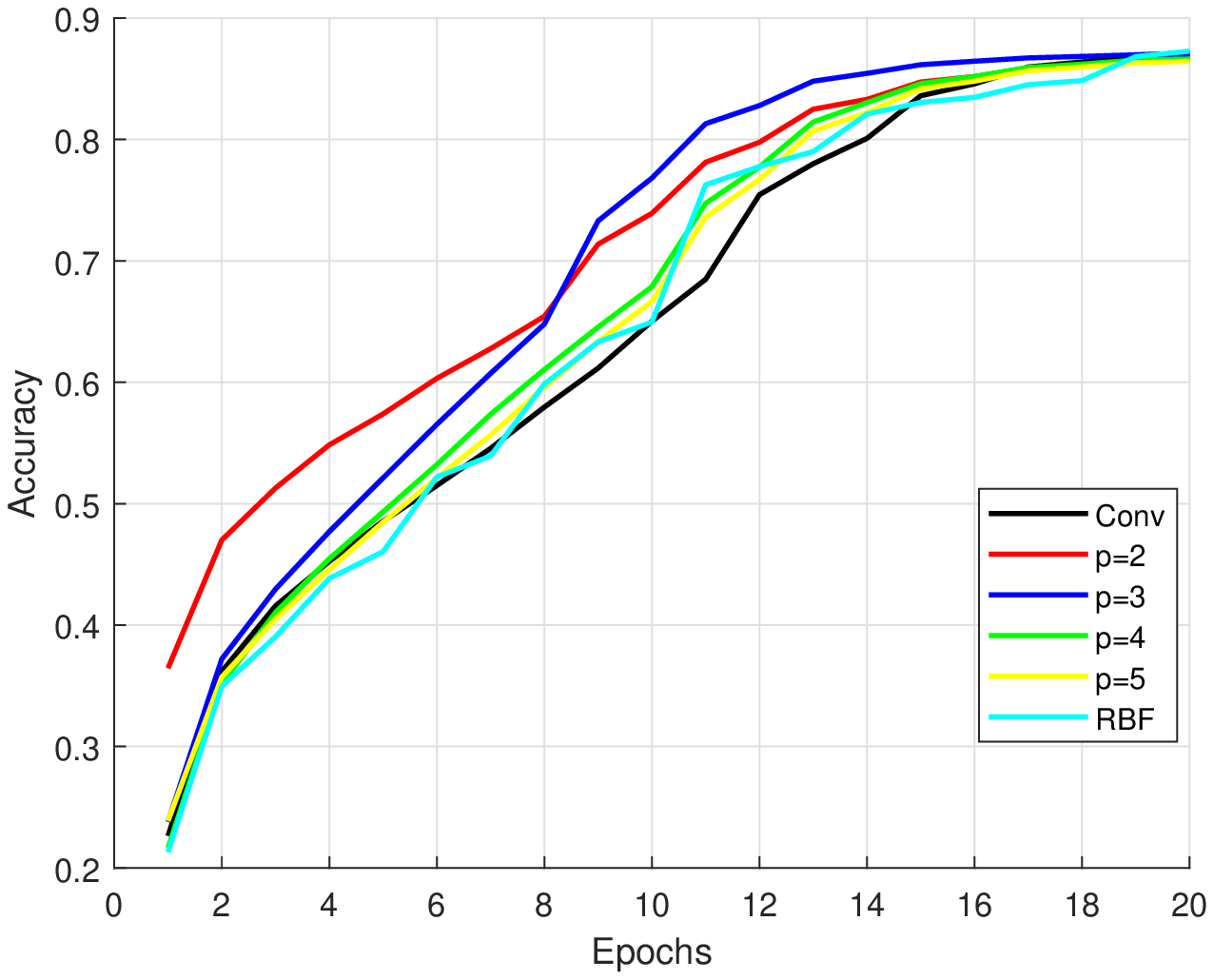}
                \caption{Cifar-10}
        \end{subfigure}
        \begin{subfigure}[b]{0.33\textwidth}
                \centering
                \includegraphics[width=.95\linewidth]{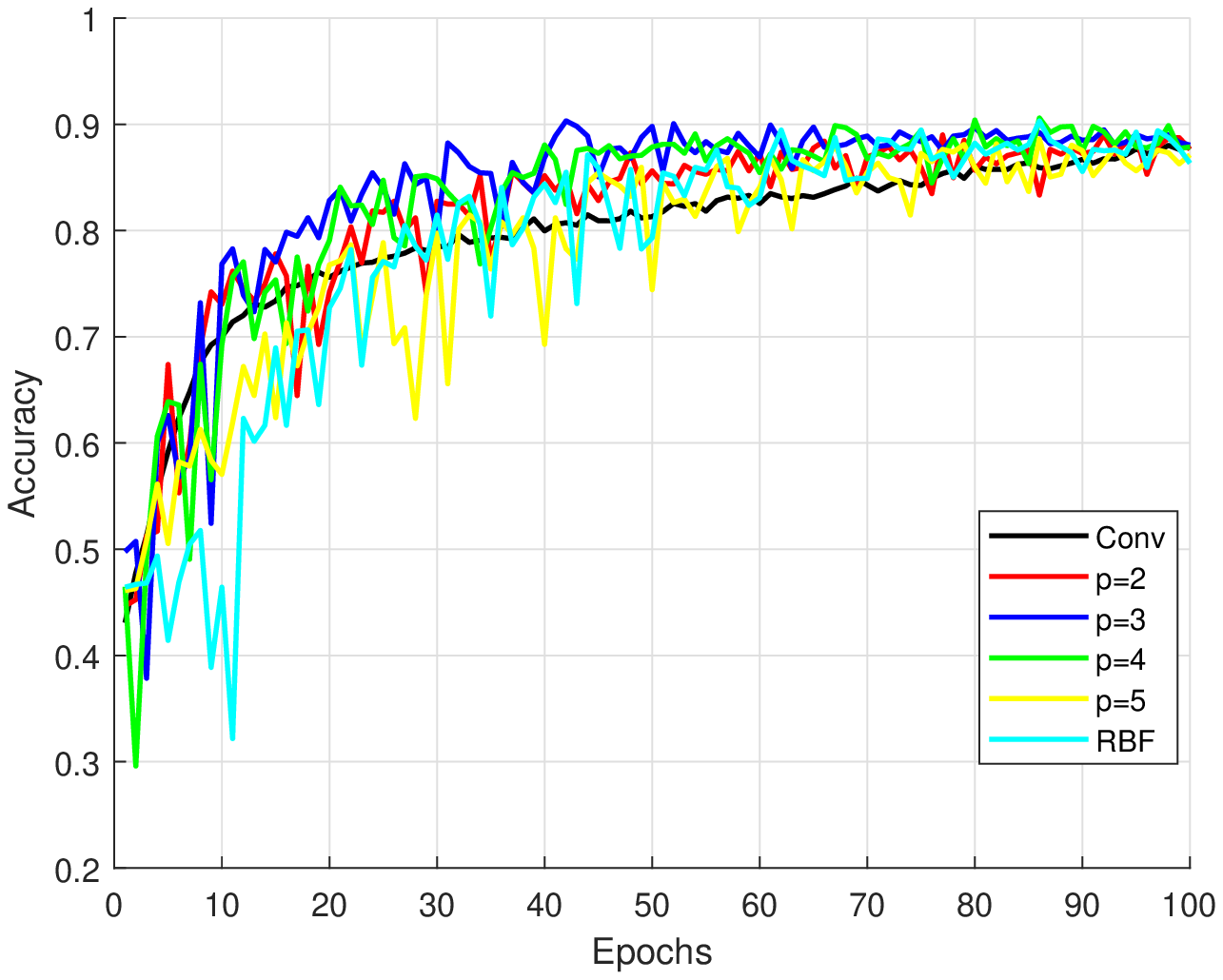}
                \caption{RAF-DB}
        \end{subfigure}%
        \begin{subfigure}[b]{0.33\textwidth}
                \centering
                \includegraphics[width=.95\linewidth]{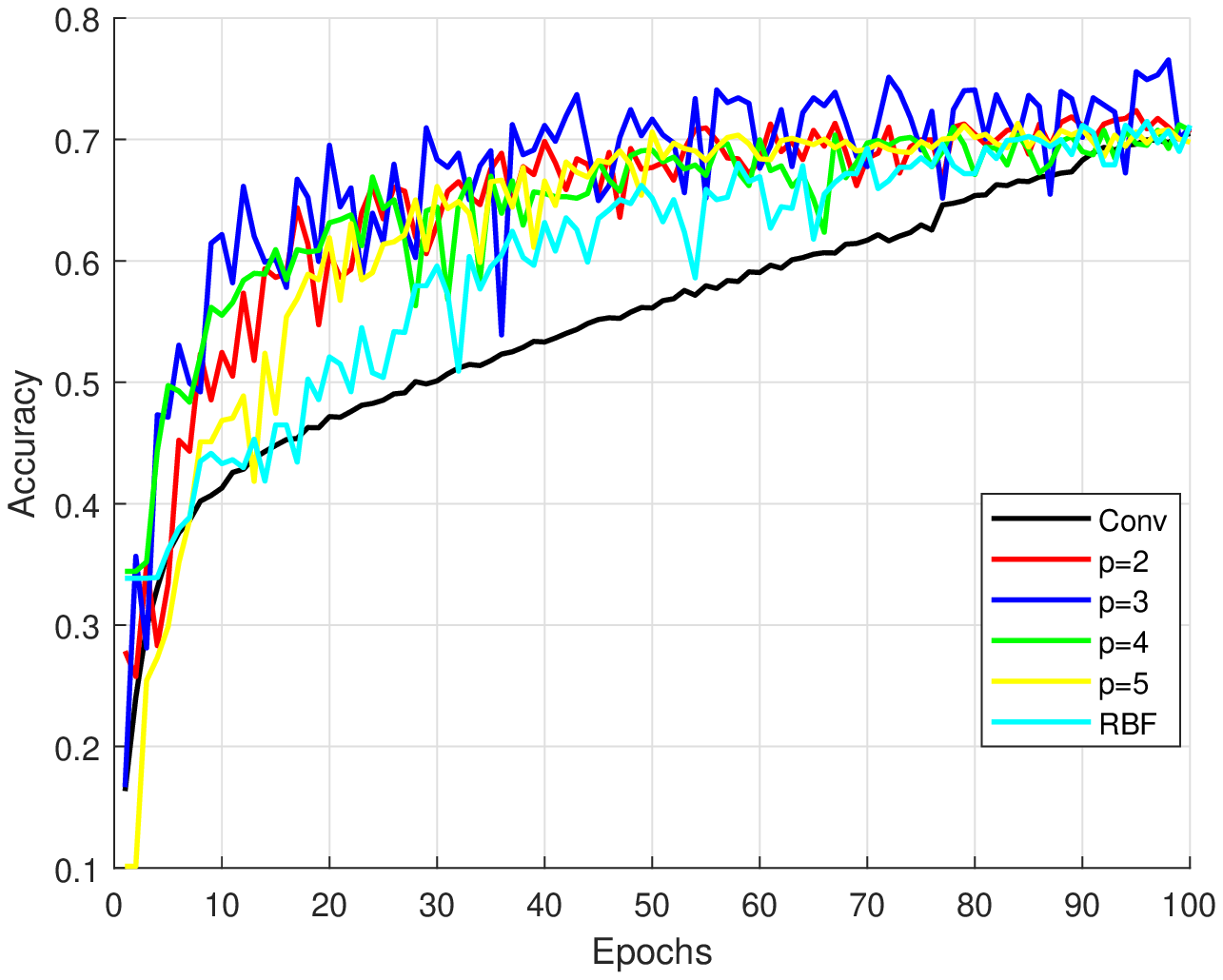}
                \caption{FER2013}
        \end{subfigure}%
        \begin{subfigure}[b]{0.33\textwidth}
                \centering
                \includegraphics[width=.95\linewidth]{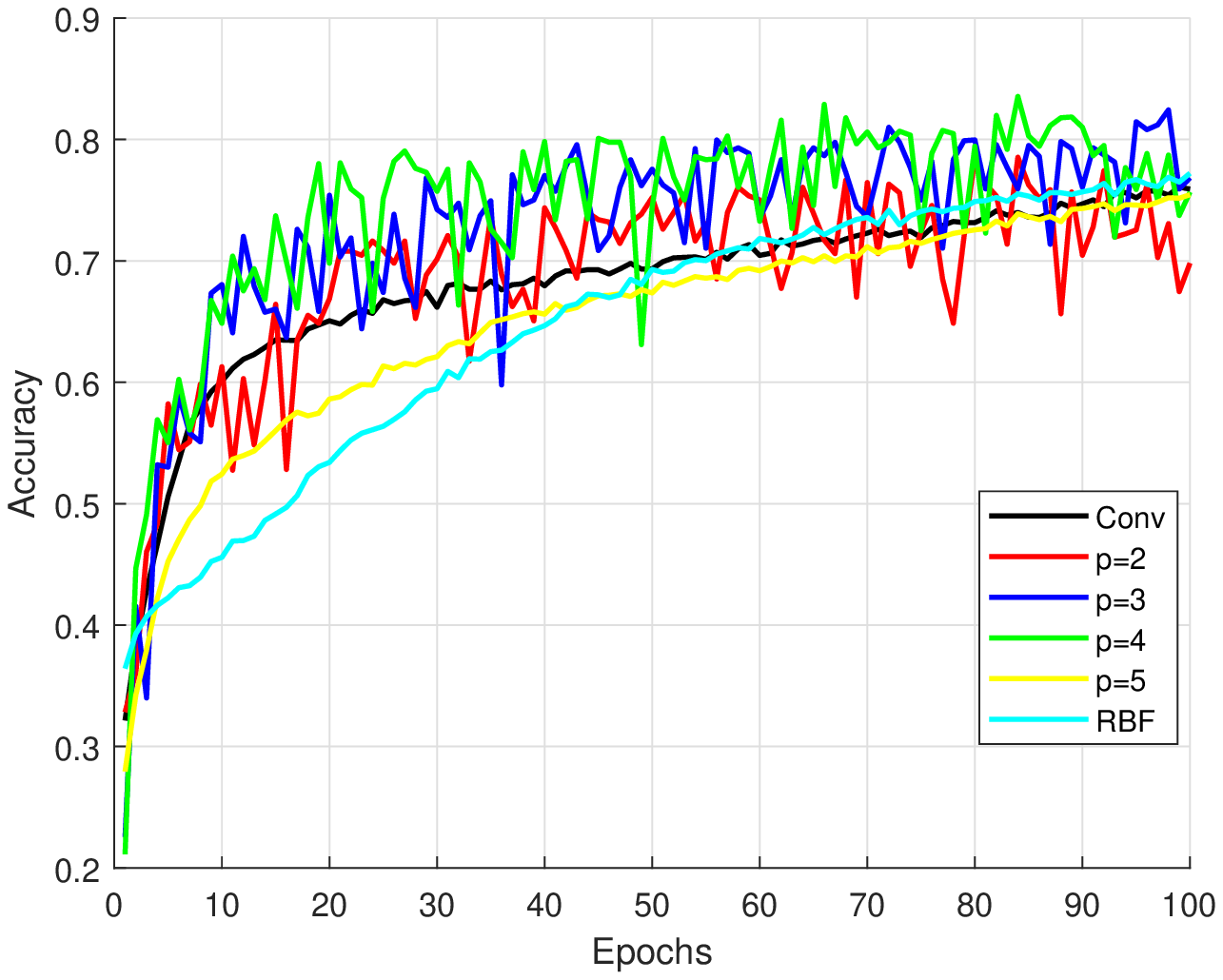}
                \caption{A ExpW}
        \end{subfigure}
        \caption{Convergence of networks with one kervolution layer at the end}
        \label{fig:kerv-last}
\end{figure*}

In this section we present the obtained results obtained with our two base models with  a full Kervolution configuration (Table \ref{tab:full-kerv}, Figure~\ref{fig:full-kerv}), a single Kervolution layer at the beginning of the network (Table \ref{tab:kerv-first}, Figure~\ref{fig:kerv-first}), and a single Kervolution layer at the end of the network (Table \ref{tab:kerv-last}, Figure~\ref{fig:kerv-last}).

\begin{table}[H]
\caption{Accuracy rates of full kervolution networks}
\label{tab:full-kerv}
\resizebox{\linewidth}{!}{
\begin{tabular}{lllllll}
\hline\noalign{\smallskip}
  & \multicolumn{2}{l}{Model-1}  & \multicolumn{4}{l}{Model-2}\\
\noalign{\smallskip}\hline\noalign{\smallskip}
Layers configuration  & MNIST & Fashion-MNIST  &  Cifar10 & RAF-DB & FER2013 & ExpW\\
\noalign{\smallskip}\hline\noalign{\smallskip}
Convolution (Linear kernel) &  98.50\%  &   88.29\% &  86.87\% & 87.05\% &70.49\% &75.91\% \\
2$^{nd}$-order Poly   & \textbf{99.05}\% &  \textbf{90.06} \%  &  87.92\% & \textbf{87.77}\% & \textbf{70.68}\% &\textbf{76.25}\% \\
3$^{rd}$-order Poly    & \textbf{99.16}\% & \textbf{90.04} \%  & \textbf{88.64}\% & \textbf{87.93}\% &\textbf{70.95}\% &\textbf{76.32}\%  \\
4$^{nd}$-order Poly   & \textbf{99.02}\% &   89.57\%  &  \textbf{88.41}\%  & \textbf{87.31}\% &\textbf{70.82}\% &\textbf{76.16}\% \\
5$^{rd}$-order Poly    & 98.99\% &   89.71\%  & 87.13\% & 86.04\% &69.65\% &75.80\% \\
Gaussian RBF $\sigma=0.9$    & 98.61\% &   88.78\%  & 87.51\% & \textbf{87.33}\% &\textbf{70.78}\% & \textbf{76.23}\% \\
\noalign{\smallskip}\hline
\end{tabular}
}
\end{table}


Table~\ref{tab:full-kerv} illustrates the accuracy rates obtained using our two base models with a full kervolution configuration. In table~\ref{tab:full-kerv} one can clearly notice that the use of kervolution layers enhance the overall accuracy of the network, especially with second, third and fourth order polynomial kernels. In comparison to full convolution network, the full kervolution network enhances the accuracy of Model-1 on MNIST dataset by 0.65\% and 1.77\% on Fashion-MNIST. On the other hand, the accuracy of Model-2 increases by 1.73\% on CIFAR-10, 0.88\% on RAF-DB, 0.46\% on FER2013 and 0.41\% on ExpW. The best accuracy rates are, in most cases, reached with third order polynomial kernel and decreases with higher order polynomial kernels. We have also noticed that Gaussian RBF kernel is more beneficial on fine-grained FER dataset than other datasets. Therefore, we can deduce that Gaussian RBF kernels are more sensitive to subtle details.

Figure~\ref{fig:full-kerv} shows the validation accuracy rates evolution of our base models with full convolution and kervolution configurations. We can clearly notice that kervolution networks, especially second and third order polynomial kernels, converge in fewer epochs than convolution network. On the other hand, Gaussian RBF kernel takes more time to converge than the other kernels. We have also noticed that, for fine-grained FER datasets, the accuracy curve fluctuates accordingly to the kernel degree. This can be explained by the fact that kernels are more sensitive to subtle changes in input data. Therefore, any small change in the weights during the training phase can cause a big change in the final classification decision.

\begin{table}[H]
\begin{center}

\caption{Accuracy rates with a single kervolution layer at the begining}
\label{tab:kerv-first}
\resizebox{\linewidth}{!}{
\begin{tabular}{lllllll}
\hline\noalign{\smallskip}
  & \multicolumn{2}{l}{Model-1}  & \multicolumn{4}{l}{Model-2}\\
\noalign{\smallskip}\hline\noalign{\smallskip}
Layers configuration  & MNIST & Fashion-MNIST  &  Cifar10 & RAF-DB & FER2013 & ExpW\\
\noalign{\smallskip}\hline\noalign{\smallskip}

Convolution (Linear kernel)&98.50\% & 88.29\% &  86.87\% & 87.05\% &70.49\% &75.91\% \\
2$^{nd}$-order Poly   & 98.87\% &   89.74\%  &  87.81\% & 87.62\% &70.82\% &76.64\% \\
3$^{rd}$-order Poly    & \textbf{99.04}\% &  \textbf{90.26}\%  & \textbf{88.73}\% & \textbf{88.06}\% &\textbf{71.06}\% &\textbf{76.85}\%  \\
4$^{nd}$-order Poly   & \textbf{99.04}\% &   \textbf{90.61}\%  &  \textbf{88.52}\%  & \textbf{87.88}\% &70.88\% &\textbf{76.53}\% \\
5$^{rd}$-order Poly    & 98.99\% &   \textbf{90.35}\%  & 86.48\% & 86.71\% &69.89\% &75.56\% \\
Gaussian RBF $\sigma=0.9$    & 98.74\% &   89.53\%  & \textbf{88.48}\% & \textbf{87.89}\% &\textbf{70.98}\% &\textbf{76.75}\% \\
\noalign{\smallskip}\hline
\end{tabular}
}
\end{center}
\end{table}

Table~\ref{tab:kerv-first} presents the accuracy rate results of our two base models, in which we replaced the first convolution layer by a kervolution layer. In table~\ref{tab:kerv-first} we can also notice that the use of one kervolution layer at the beginning of the network increases the accuracy of the network comparing to full convolution network. Furthermore, the use of only one kervolution layer in the beginning of the network allows to reach better results than full-kervolution network. Compared to full convolution network, the accuracy rate increases up to 1.97\% on Fashion-MNIST,  1.86\% on Cifar-10, 2\% on RAF-DB, 0.57\% on FER2013 and 0.94 on ExpW. Gaussian RBF kernels are slightly less accurate than third and fourth polynomial kernels, yet they also increase the accuracy rate compared to full-convolution network. They enhance the accuracy on Cifar-10 by 1.61\%, 0.84\% on REF-DB, 0.49\% on FER2013 and 0.84\% on ExpW. Another remark is that higher order kernels are more accurate than lower order kernel when used only at the beginning of the network. This may be explained by the fact that higher order kernels can fit to subtle details more efficiently than lower order kernels. Therefore, the network learns to detect more useful information than full-convolution network and will be less prone to over-fitting than full-kervolution network.

Figure~\ref{fig:kerv-first} shows the accuracy convergence of our two base models, in which we replaced the first convolution layer by a kervolution layer. As illustrated in figure~\ref{fig:kerv-first}, using one kervolution layer at the beginning of the network allows the latter to converge in less time than the full convolution network. Similarly to full-kervolution configuration, second and third degree polynomial are the fastest to converge in all cases. Higher degree polynomial layers are also faster to converge than full convolution configuration except for FER datasets. On the other hand, Gaussian RBF layers are the slowest to converge, though they surpass convolution on MNIST, Fashion MNIST and Cifar-10.



\begin{table}[H]
\caption{Accuracy rates with a single kervolution layer at the end}
\label{tab:kerv-last}
\resizebox{\linewidth}{!}{
\begin{tabular}{lllllll}
\hline\noalign{\smallskip}
  & \multicolumn{2}{l}{Model-1}  & \multicolumn{4}{l}{Model-2}\\
\noalign{\smallskip}\hline\noalign{\smallskip}
Layers configuration  & MNIST & Fashion-MNIST  &  Cifar10 & RAF-DB & FER2013 & ExpW\\
\noalign{\smallskip}\hline\noalign{\smallskip}

Convolution (Linear kernel) &  98.50\%  &   88.29\% &  86.87\% & 87.05\% &70.49\% &75.91\% \\
2$^{nd}$-order Poly   & 98.89\% &   88.37\%  &  87.11\% & 87.63\% &70.81\% &76.24\% \\
3$^{rd}$-order Poly    & 98.84\% &  88.45\%  & 87.23\% & 88.03\% & 70.95\% &76.82\%  \\
4$^{nd}$-order Poly   & 98.62\% &   87.76\%  &  87.09\%  & 87.85\% &70.75\% &76.66\% \\
5$^{rd}$-order Poly    & 98.72\% &   87.45\%  & 86.97\% & 87.15\% &69.76\% &75.45\% \\
Gaussian RBF $\sigma=0.9$    & \textbf{98.75}\% &   \textbf{89.33}\%  & \textbf{90.13}\% & \textbf{89.36}\% &\textbf{71.15}\% &\textbf{77.21}\% \\
\noalign{\smallskip}\hline
\end{tabular}
}
\end{table}

Table~\ref{tab:kerv-last} presents the accuracy rate results of our two base models, in which we replaced the last convolution layer by a kervolution layer. In table~\ref{tab:kerv-last} one can notice that the use of polynomial kervolution layers at the end of the network decreases its accuracy compared to full-kervolution configuration and kervolution at the beginning. Yet, it performs better than full-convolution configuration. Indeed, we could surpass full-convolution configuration by 0.34\% on MNIST, 0.16\% on Fashion MNIST, 0.36\% on Cifar-10, .0.98\% on RAF-DB, 0.46\% on FER2013 and 0.91\% on ExpW. The best accuracy rates where reached with third degree polynomial kernel. Other polynomial kernels also perform slightly better than full-convolution configuration, but are less accurate than the other kervolution configurations. On the other hand, Gaussian RBF kernels increase remarkably the overall accuracy of the network compared to full-convolution configuration and other Gaussian RBF kervolution configurations. Indeed, it increases the overall accuracy rates by 0.25\% on MNIST, 1.04\% on Fashion MNIST, 3.26\% on Cifar-10, 2.31\% on RAF-DB, 0.66\% on FER2013 and 1.30\% on ExpW. With this has been said, we deduce that polynomial kervolution layers are more suited for feature extraction which explains why they are more accurate when plugged at the beginning of the network. On the other hand, Gaussian RBF kervolution layers are more beneficial when plugged at the end of the network than the beginning. This can be explained by the fact that Gaussian RBF kernels are more accurate for classification than for feature extraction.


Figure~\ref{fig:kerv-last} shows the accuracy convergence of our two base models, in which we replaced the last convolution layer by a kervolution layer. We can clearly notice that the use of one kervolution layer at the end of the network quicken remarkably its convergence. This impact on convergence speed is valid with all kernels. On the other hand, even-though Gaussian RBF kernel reaches the highest accuracy rates among all kernels, it is still the slowest kernel to converge. Polynomial kernels, on the other hand, are still the fastest in convergence.

According to these results, we can say that the use of kervolution layers have clearly a beneficial impact on the accuracy rate and convergence speed of the network. As we have seen above, this positive impact is noticeable wherever kervolution layers are used. In terms of accuracy, using one kervolution layer at the beginning of the network shown to be the most efficient configuration. Whereas, in terms of convergence speed, using one kervolution layer at the end of the network shown to be the fastest to converge. Full-kervolution configuration is, for its part, average in both accuracy and convergence speed compared with the above configurations. In addition, full-kervolution configuration is more prone to overfitting than the other kervolution configurations.

\subsubsection{Learnable weights pooling}

\begin{figure*}[h]
        \begin{subfigure}[b]{0.33\textwidth}
                \centering
                \includegraphics[width=.95\linewidth]{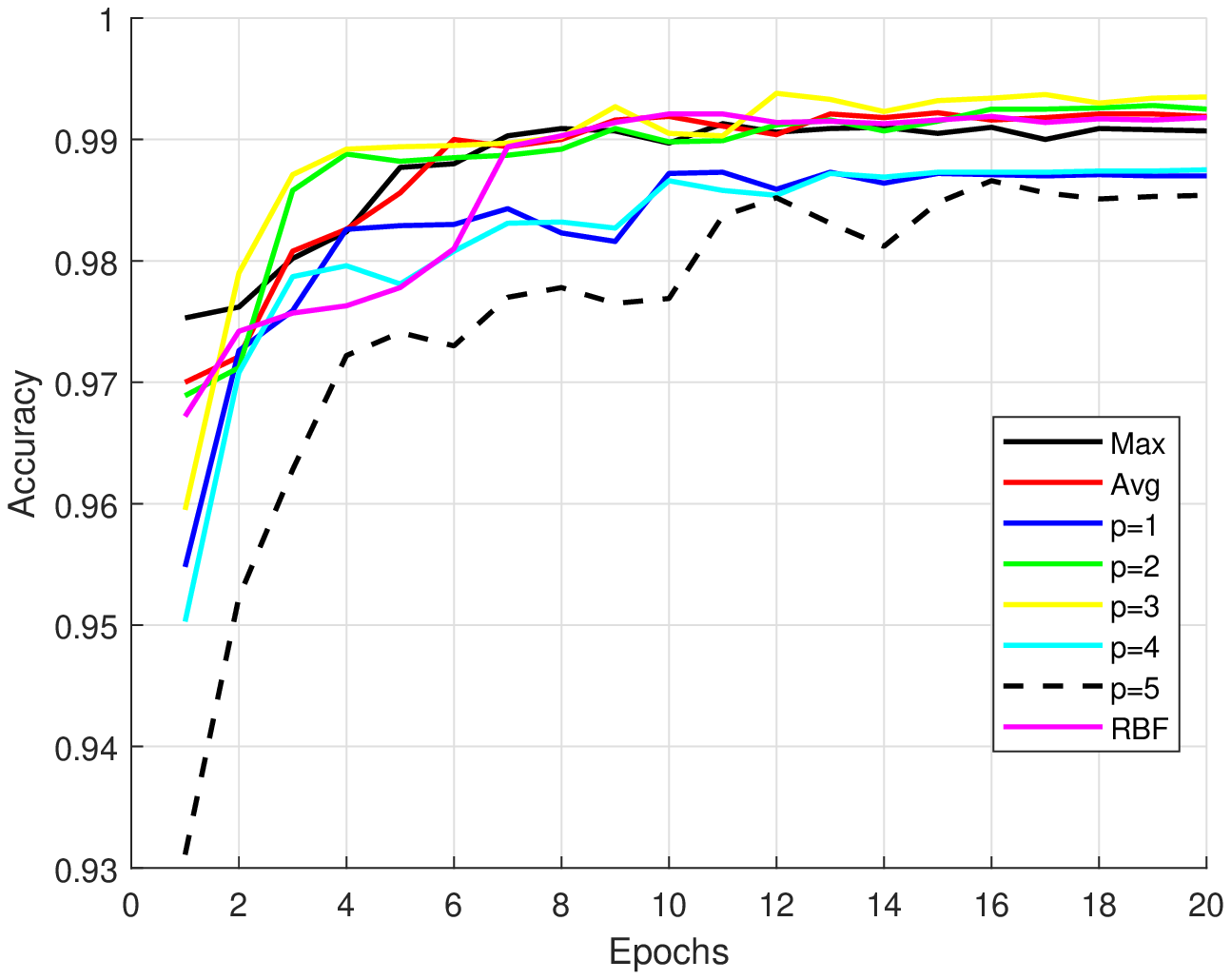}
                \caption{MNIST}
        \end{subfigure}%
        \begin{subfigure}[b]{0.33\textwidth}
                \centering
                \includegraphics[width=.95\linewidth]{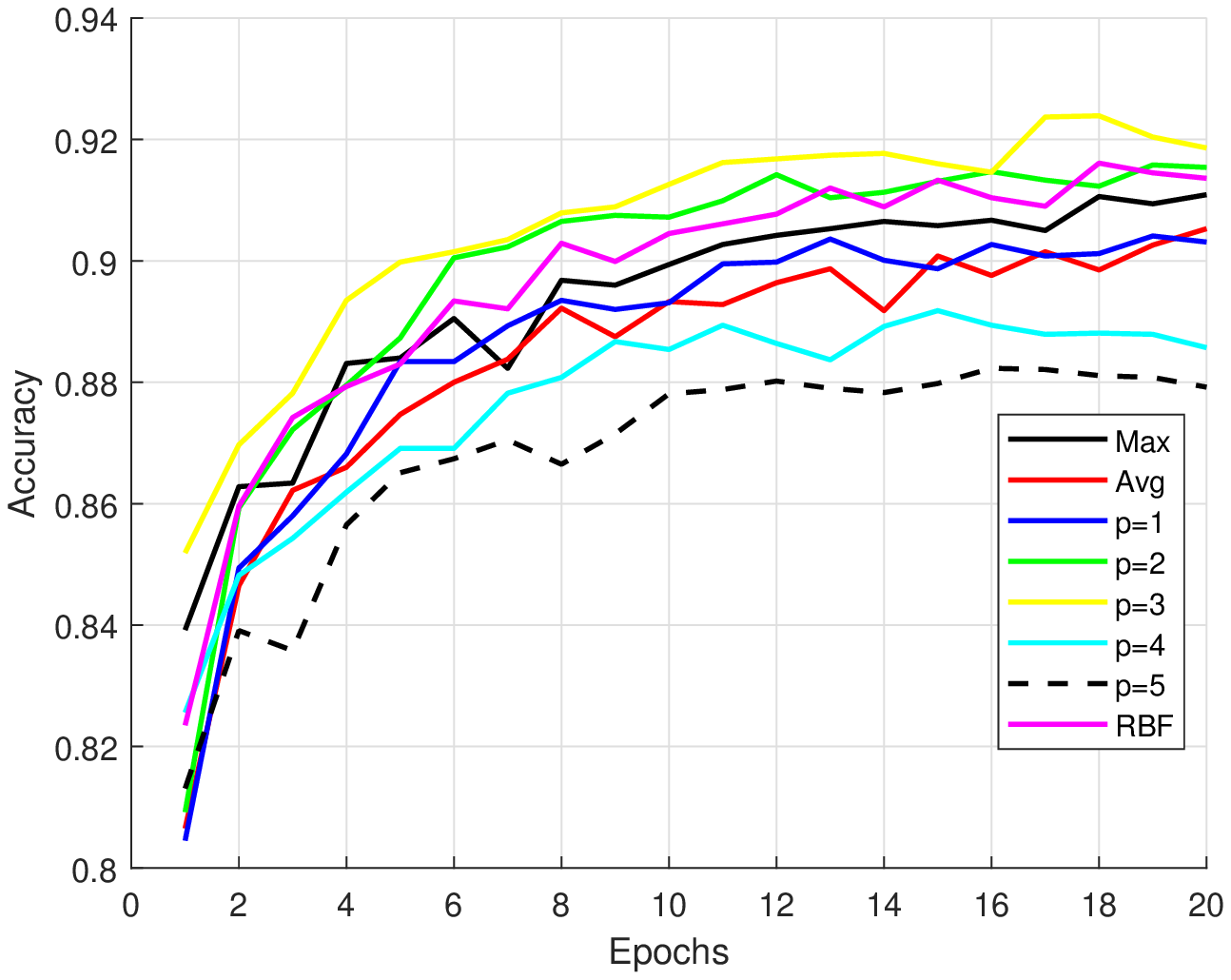}
                \caption{Fashion MNIST}
        \end{subfigure}%
        \begin{subfigure}[b]{0.33\textwidth}
                \centering
                \includegraphics[width=.95\linewidth]{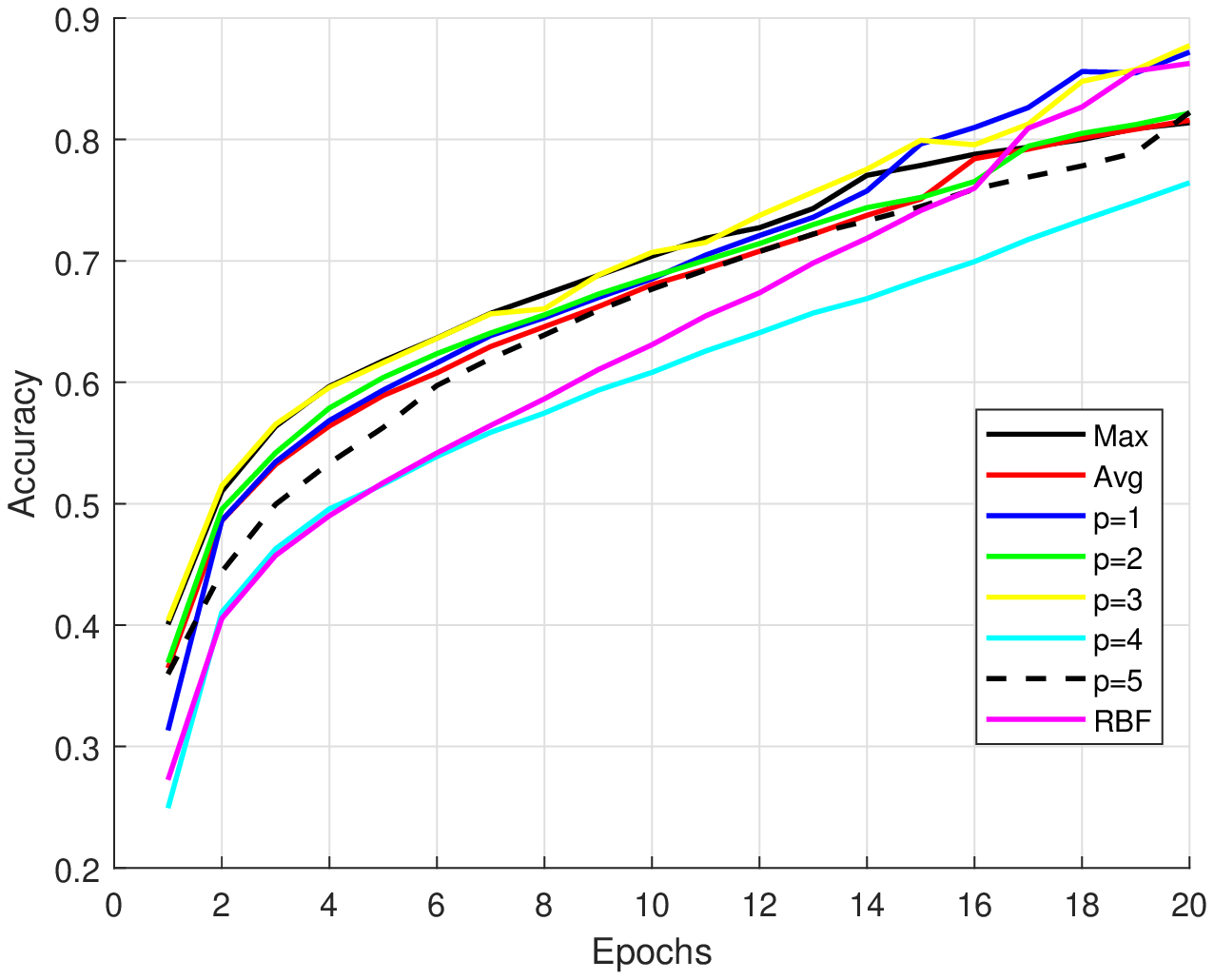}
                \caption{FER20132}
        \end{subfigure}
        \begin{subfigure}[b]{0.33\textwidth}
                \centering
                \includegraphics[width=.95\linewidth]{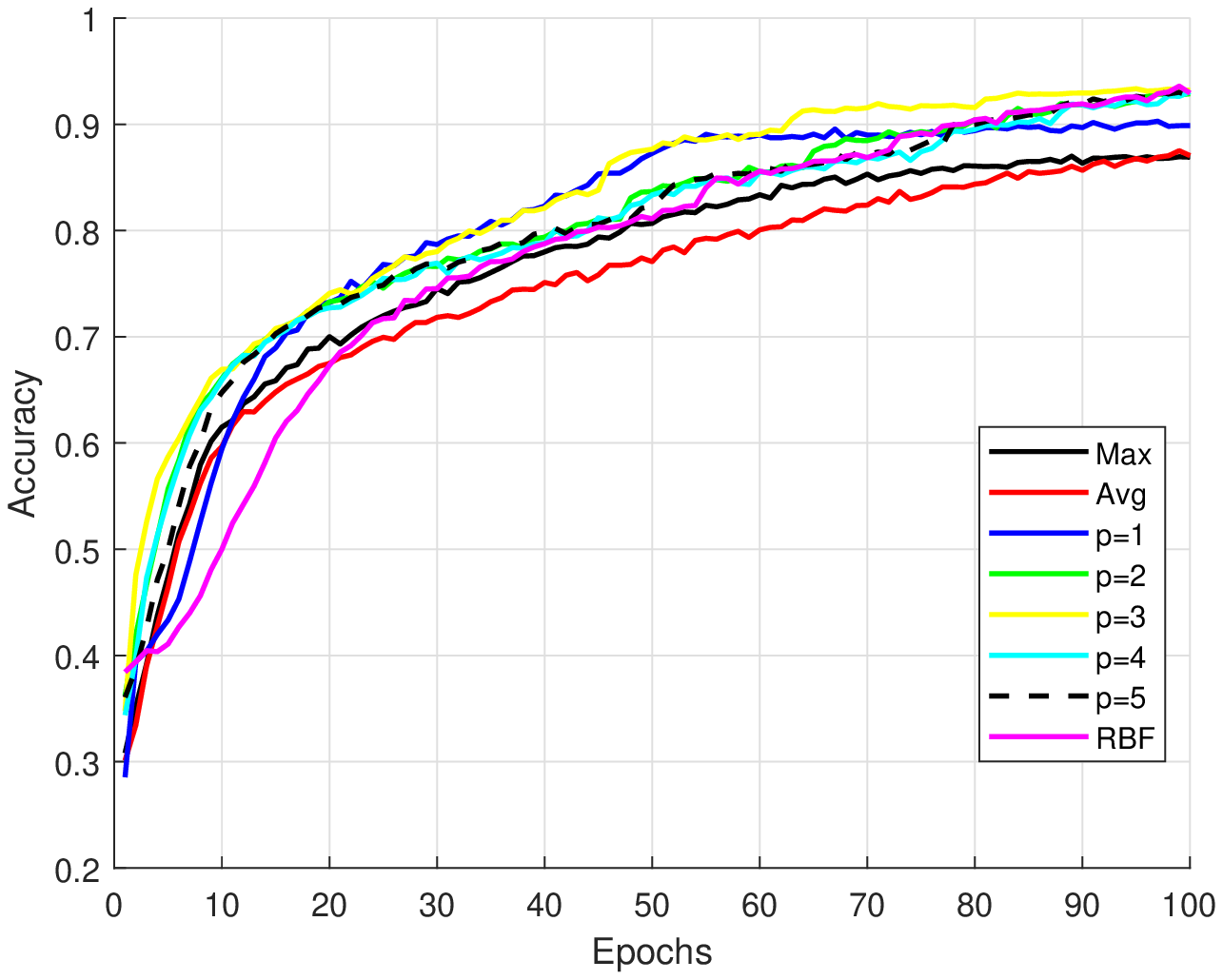}
                \caption{RAF-DB}
        \end{subfigure}%
        \begin{subfigure}[b]{0.33\textwidth}
                \centering
                \includegraphics[width=.95\linewidth]{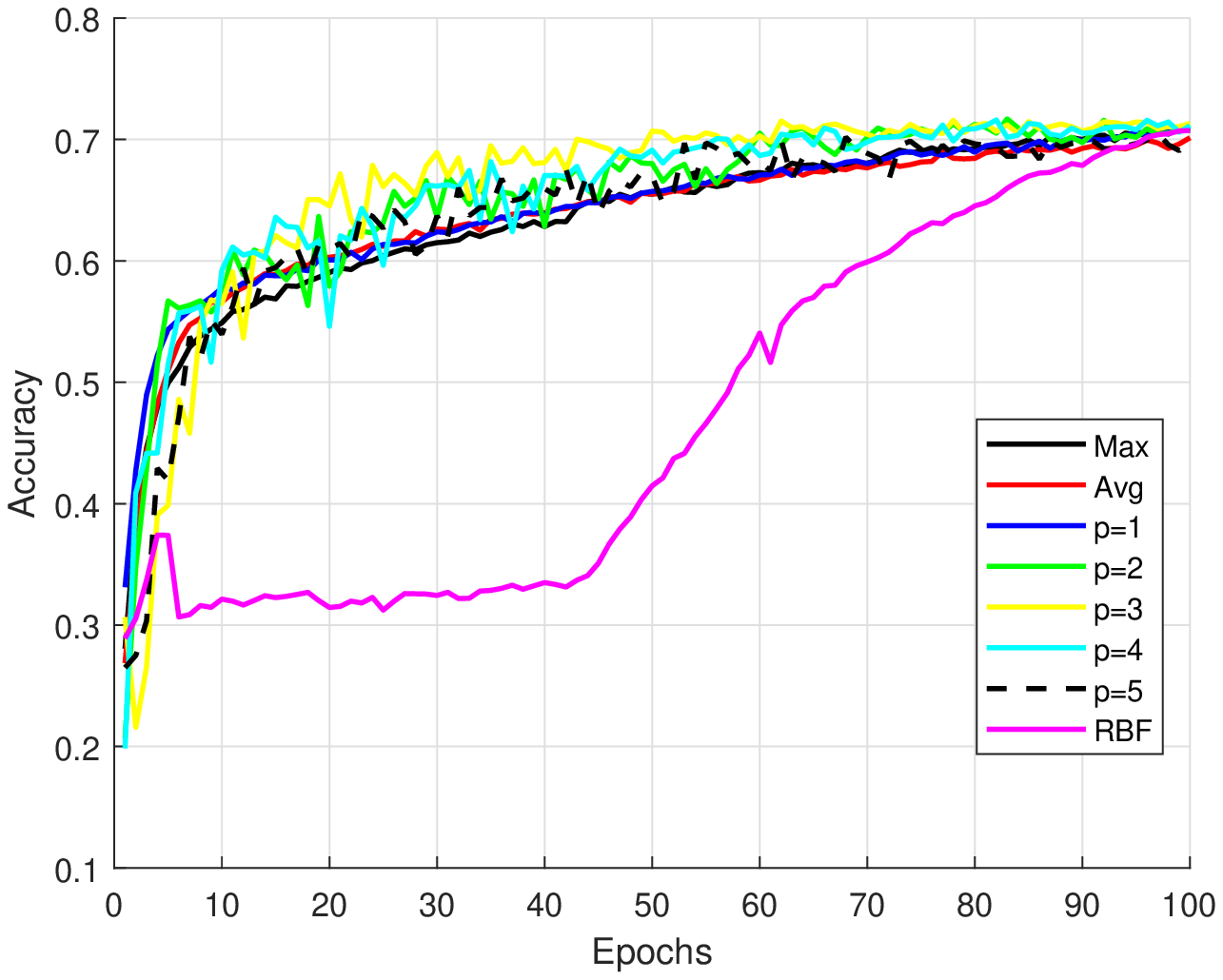}
                \caption{FER2013}
        \end{subfigure}%
        \begin{subfigure}[b]{0.33\textwidth}
                \centering
                \includegraphics[width=.95\linewidth]{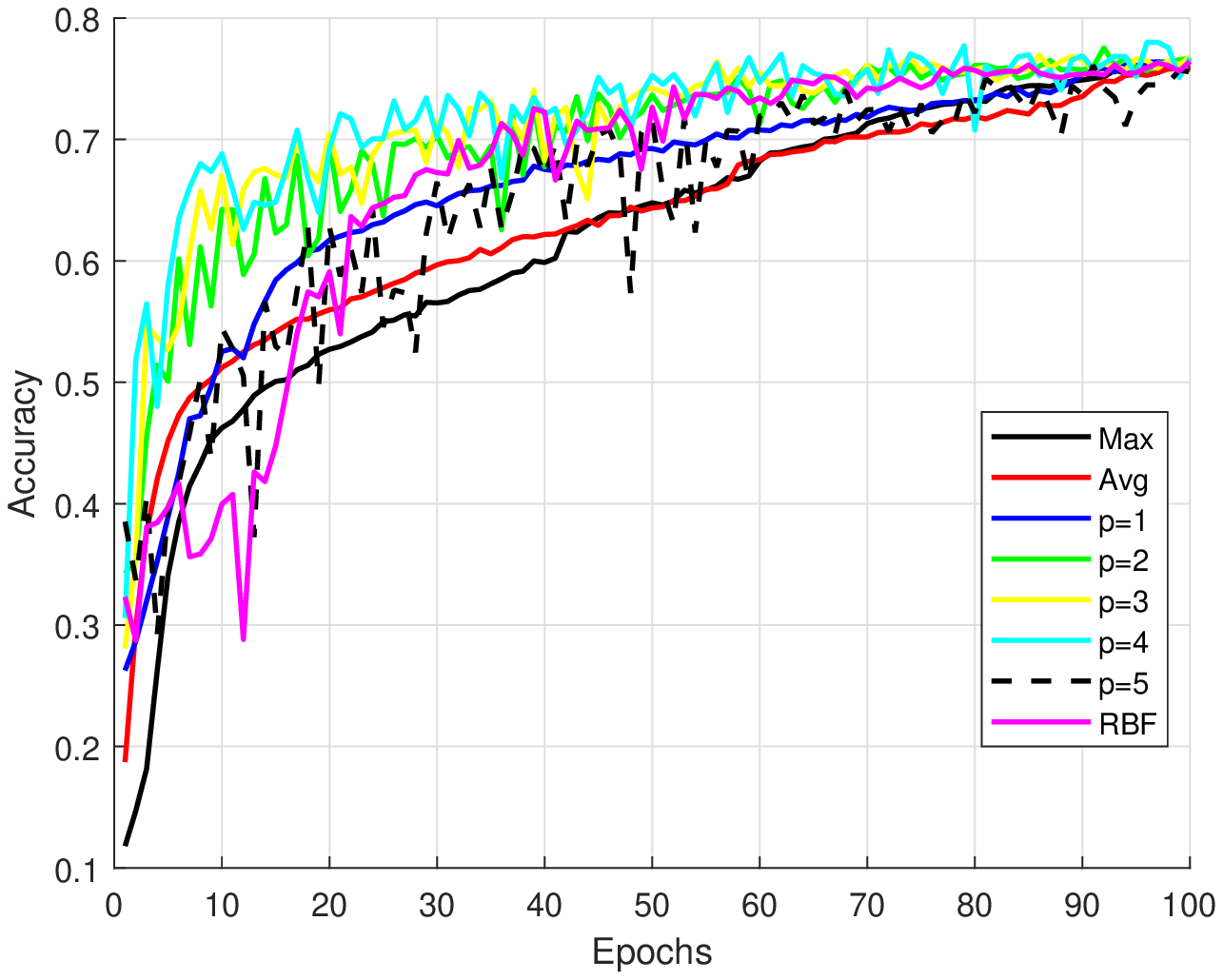}
                \caption{A ExpW}
        \end{subfigure}
        \caption{Convergence of networks with learnable pooling}
        \label{fig:full-learnable pooling}
\end{figure*}

\begin{figure*}[h]
        \begin{subfigure}[b]{0.33\textwidth}
                \centering
                \includegraphics[width=.95\linewidth]{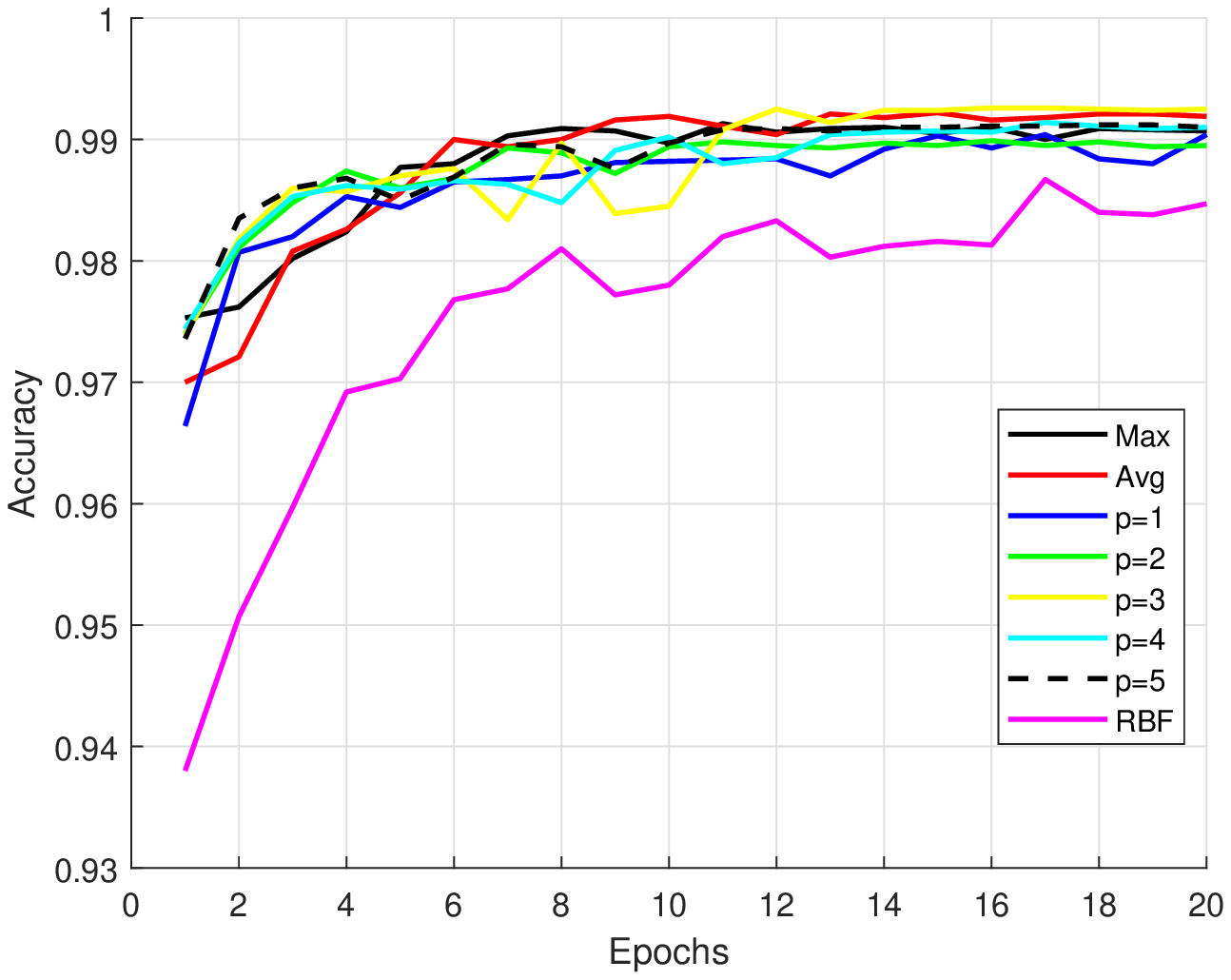}
                \caption{MNIST}
        \end{subfigure}%
        \begin{subfigure}[b]{0.33\textwidth}
                \centering
                \includegraphics[width=.95\linewidth]{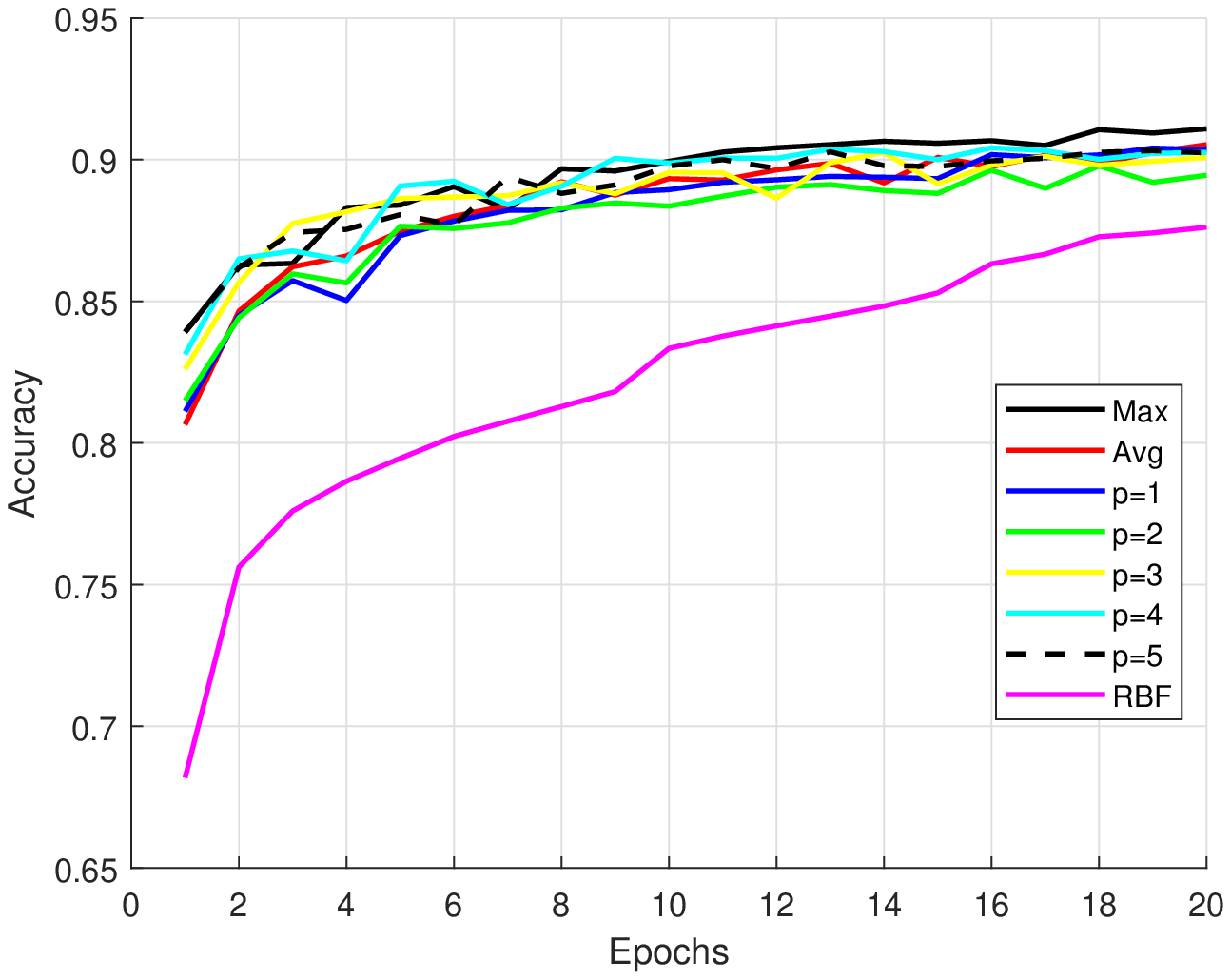}
                \caption{Fashion MNIST}
        \end{subfigure}%
        \begin{subfigure}[b]{0.33\textwidth}
                \centering
                \includegraphics[width=.95\linewidth]{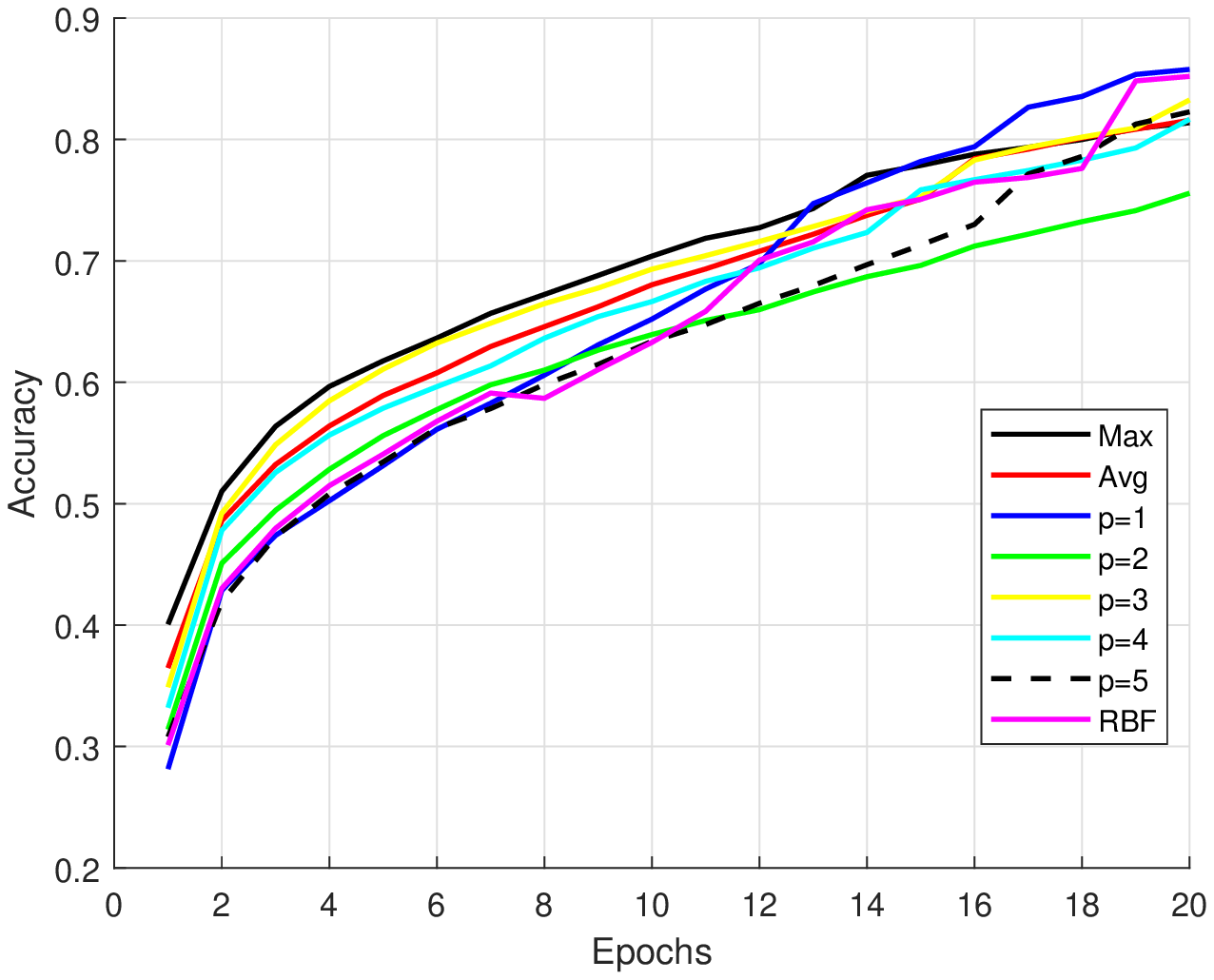}
                \caption{Cifar-10}
        \end{subfigure}
        \begin{subfigure}[b]{0.33\textwidth}
                \centering
                \includegraphics[width=.95\linewidth]{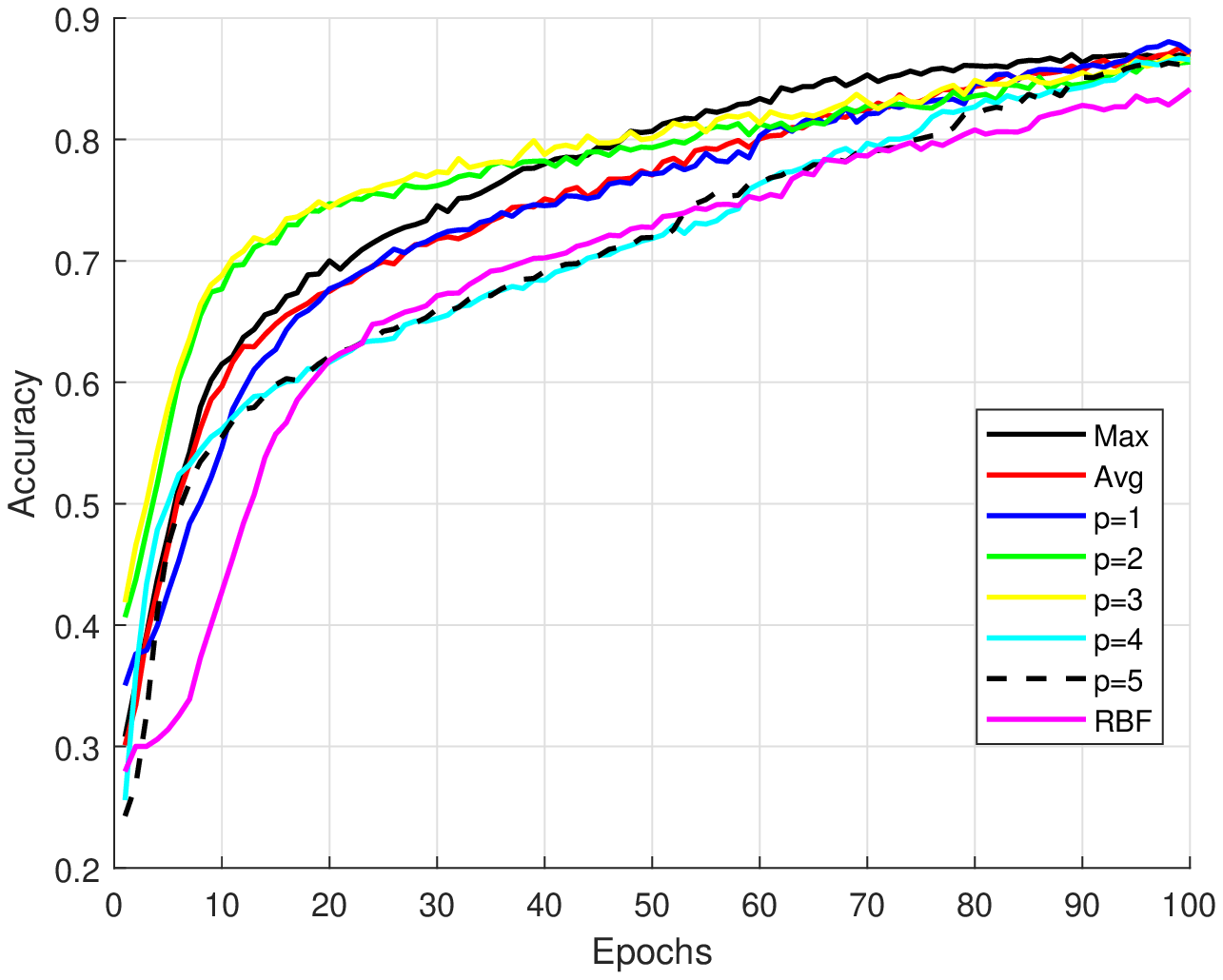}
                \caption{RAF-DB}
        \end{subfigure}%
        \begin{subfigure}[b]{0.33\textwidth}
                \centering
                \includegraphics[width=.95\linewidth]{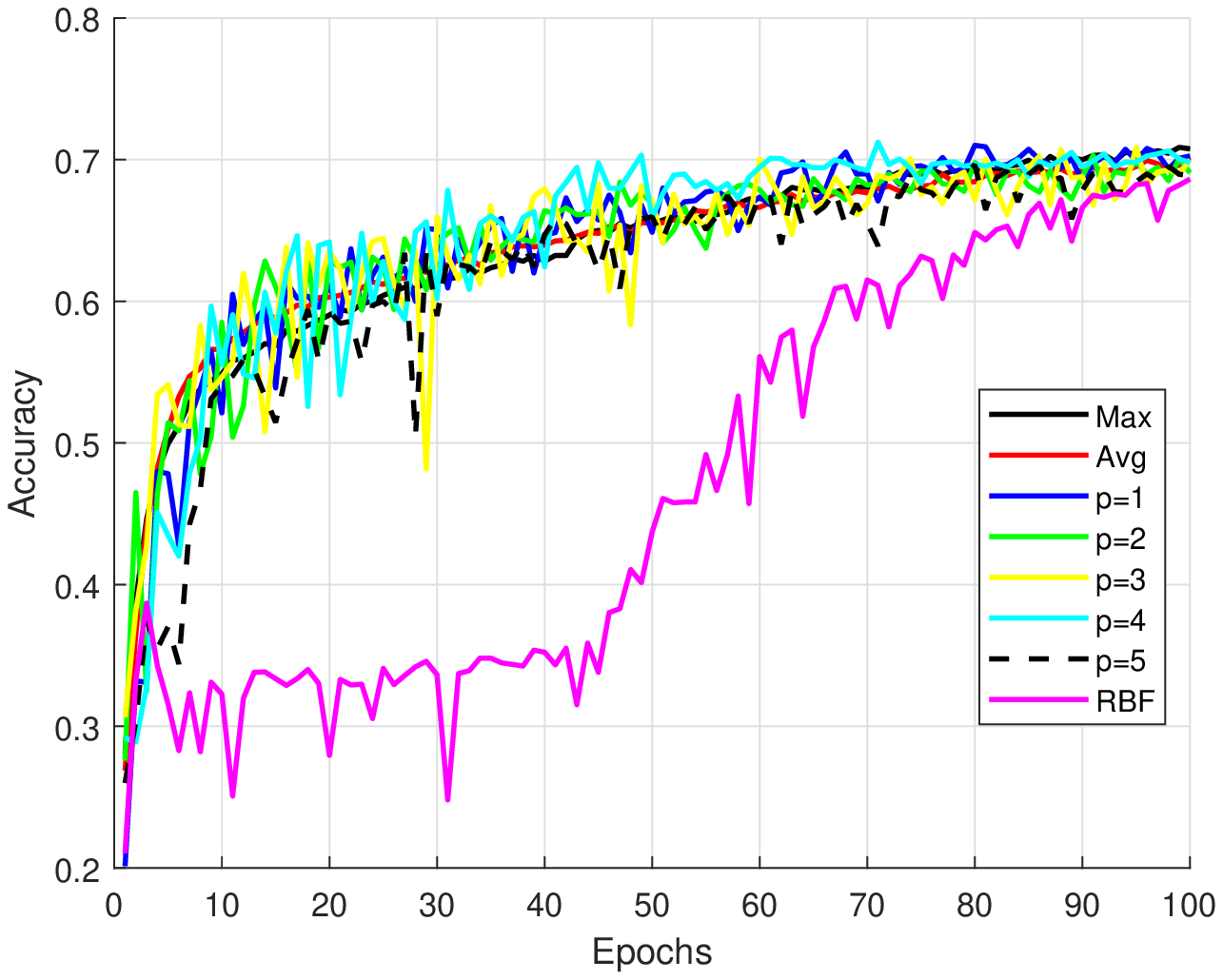}
                \caption{FER2013}
        \end{subfigure}%
        \begin{subfigure}[b]{0.33\textwidth}
                \centering
                \includegraphics[width=.95\linewidth]{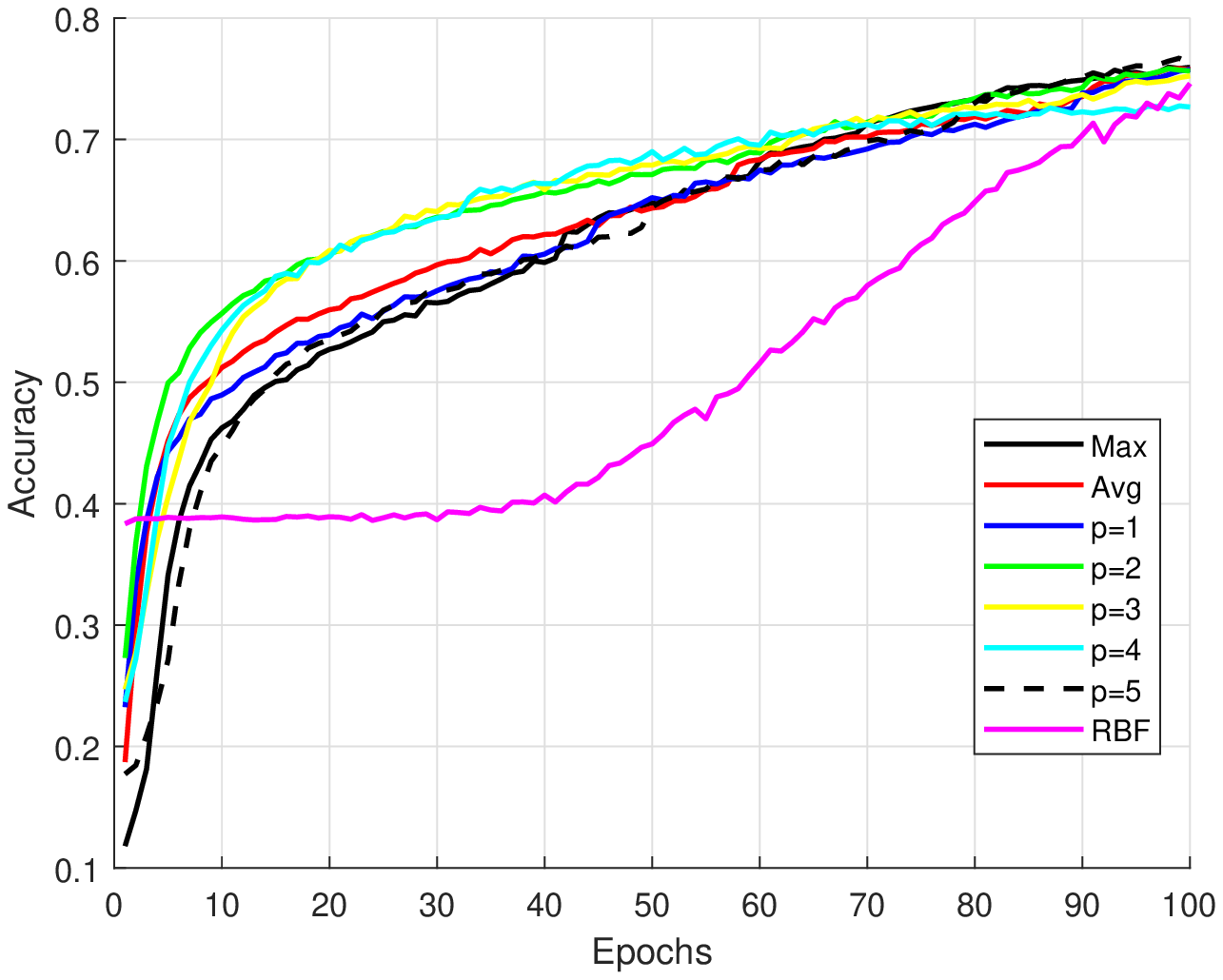}
                \caption{A ExpW}
        \end{subfigure}
        \caption{Networks convergence with learnable pooling at the beginning}
        \label{fig:learnablepoolingfirst}
\end{figure*}

\begin{figure*}
        \begin{subfigure}[b]{0.33\textwidth}
                \centering
                \includegraphics[width=.95\linewidth]{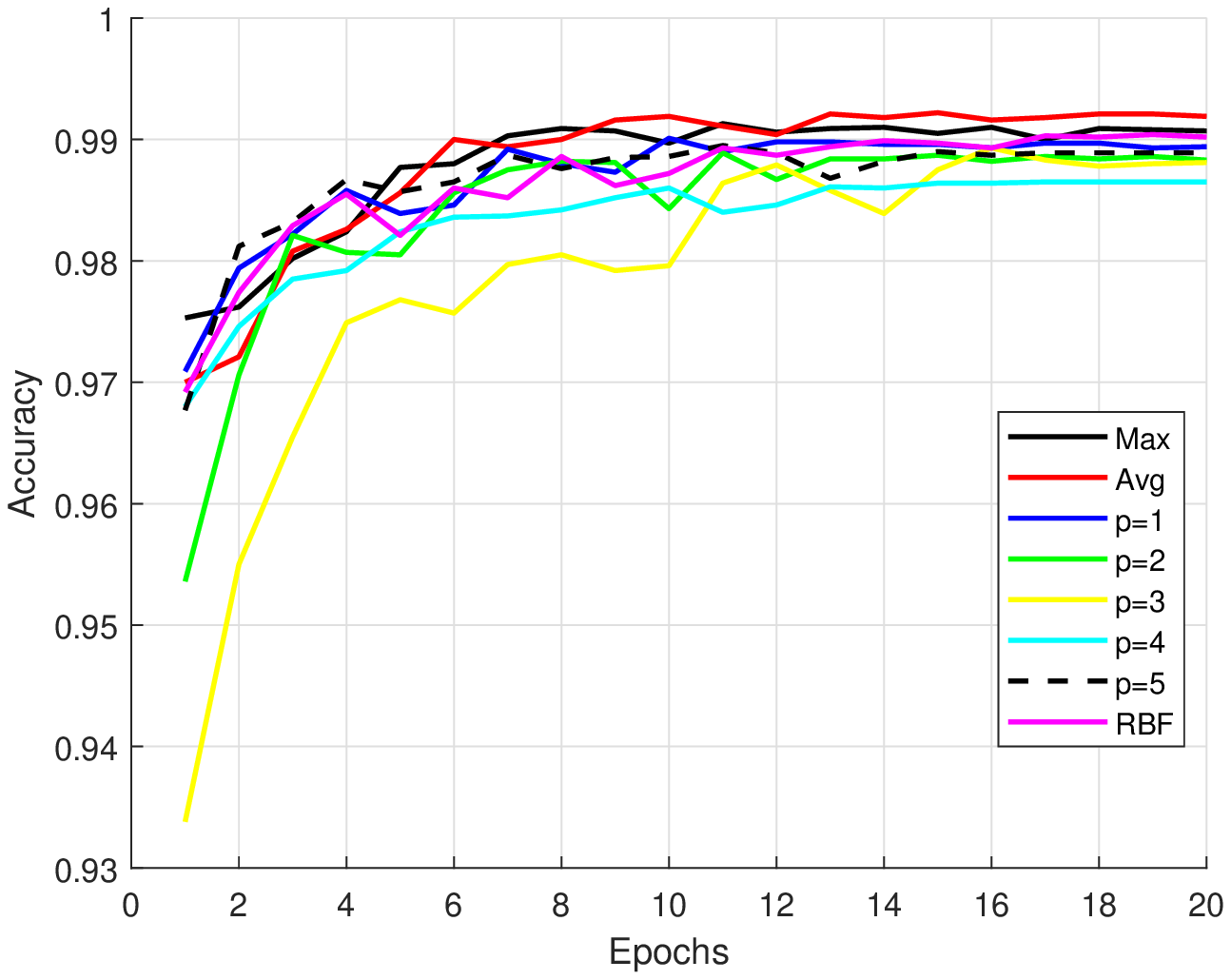}
                \caption{MNIST}
        \end{subfigure}%
        \begin{subfigure}[b]{0.33\textwidth}
                \centering
                \includegraphics[width=.95\linewidth]{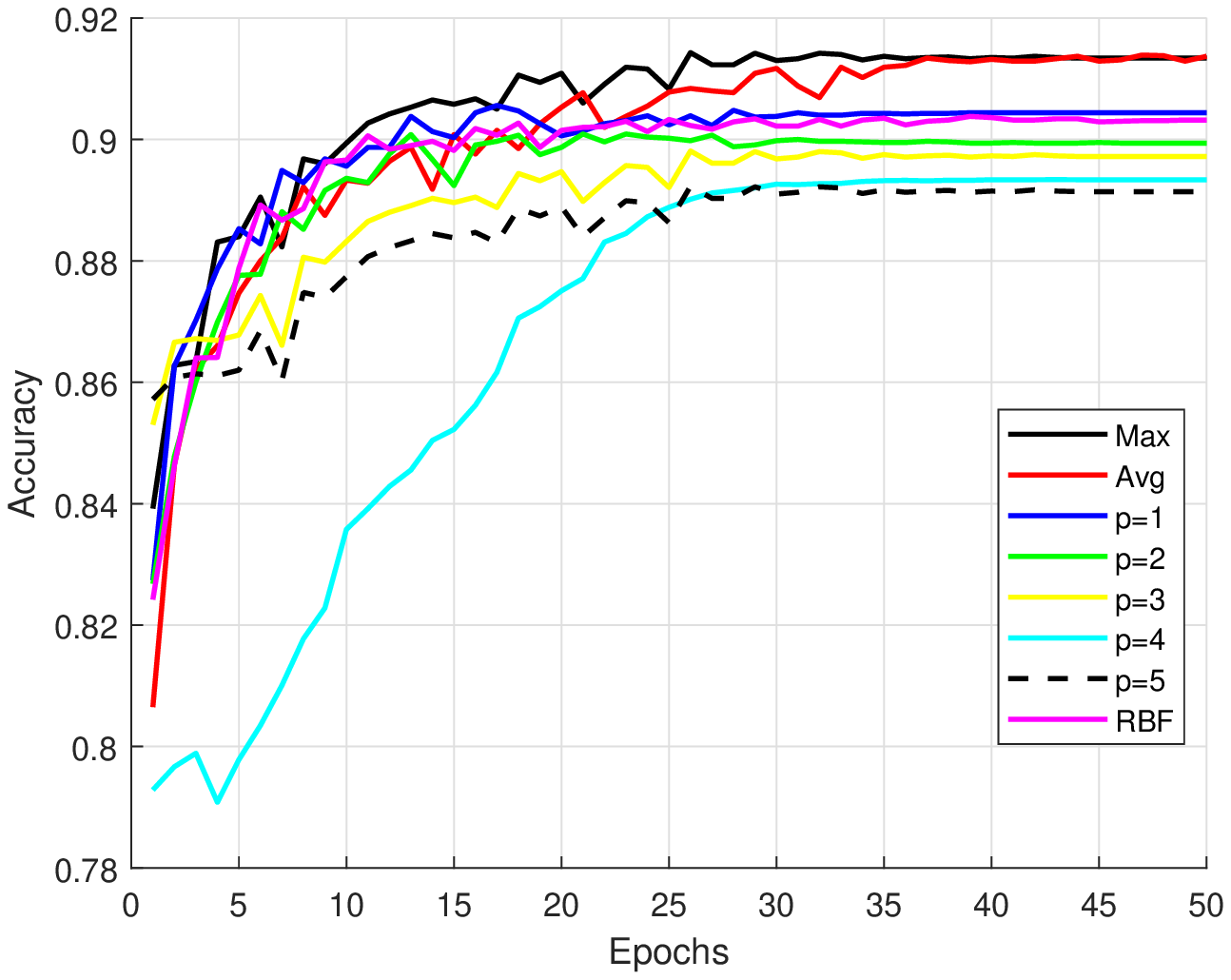}
                \caption{Fashion MNIST}
        \end{subfigure}%
        \begin{subfigure}[b]{0.33\textwidth}
                \centering
                \includegraphics[width=.95\linewidth]{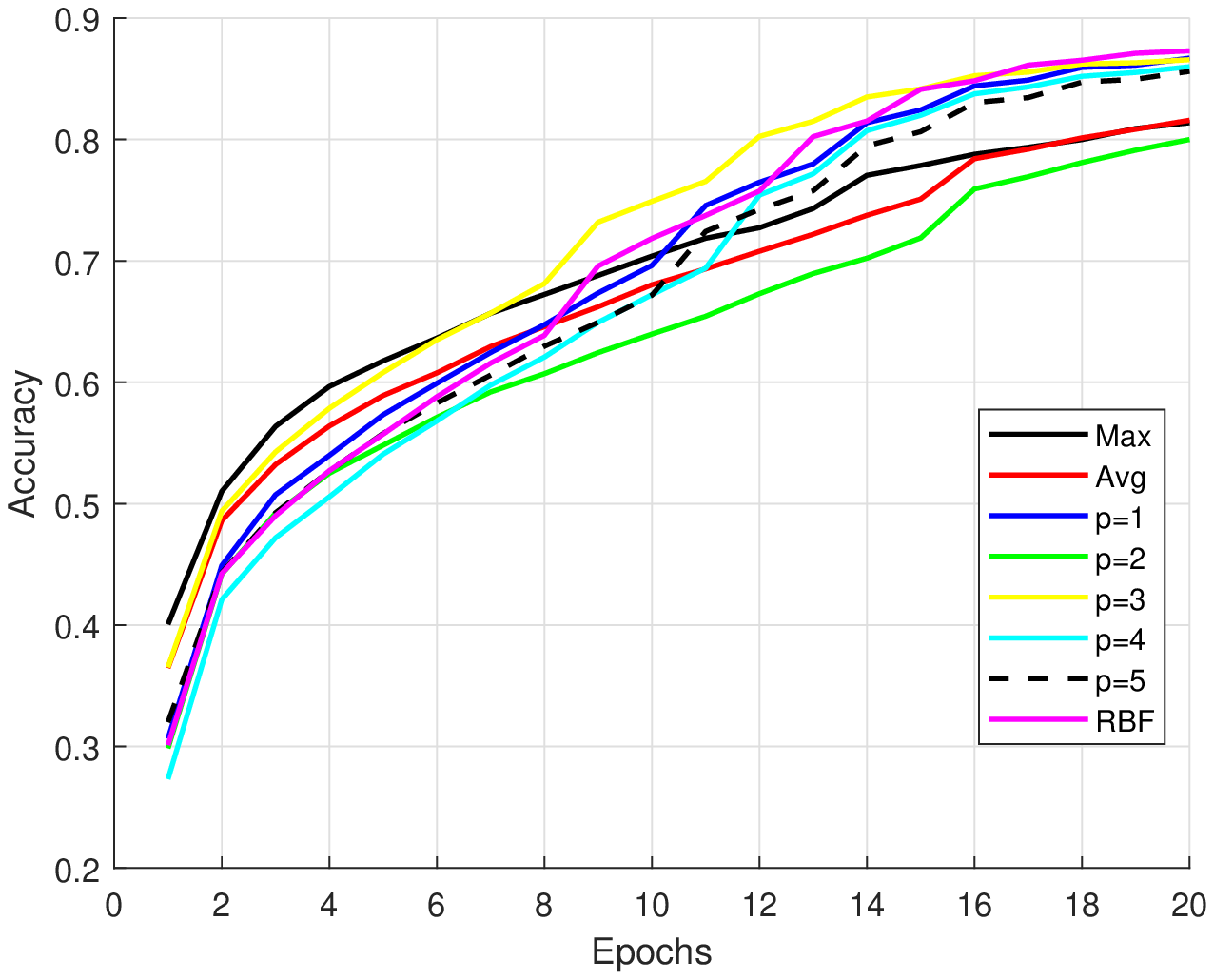}
                \caption{Cifar-10}
        \end{subfigure}
        \begin{subfigure}[b]{0.33\textwidth}
                \centering
                \includegraphics[width=.95\linewidth]{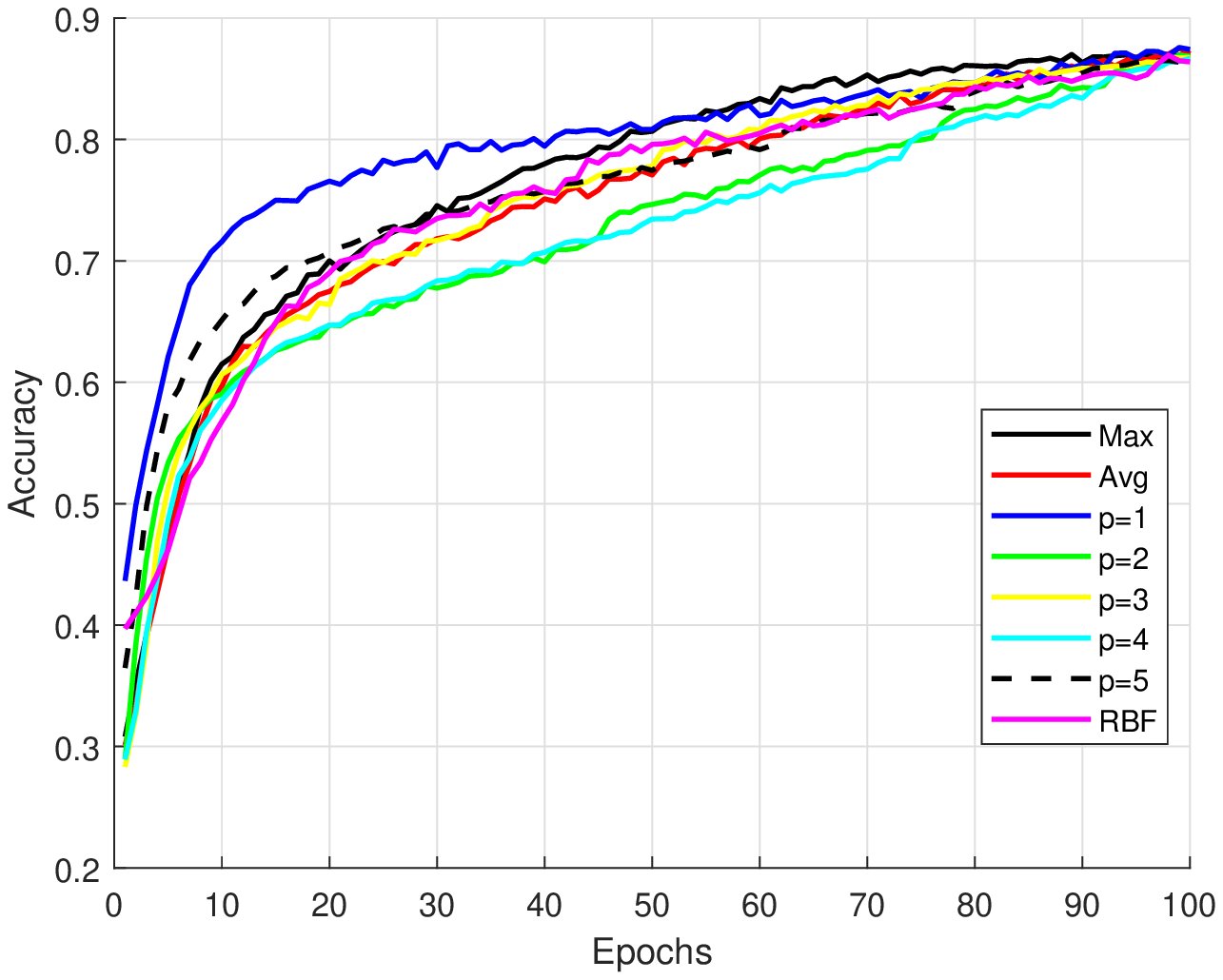}
                \caption{RAF-DB}
        \end{subfigure}%
        \begin{subfigure}[b]{0.33\textwidth}
                \centering
                \includegraphics[width=.95\linewidth]{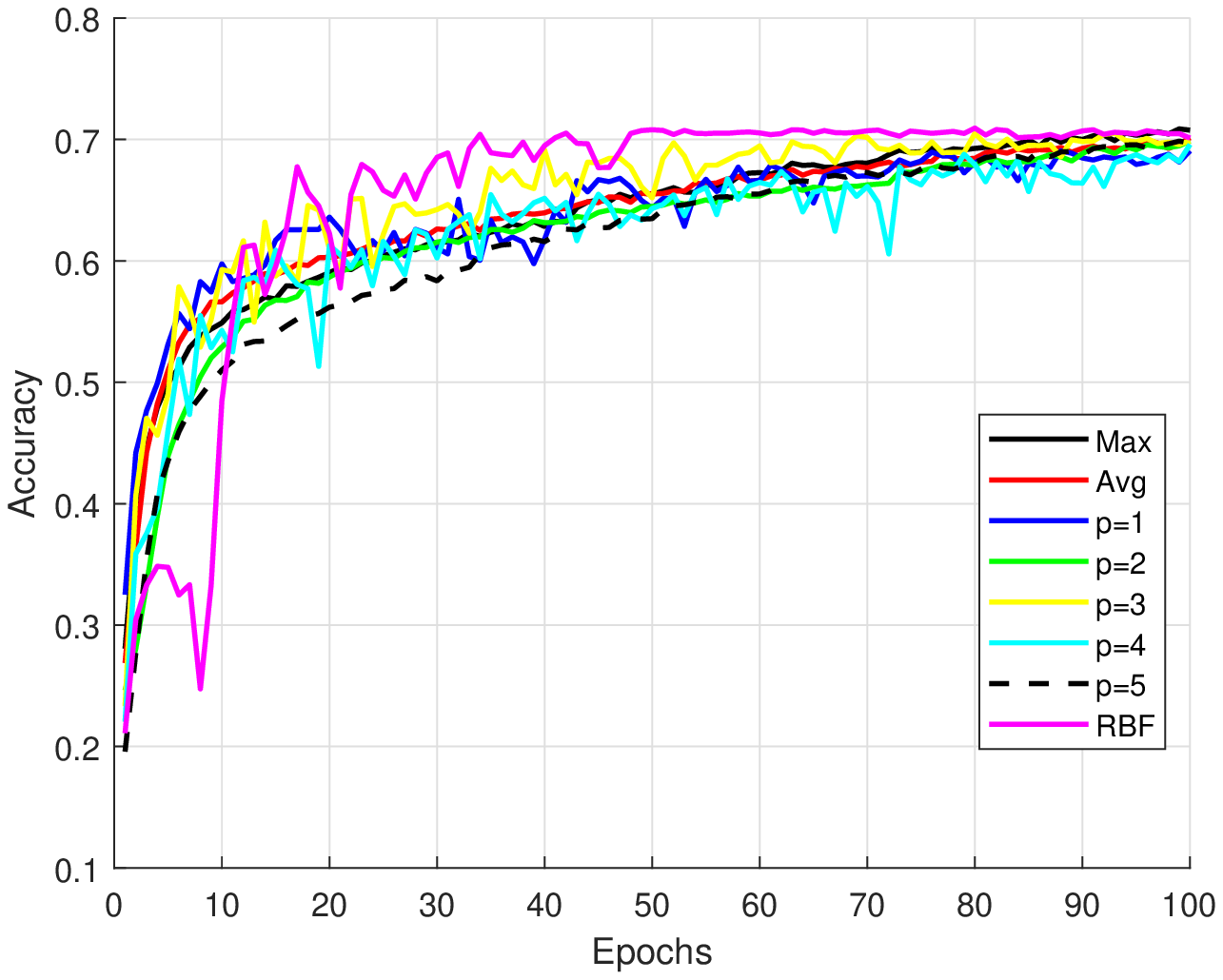}
                \caption{FER2013}
        \end{subfigure}%
        \begin{subfigure}[b]{0.33\textwidth}
                \centering
                \includegraphics[width=.95\linewidth]{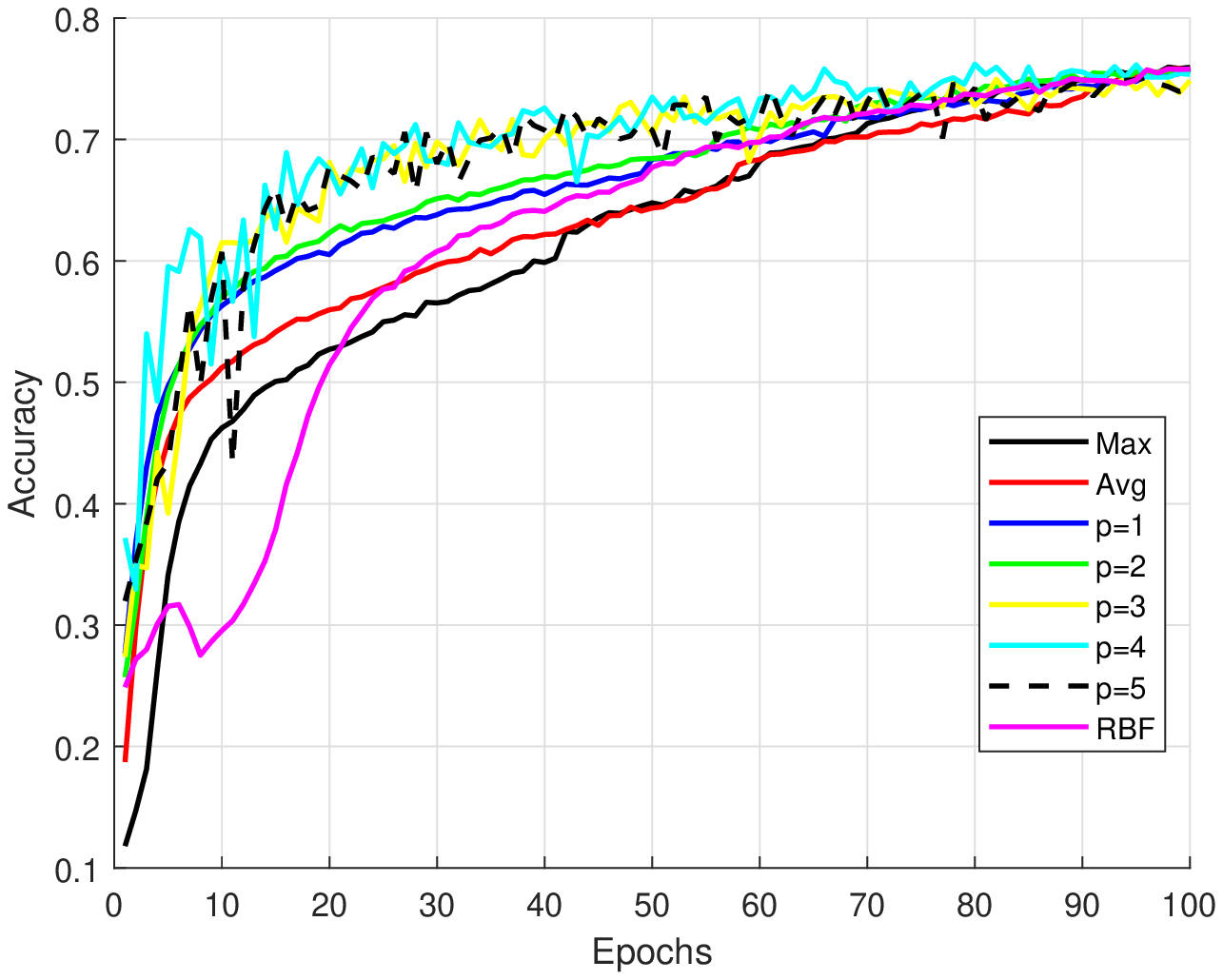}
                \caption{ExpW}
        \end{subfigure}
        \caption{Networks convergence with a single learnable pooling layer at the end}
        \label{fig:Learnablepoolinglast}
\end{figure*}

Similarly to kervolution, we explore the impact of using learnable pooling with three configurations, namely: full learnable pooling network (Table~\ref{tab:full-pooling}, Figure~\ref{fig:full-learnable pooling}), learnable pooling after the first convolution block (Table~\ref{tab:pooling-first}, Figure~\ref{fig:learnablepoolingfirst}) and learnable pooling at the end of the network (Table~\ref{tab:pooling-last}, Figure~\ref{fig:Learnablepoolinglast}). This time, we compare the performance of the network with the usual pooling methods, namely: Max and average pooling.

\begin{table}[H]
\caption{Accuracy rates of full learnable pooling network}
\label{tab:full-pooling}
\resizebox{\linewidth}{!}{
\begin{tabular}{lllllll}
\hline\noalign{\smallskip}
  & \multicolumn{2}{l}{Model-1}  & \multicolumn{4}{l}{Model-2}\\
\noalign{\smallskip}\hline\noalign{\smallskip}
Layers configuration  & MNIST & Fashion-MNIST  &  Cifar10 & RAF-DB & FER2013 & ExpW\\
\noalign{\smallskip}\hline\noalign{\smallskip}
Max     &  \textbf{99.05}\%  &   \textbf{91.34}\% &  88.42\% & 87.05\% & 70.49\% & 75.91\% \\
AVG   & \textbf{99.17}\% &   \textbf{91.37}\%  &  88.12\% & 86.89\% &70.13\% & 75.74\% \\
Linear kernel     &  98.85\%  &   \textbf{90.05}\% &  \textbf{90.25}\% & \textbf{90.81}\% & 70.96\% & 76.28\% \\
2$^{nd}$-order Poly   & \textbf{99.26}\% &   \textbf{91.42}\%  &  \textbf{90.62}\% & \textbf{92.87}\% &70.88\% & \textbf{76.64}\% \\
3$^{rd}$-order Poly    & \textbf{99.35}\% &  \textbf{91.79}\%  & \textbf{90.97}\% & \textbf{93.21}\% &\textbf{71.35}\% & \textbf{76.81}\%  \\
4$^{nd}$-order Poly   & 98.73\% &   88.56\%  &  89.88\%  & \textbf{93.03}\% &\textbf{71.13}\% &\textbf{76.73}\% \\
5$^{rd}$-order Poly    & 98.52\% &   87.91\%  & 89.61\% & \textbf{92.64}\% &69.91\% &75.56\% \\
Gaussian RBF $\sigma=0.9$    & \textbf{99.11}\% &   \textbf{91.32}\%  & \textbf{90.45}\% & \textbf{92.74}\% &70.74\% & 76.42\% \\
\noalign{\smallskip}\hline
\end{tabular}
}
\end{table}

Table~\ref{tab:full-pooling} presents the results of our base models with learnable pooling only. One can clearly notice that learnable pooling enhance the accuracy of the network with all kernel functions. Second and third order polynomial kernels are the most accurate kernels for pooling with all datasets. With third order polynomial kernel Model-1 surpassed usual pooling methods by 0.18\% on MNIST and 0.42\% on Fashion-MNIST. On the other hand, Model-2 outperformed usual pooling method by 2.55\% on Cifar-10, 6.15\% on RAF-DB, 0.86\% on FER2013 and 0.90\% on ExpW. We can also notice that for fine-grained FER datasets, higher order polynomial kernels are also beneficial for the model for the same reasons cited for kervolution at the beginning of the network. Indeed, learnable pooling uses less weights than convolution or kervolution which allows it to fit subtle details without over-fitting. Finally, Gaussian RBF learnable pooling performs better than usual pooling methods even-though it does not outperform polynomial kernels. Although it did not surpass usual pooling method on model-1, it outperformed ordinary pooling methods with Model-2 by 2.03\% on Cifar-10, 5.69\% on RAF-DB, 0.25\% on FER2013 and 0.51 on ExpW.

The convergence progression of full learnable pooling networks is shown in Figure~\ref{fig:full-learnable pooling}. We can notice that second and third order polynomial kernels are the fastest to converge. On the other hand, higher order polynomial kernels convergence is the slowest among all kernels, even if they reach higher accuracy rates than Max and average pooling for FER datasets. Finally, Gaussian RBF kernel has similar accuracy progression to Max and average pooling.


\begin{table}[h]
\caption{Accuracy rates networks with a single learnable pooling layer at the begining}
\label{tab:pooling-first}
\resizebox{\linewidth}{!}{
\begin{tabular}{lllllll}
\hline\noalign{\smallskip}
  & \multicolumn{2}{l}{Model-1}  & \multicolumn{4}{l}{Model-2}\\
\noalign{\smallskip}\hline\noalign{\smallskip}
Layers configuration  & MNIST & Fashion-MNIST  &  Cifar10 & RAF-DB & FER2013 & ExpW\\
\noalign{\smallskip}\hline\noalign{\smallskip}
Max     &  \textbf{99.05}\%  &   \textbf{91.34}\% &  88.42\% & 87.05\% & 70.49\% & 75.91\% \\
AVG   & \textbf{99.17}\% &   \textbf{91.37}\%  &  88.12\% & 86.89\% &70.13\% & 75.74\% \\
Linear kernel     &  \textbf{99.11}\%  &   \textbf{90.58}\% &  \textbf{88.15}\% & \textbf{87.19}\% & \textbf{70.26}\% & \textbf{75.95}\% \\
2$^{nd}$-order Poly   & 98.97\% &   89.80\%  &  87.71\% & 86.92\% & 70.08\% & 75.82\% \\
3$^{rd}$-order Poly    & \textbf{99.24}\% &  89.88\%  & 87.54\% & 86.79\% & 69.92\% & 75.23\%  \\
4$^{nd}$-order Poly   & \textbf{99.12}\% &   \textbf{90.32}\%  &  87\%  & 86.57\% & 69.79\% & 74.84\% \\
5$^{rd}$-order Poly    & \textbf{99.10}\% &   \textbf{90.24}\%  & 86.89\% & 86.38\% & 69.54\% & 74.88\% \\
Gaussian RBF $\sigma=0.9$    & 98.43\% &   88.93\%  & 85.71\% & 84.12\% &68.63\% & 73.21\% \\
\noalign{\smallskip}\hline
\end{tabular}
}
\end{table}
In table~\ref{tab:pooling-first} we notice that using one learnable pooling layer after the first convolution block does not perform as well as the full learnable pooling network. Even-though linear kernel outperforms the usual pooling methods. As stated before, linear kernel pooling can learn a suitable pooling from a continuum of methods that ranges from average to max pooling. This is the reason why linear kernel pooling performs well wherever it is plugged.

The convergence progression of learnable pooling at the beginning of the networks is shown in Figure~\ref{fig:learnablepoolingfirst}. We can notice that polynomial kernels are, in general, the fastest to converge. Second and third order polynomial kernels are still the most efficient among these kernels. On the other hand, Gaussian RBF kernel is the slowest kernel to converge, specially for the biggest FER datasets (i.e. FER2013 and ExpW). This kernel is more sensitive to the learning rate used. As we can see in Figure~\ref{fig:learnablepoolingfirst}, Gaussian RBF kernel does not start to converge until reaching a specific learning rate.

\begin{table}[h]
\caption{Accuracy rates networks with a single learnable pooling layer at the end}
\label{tab:pooling-last}
\resizebox{\linewidth}{!}{
\begin{tabular}{lllllll}
\hline\noalign{\smallskip}
  & \multicolumn{2}{l}{Model-1}  & \multicolumn{4}{l}{Model-2}\\
\noalign{\smallskip}\hline\noalign{\smallskip}
Layers configuration  & MNIST & Fashion-MNIST  &  Cifar10 & RAF-DB & FER2013 & ExpW\\
\noalign{\smallskip}\hline\noalign{\smallskip}

Max     &  \textbf{99.05}\%  &   \textbf{91.34}\% &  88.42\% & 87.05\% & 70.49\% & 75.91\% \\
AVG   & \textbf{99.17}\% &   \textbf{91.37}\%  &  88.12\% & 86.89\% &70.13\% & 75.74\% \\
Linear kernel     &  98.93\%  &   90.42\% &  87.52\% & \textbf{87.12}\% & \textbf{70.06}\% &\textbf{75.87}\% \\
2$^{nd}$-order Poly   & 98.85\% &   89.94\%  &  87.28\% & 86.89\% & 69.87\% &75.66\% \\
3$^{rd}$-order Poly    & 98.78\% &  89.72\%  & 87.04\% & 86.75\% & 69.71\% &75.48\%  \\
4$^{nd}$-order Poly   & 98.62\% &   89.37\%  &  86.79\%  & 86.69\% & 69.58\% &75.36\% \\
5$^{rd}$-order Poly    & 98.90\% &   89.10\%  & 86.63\% & 86.42\% & 69.22\% &74.81\% \\
Gaussian RBF $\sigma=0.9$    & \textbf{99.03}\% &   \textbf{90.36}\%  & \textbf{88.09}\% & \textbf{86.94}\% & \textbf{70.11}\% & \textbf{75.79}\% \\
\noalign{\smallskip}\hline
\end{tabular}
}
\end{table}

Similarly to learnable pooling at the beginning of the network (Table~\ref{tab:pooling-first}), using one learnable pooling layer after the last convolution layer and before fully connected layers (Table~\ref{tab:pooling-last}) does not perform as well as the full learnable pooling network. One exception is the Gaussian RBF kernel which slightly performs better at the end than at the beginning. This last remark strengthens the deduction made on kervolution at the end of the network which states that Gaussian RBF kernel is best suited for classification than feature extraction or even pooling.

The convergence speed of learnable pooling at the end of the network is illustrated in Figure~\ref{fig:Learnablepoolinglast}. One can clearly notice that kernel based pooling does not perform as well as when plugged at the beginning or in a full configuration. It performs similarly to the usual pooling methods. On the other hand, Gaussian RBF kernels have a faster convergence when used at the end of the network. Moreover, it outperforms other pooling method, specially with FER dataset. This confirms the assumption made above that Gaussian RBF kernel is more efficient when used just before the fully connected layers.


\subsubsection{kernelized Dense Layers}

kernelized Dense Layers are only plugged at the end of the network since they replace the fully connected layers. Table~\ref{tab:KDL} shows the result of using kernelized Dense Layers instead of the usual fully-connected layers. One can clearly notice that using kernel function for classification improves the accuracy of the network. For instance, polynomial kernel enhances the accuracy of the network for about 1\% with fine-grained FER datasets. Third and Fourth order polynomial kernels are those who gave the best result with all datasets. On the other hand, Gaussian RBF kernel results confirm the deduction made trough this experimental section that it is best suited for classification than feature extraction. Whenever it is used at the end of the network it performs better than when used in the beginning.

\begin{table}
\caption{Accuracy rates of networks with KDL}
\label{tab:KDL}
\resizebox{\linewidth}{!}{
\begin{tabular}{lllllll}
\hline\noalign{\smallskip}
  & \multicolumn{2}{l}{Model-1}  & \multicolumn{4}{l}{Model-2}\\
\noalign{\smallskip}\hline\noalign{\smallskip}
Layers configuration  & MNIST & Fashion-MNIST  &  Cifar10 & RAF-DB & FER2013 & ExpW\\
\noalign{\smallskip}\hline\noalign{\smallskip}

Linear kernel     &  \textbf{99.07\%}  &   \textbf{90.07}\% &  86.92\% & \textbf{87.05}\% & 70.49\% & 75.91\% \\
2$^{nd}$-order Poly   & \textbf{99.12\%} &   \textbf{90.73}\%  &  87.61\% & \textbf{87.64}\% &70.85\% &76.13\% \\
3$^{rd}$-order Poly    & \textbf{99.11}\% &  \textbf{90.63}\%  & \textbf{89.37}\% & \textbf{88.12}\% &\textbf{71.28}\% &\textbf{76.64}\%  \\
4$^{nd}$-order Poly   & \textbf{99.09\%} &   \textbf{90.30}\%  &  \textbf{89.09}\%  & \textbf{87.83}\% &\textbf{71.13}\% &\textbf{76.42}\% \\
5$^{rd}$-order Poly    & \textbf{99.01}\% &   \textbf{91.07}\%  & 88.95\% & 86.93\% &70.62\% &75.86\% \\
Gaussian RBF $\sigma=0.9$    & \textbf{99.23}\% &   \textbf{90.79}\%  & \textbf{89.11}\% & \textbf{88.03}\% & \textbf{71.06}\% & \textbf{76.51}\% \\
\noalign{\smallskip}\hline
\end{tabular}
}
\end{table}

\begin{figure*}
        \begin{subfigure}[b]{0.33\textwidth}
                \centering
                \includegraphics[width=.95\linewidth]{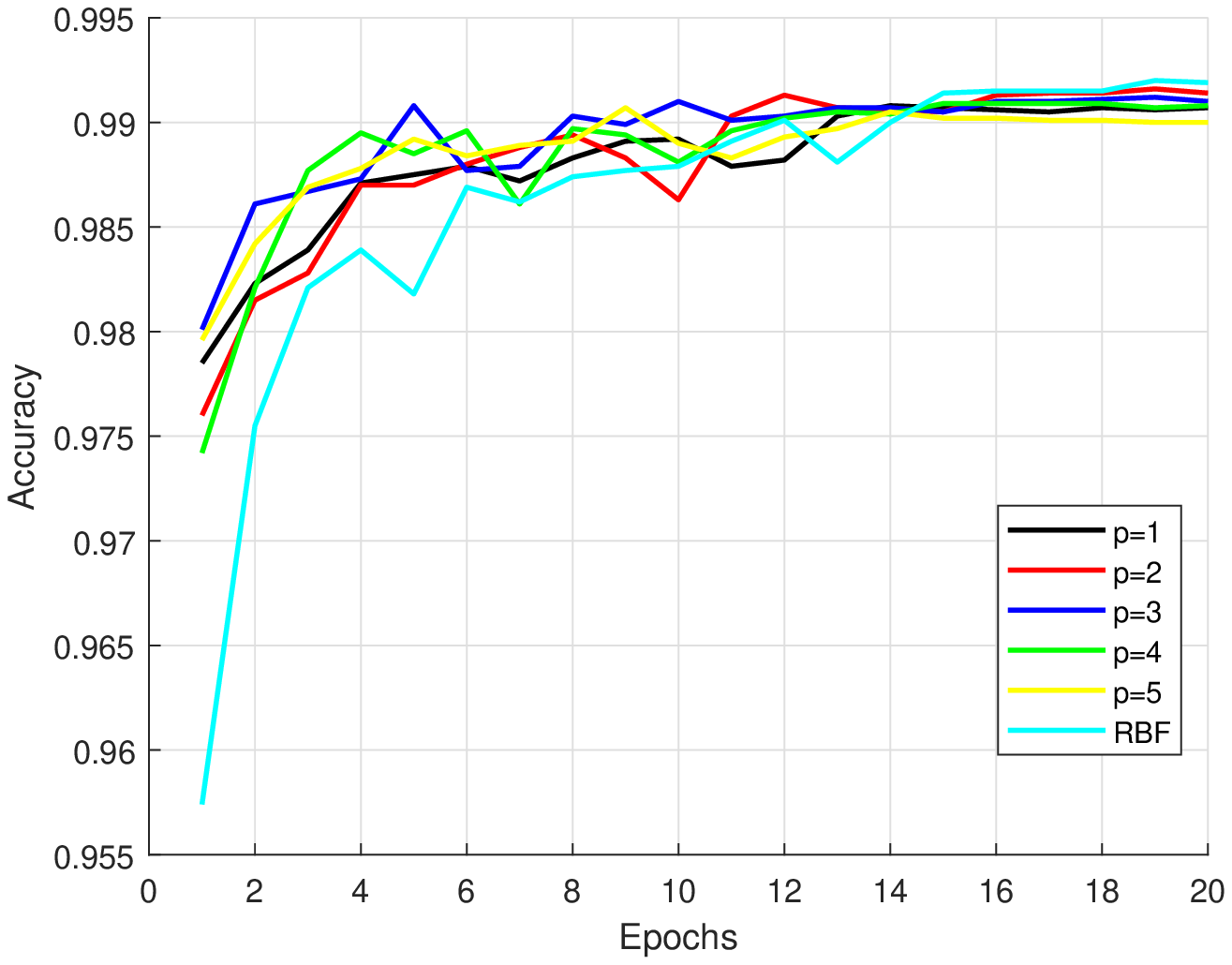}
                \caption{MNIST}
        \end{subfigure}%
        \begin{subfigure}[b]{0.33\textwidth}
                \centering
                \includegraphics[width=.95\linewidth]{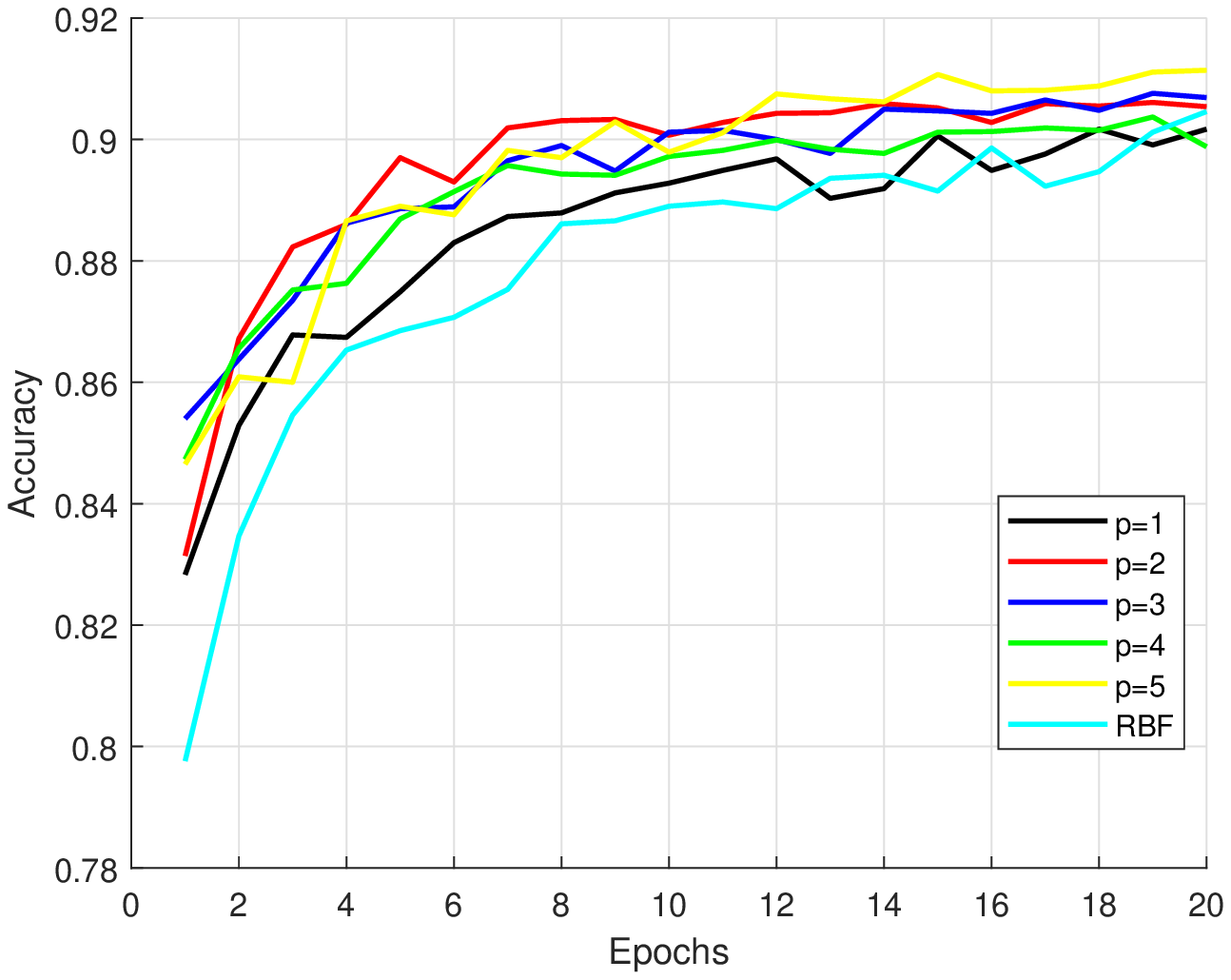}
                \caption{Fashion MNIST}
        \end{subfigure}%
        \begin{subfigure}[b]{0.33\textwidth}
                \centering
                \includegraphics[width=.95\linewidth]{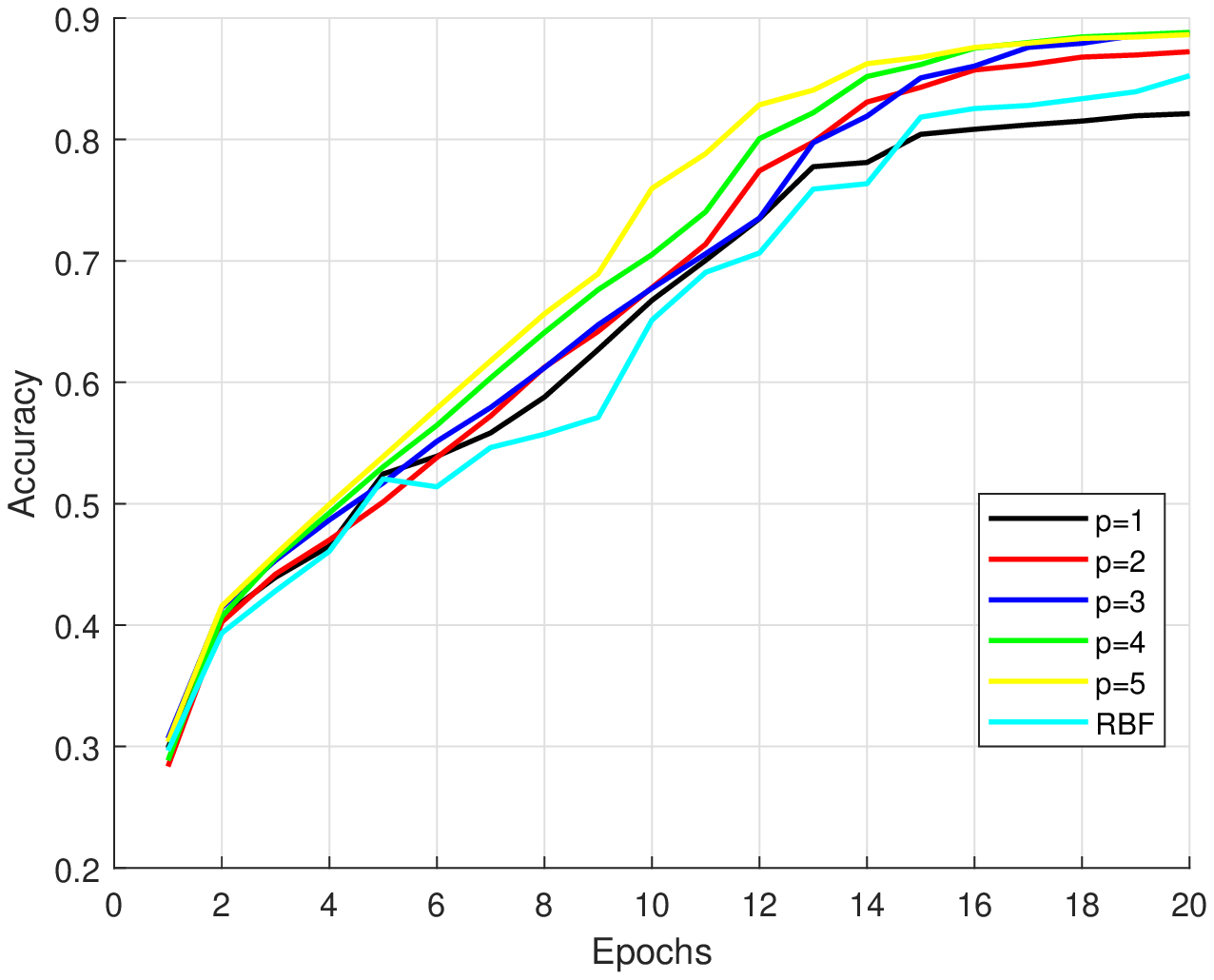}
                \caption{Cifar 10}
        \end{subfigure}
        \begin{subfigure}[b]{0.33\textwidth}
                \centering
                \includegraphics[width=.95\linewidth]{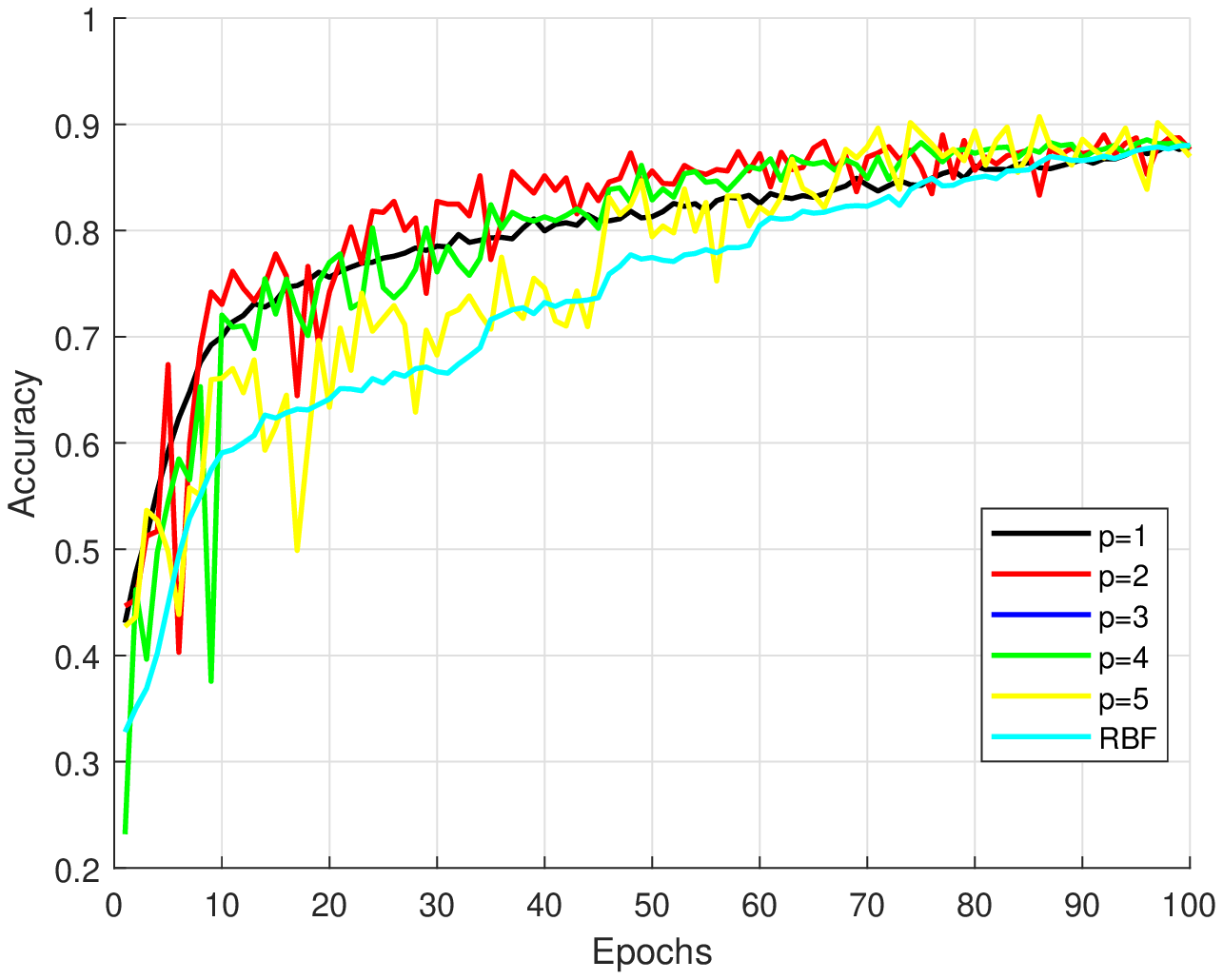}
                \caption{RAF-DB}
        \end{subfigure}%
        \begin{subfigure}[b]{0.33\textwidth}
                \centering
                \includegraphics[width=.95\linewidth]{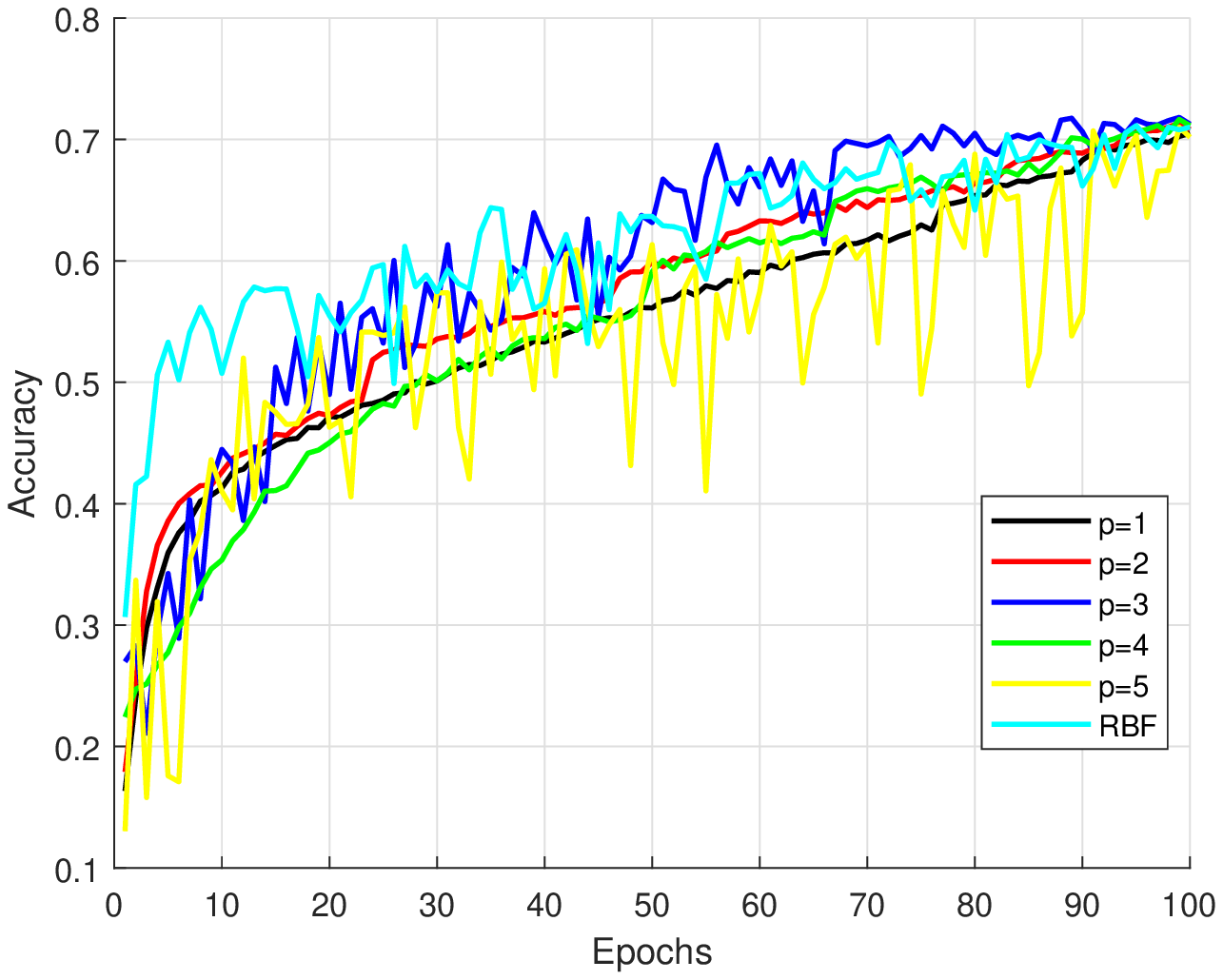}
                \caption{FER2013}
        \end{subfigure}%
        \begin{subfigure}[b]{0.33\textwidth}
                \centering
                \includegraphics[width=.95\linewidth]{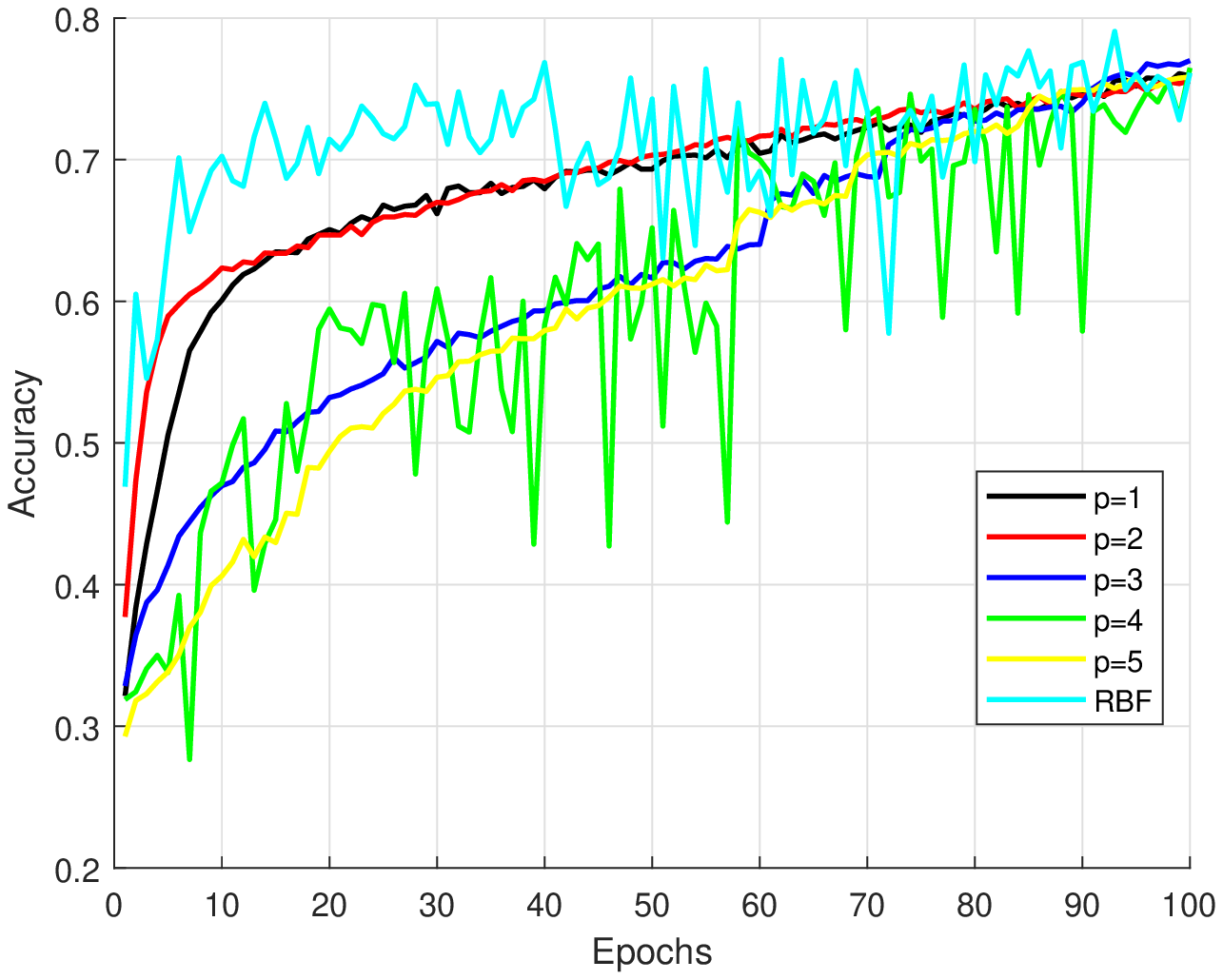}
                \caption{A ExpW}
        \end{subfigure}
        \caption{Networks convergence with KDL}
        \label{fig:KDL}
\end{figure*}

Figure~\ref{fig:KDL} shows the convergence of networks with KDL. We can say that polynomial kernels converge faster than other kernels. Even-though second and third order polynomial reach better accuracy rates than other polynomial kernels, they have not better convergence. On the other hand, Gaussian RBF kernel shows the slowest convergence even if it gives the best accuracy rates.

We have also tried to combine these kernel-based methods together in the same network. The number of all possible configurations is very big, therefore we only show the configuration which gave the best results. These results are shown in Table~\ref{tab:best-conf}. All the combinations tested shown a dramatically decreasing in performance. The only reasonable performances we could attain were using one kervolution layer at the beginning of the network and KDL at the end. Yet we have noticed that this configuration shows a clear over fitting.

\begin{table}
\caption{Accuracy rates of networks with best combinations}
\label{tab:best-conf}
\resizebox{\linewidth}{!}{
\begin{tabular}{lllllll}
\hline\noalign{\smallskip}
  & \multicolumn{2}{l}{Model-1}  & \multicolumn{4}{l}{Model-2}\\
\noalign{\smallskip}\hline\noalign{\smallskip}
Layers configuration  & MNIST & Fashion-MNIST  &  Cifar10 & RAF-DB & FER2013 & ExpW\\
\noalign{\smallskip}\hline\noalign{\smallskip}

\multicolumn{7}{l}{Single Kervolution layer at the beginning and KDL}\\
\hline
2$^{rd}$-order Poly (1kerv-KDL)   & \textbf{99.16}\% & \textbf{91.19}\%  & 65.86\% & 72.81\% &62.41\% & 58.72\%  \\
3$^{rd}$-order Poly (1kerv-KDL)   & 98.96\% &  90.23\%  & 65.75\% & 72.64\% & 61.73\% & 57.96\%  \\
4$^{nd}$-order Poly (1kerv-KDL)  & 98.88\% &   89.64\%  &  61.87\%  & 69.58\% &59.86\% &57.41\% \\
Gaussian RBF $\sigma=0.9$ (1kerv-KDL)   & 98.57\% &   88.92\%  & 61.20\% & 68.12\% & 59.12\% &56.87\% \\
\noalign{\smallskip}\hline
\end{tabular}
}
\end{table}

\subsection{Comparison with the State-of-the-Art}
\label{comp-sota}

In this section, we compare the performance of the proposed methods, namely: kervolution, learnable pooling and KDL, with respect to several state-of-the-art methods. According to table~\ref{tab:sota}, we obtained state-of-the-art results on two FER datasets (RAF-DB and ExpW) with all the proposed methods. For the remaining datasets namely MNIST, Fashion MNIST,  Cifar-10 and FER2013, we could not surpass state-of-the-art results, yet we obtained very close results. In the following, we give the best results of each configuration.

With full kervolution network (Table~\ref{tab:full-kerv}), the highest accuracy rate we reached, was using third order polynomial kernel. It gave \textbf{99.16\%} on MNIST, \textbf{87.93\%} on RAF-DB and \textbf{76.32\%} on ExpW. Kervolution attains better results when used at the beginning of the network (Table~\ref{tab:kerv-first}). With this configuration, we got \textbf{99.04} on MNIST, \textbf{88.73\%} on RAF-DB and \textbf{76.85\%} on ExpW using third order polynomial kernel. On the other hand, when using kervolution at the end of the network (Table~\ref{tab:kerv-last}), the Gaussian RBF reaches the best results.  It attains \textbf{89.36\%} on RAF-DB and \textbf{77.21\%} on ExpW.

Learnable pooling is the method that allowed to reach the best results. With the full learnable pooling configuration (Table~\ref{tab:full-pooling}), we could reach the best results on all datasets using third order polynomial kernel. We got very close results to state-of-the-art with \textbf{99.35\%} on MNIST, \textbf{91.79\%} on Fashion MNIST, \textbf{90.97\%} on Cifar-10 and \textbf{71.35\%} on FER2013. We have also reached state-of-the-art result on RAF-DB and ExpW with respectively \textbf{93.21\%} and \textbf{ 76.81\%}. Using learnable pooling after the first convolution block (Table~\ref{tab:pooling-first}), we could not surpass full learnable pooling network. The best results were obtained with linear kernel with \textbf{87.19\%} on RAF-DB and \textbf{75.95\%} on Expw. The same results were obtained when using learnable pooling at the end of the network, with a slight improvement for Gaussian RBF kernel. Yet linear kernel remains the more accurate with \textbf{87.12\%} for RAF-DB and \textbf{75.87\%} for ExpW.

Finally, KDL (Table~\ref{tab:KDL}) gives very good results with all the used kernels. On MNIST dataset, all the results are above \textbf{99\%} and reached \textbf{99.23\%} with Gaussian RBF kernel. It also reached \textbf{88.03\%} on RAF-DB and \textbf{76.51} on ExpW. The best KDL results were obtained with third order polynomial kernel. It reached \textbf{88.12\%} on RAF-DB and \textbf{76.64\%} on ExpW.

\section{Discussion}
\label{sec:Discussion}

In this paper, we extensively studied the impact of using kernel functions on different levels of a CNN, in particular, convolution, pooling and fully-connected layers. We replaced convolution layer by a non-linear layer as introduced by Wang et al.~\citep{wang2019kervolutional}. We tested the performance of this layer following three network configurations. We first tested them solely, in a full configuration network. After that, we tested them jointly with the usual convolution layers, plugging them either at the beginning of the network or at the end.

We have also used a novel pooling layer, based on kernel functions, that keeps the down-scaling aspect of the standard pooling function and brings various new features. This pooling layer relay on learnable weights that generalize ordinary pooling operations (i.e. average pooling and max pooling). Furthermore, it encodes patch-wise non-linearity. In this manner, the discrimination power of the full network is enhanced. The novel pooling, called learnable weights pooling, can be used at any level of the network and is fully differentiable, which allows the network to be trained in an end-to-end training. We have also explored their impact following the same network configuration as stated above.

We also use a novel fully-connected layer in which we use kernel functions to create a neuron unit that uses a higher degree kernel function on its inputs rather than computing the weighted sum. The used layer, called Kernelized Dense Layers (KDL), is also differentiable and demonstrates its usefulness in the improvement of the discrimination power of the full network.

For the three proposed layers, we explored their impact on the overall accuracy and convergence speed of the network. The results illustrated in Tables~\ref{tab:full-kerv}-~\ref{tab:KDL} and Figures~\ref{fig:full-kerv}-~\ref{fig:KDL} allow us to deduce the following. Kervolution layer is best suited for feature extraction and gives the best of its results when used at beginning of the network conjointly with convolution layers. It is more accurate with fine-grained FER datasets which means that it is more sensitive to subtle details. Adding more kervolution layers only increases over-fitting. While using kervolution at the end of the network decreases its performance. On the other hand, learnable pooling works best when used in full network configuration. It has the same advantages of kervolution in term of being more sensitive to subtle details with fewer parameters, which prevents it from over-fitting. The performance of learnable pooling decreases when used only at the beginning of the network or at the end, compared to the full network configuration. Finally, KDL is the most stable proposed layer. It shows good performance with all kernels and datasets.

\begin{table}

\caption{Accuracy rates of networks with best combinations}
\label{tab:sota}
\resizebox{\linewidth}{!}{
\begin{tabular}{lllllll}
\hline\noalign{\smallskip}
  & \multicolumn{2}{l}{Model-1}  & \multicolumn{4}{l}{Model-2}\\
\noalign{\smallskip}\hline\noalign{\smallskip}
Layers configuration  & MNIST & Fashion-MNIST  &  Cifar10 & RAF-DB & FER2013 & ExpW\\
\noalign{\smallskip}\hline\noalign{\smallskip}

\multicolumn{7}{l}{Full kervolution networks}\\
\noalign{\smallskip}\hline
3$^{rd}$-order Poly    & \textbf{99.16}\% & \textbf{90.04} \%  & \textbf{88.64}\% & \textbf{87.93}\% &\textbf{70.95}\% &\textbf{76.32}\%  \\
Gaussian RBF $\sigma=0.9$    & 98.61\% &   88.78\%  & 87.51\% & \textbf{87.33}\% &\textbf{70.78}\% & \textbf{76.23}\% \\
\noalign{\smallskip}\hline
\multicolumn{7}{l}{Single kervolution layer at the beginning}\\
\noalign{\smallskip}\hline
3$^{rd}$-order Poly    & \textbf{99.04}\% &  \textbf{90.26}\%  & \textbf{88.73}\% & \textbf{88.06}\% &\textbf{71.06}\% &\textbf{76.85}\%  \\
Gaussian RBF $\sigma=0.9$    & 98.74\% &   89.53\%  & \textbf{88.48}\% & \textbf{87.89}\% &\textbf{70.98}\% &76.75\% \\
\noalign{\smallskip}\hline
\multicolumn{7}{l}{Single kervolution layer at the end}\\
\noalign{\smallskip}\hline
3$^{rd}$-order Poly    & 98.84\% &  88.45\%  & 87.23\% & 88.03\% & 70.95\% &76.82\%  \\
Gaussian RBF $\sigma=0.9$    & \textbf{98.75}\% &   \textbf{89.33}\%  & \textbf{90.13}\% & \textbf{89.36}\% &\textbf{71.15}\% &\textbf{77.21}\% \\
\noalign{\smallskip}\hline
\multicolumn{7}{l}{Full learnable pooling}\\
\noalign{\smallskip}\hline
3$^{rd}$-order Poly    & \textbf{99.35}\% &  \textbf{91.79}\%  & \textbf{90.97}\% & \textbf{93.21}\% &\textbf{71.35}\% & \textbf{76.81}\%  \\
Gaussian RBF $\sigma=0.9$    & \textbf{99.11}\% &   \textbf{91.32}\%  & \textbf{90.45}\% & \textbf{92.74}\% &70.74\% & 76.42\% \\
\noalign{\smallskip}\hline
\multicolumn{7}{l}{Learnable pooling at the beginning}\\
\hline
3$^{rd}$-order Poly    & \textbf{99.24}\% &  89.88\%  & 87.54\% & 86.79\% & 69.92\% & 75.23\%  \\
Gaussian RBF $\sigma=0.9$    & 98.74\% &   89.53\%  & \textbf{88.48}\% & \textbf{87.89}\% &\textbf{70.98}\% &\textbf{76.75}\% \\
\noalign{\smallskip}\hline
\multicolumn{7}{l}{Learnable pooling at the end}\\
\noalign{\smallskip}\hline
Linear kernel     &  98.93\%  &   90.42\% &  87.52\% & \textbf{87.12}\% & \textbf{70.06}\% &\textbf{75.87}\% \\
Gaussian RBF $\sigma=0.9$    & \textbf{99.03}\% &   \textbf{90.36}\%  & \textbf{88.09}\% & \textbf{86.94}\% & \textbf{70.11}\% & \textbf{75.79}\% \\
\noalign{\smallskip}\hline
\multicolumn{7}{l}{Kernelized Dense Layer}\\
\noalign{\smallskip}\hline
3$^{rd}$-order Poly    & \textbf{99.11}\% &  \textbf{90.63}\%  & \textbf{89.37}\% & \textbf{88.12}\% &\textbf{71.28}\% &\textbf{76.64}\%  \\
Gaussian RBF $\sigma=0.9$    & \textbf{99.23}\% &   \textbf{90.79}\%  & \textbf{89.11}\% & \textbf{88.03}\% & \textbf{71.06}\% & \textbf{76.51}\% \\
\noalign{\smallskip}\hline
\multicolumn{7}{l}{State-of-the-art results}\\
\noalign{\smallskip}\hline
Byerly et al~\citep{byerly2020branching}&\textbf{99.84\%}&--&--&--&--&--\\
Assiri~\citep{assiri2020stochastic}&\textbf{99.83\%}&--&--&--&--&--\\
Jayasundara et al~\citep{jayasundara2019textcaps}&\textbf{99.71\%}&\textbf{96.36\%}&--&--&--&--\\
Kolesnikov et al~\citep{kolesnikov2019large}&--&--&\textbf{99.30\%}&--&--&--\\
Huang et al~\citep{huang2019gpipe}&--&--&\textbf{99\%}&--&--&--\\
Ridnik et al~\citep{ridnik2020tresnet}&--&--&\textbf{99\%}&--&--&--\\
Tang et al.~\citep{tang2013deep}  & --&--&--&  -- & --   &   71.16\%  \\
Guo et al.~\citep{guo2016deep}&--&--&--& --&--&   71.33\%\\
Kim et al.~\citep{kim2016fusing}&--&--&--&--&--&   \textbf{73.73\%}\\
Bishay et al.~\citep{bishay2019schinet}  &--&--&--& -- & \textbf{73.10\%} &   --    \\
Lian et al.~\citep{lianexpression}  &--&--&--& -- & 71.90 \% &   --    \\
Acharya et al.~\citep{acharya2018covariance} &--&--&--&\textbf{87\%} & --  & --  \\
S Li et al.~\citep{li2018reliable} &--&--&--&74.20\% & --  &--  \\
Z.Liu et al.~\citep{liu2017boosting} &--&--&--&73.19\% & --  &--  \\
\noalign{\smallskip}\hline
\end{tabular}
}
\end{table}

In terms of kernels, we have also noticed that they may perform differently when used at different levels. For instance, polynomial kernels work best as feature extractors. They work better when used at the beginning of the network only; as stated in~\citep{wang2019kervolutional}. On the other hand, Gaussian RBF kernels work best when used at the end of the network. Either as feature extraction or classification layer.

\section{Conclusion}

In light of the results obtained in the experiments, we have conducted in this work, we deduce that kernel functions impact positively the CNN performance. Indeed, the experimental results prove that the use of kernel functions, instead of the linear functions used in CNN layers, improves the accuracy rate of the whole model and its convergence speed. Furthermore, we concluded that the impact of these kernel functions varies according to the level in which they are plugged into and the specific kernel in use. We have also concluded that high order kernels prone to overfitting, whereas low order kernels might not be sufficiently effective to fit the input data distributions. Therefore, a trade-off between complexity and performance should be reached.


%

\bibliography{main}

\end{document}